\documentclass[manuscript,screen, anonymous=False, nonacm]{acmart}

\usepackage{xcolor}
\usepackage{pifont}   
\usepackage{tikz}
\usetikzlibrary{shapes.geometric, arrows.meta, positioning}

\usepackage{url}
\usepackage{tcolorbox}

\usepackage{microtype}
\usepackage{graphicx}
\usepackage{booktabs} 
\usepackage{algorithm}
\usepackage{multirow}
\usepackage{algpseudocode}
\usepackage[table]{xcolor}
\definecolor{best}{RGB}{128, 179, 255}
\definecolor{2best}{RGB}{252, 186, 182}
\definecolor{cornflowerblue}{rgb}{0.0, 0.72, 0.92}
\usepackage{appendix}
\usepackage{float} 
\usepackage{stfloats}
\usepackage{subcaption}
\usepackage{pgfplots}
\usepackage{orcidlink}
\usepackage{tikz}
\pgfplotsset{compat=1.18}
\usetikzlibrary{pgfplots.fillbetween}
\usepgfplotslibrary{groupplots}
\usepackage{graphicx}
\usepackage{subcaption}
\usepackage{tikz}
\definecolor{cvprblue}{rgb}{0.21,0.49,0.74}
\usepackage[capitalize,noabbrev]{cleveref}

\newenvironment{conditions}
  {\par\vspace{\abovedisplayskip}\noindent
   \begin{tabular}{>{$}l<{$} @{} >{${}}c<{{}$} @{} l}}
  {\end{tabular}\par\vspace{\belowdisplayskip}}
\AtBeginDocument{%
  \providecommand\BibTeX{{%
    \normalfont B\kern-0.5em{\scshape i\kern-0.25em b}\kern-0.8em\TeX}}}

\AtBeginDocument{%
  \providecommand\BibTeX{{%
    Bib\TeX}}}

\setcopyright{acmlicensed}
\copyrightyear{2026}
\acmYear{2026}
\acmDOI{XXXXXXX.XXXXXXX}

\setcopyright{rightsretained}
\acmJournal{TOPML}
\acmYear{2026} \acmVolume{X}
\begin{document}

\title{Last Layer Hamiltonian Monte Carlo}


\author{Koen Vellenga}
\affiliation{%
\orcidlink{0000-0003-2135-6615} 
  \institution{University of Sk\"{o}vde  \& Volvo Car Corporation}
  \country{Sweden}
}

\author{H. Joe Steinhauer}
\affiliation{%
\orcidlink{0000-0003-2949-4123}
  \institution{University of Sk\"{o}vde}
  \country{Sweden}
}

\author{G\"{o}ran Falkman}
\affiliation{%
\orcidlink{0000-0001-8884-2154}
  \institution{University of Sk\"{o}vde}
  \country{Sweden}
}

\author{Jonas Andersson}
\affiliation{%
  \institution{Volvo Car Corporation}
  \country{Sweden}
}

\author{Anders Sj\"{o}gren}
\affiliation{%
\orcidlink{0000-0003-2579-7063}
  \institution{Volvo Car Corporation}
  \country{Sweden}
}

\authorsaddresses{}








\renewcommand{\shortauthors}{Vellenga et al.}

\begin{abstract}
  We explore the use of Hamiltonian Monte Carlo (HMC) sampling as a probabilistic last layer approach for deep neural networks (DNNs). While HMC is widely regarded as a gold standard for uncertainty estimation, the computational demands limit its application to large-scale datasets and large DNN architectures. Although the predictions from the sampled DNN parameters can be parallelized, the computational cost still scales linearly with the number of samples (similar to an ensemble). Last layer HMC (LL--HMC) reduces the required computations by restricting the HMC sampling to the final layer of a DNN, making it applicable to more data-intensive scenarios with limited computational resources. In this paper, we compare LL-–HMC against five last layer probabilistic deep learning (LL--PDL) methods across three real-world video datasets for driver action and intention. We evaluate the in-distribution classification performance, calibration, and out-of-distribution (OOD) detection. Due to the stochastic nature of the probabilistic evaluations, we performed five grid searches for different random seeds to avoid being reliant  on a single initialization for the hyperparameter configurations. The results show that LL--HMC achieves competitive in-distribution classification and OOD detection performance. Additional sampled last layer parameters do not improve the classification performance, but can improve the OOD detection. Multiple chains or starting positions did not yield consistent improvements. Implementation is available at \url{https://github.com/koenvellenga/LL-HMC/}. 
\end{abstract}
%



\maketitle

\section{Introduction}
The rapid progress in Artificial Intelligence (AI) has amplified the need for robust and transparent decision making \citep{hendrycks2021unsolved,papamarkou2024position,bengio2024managing,manchingal2025position}. When using deep neural networks (DNN) in safety-critical environments, it is important that the model is able to identify instances it cannot confidently predict. Regular DNNs are unable to express uncertainty about their outputs, suffer from opaque decision-making processes \citep{rudin2019stop}, can be poorly calibrated \citep{guo2017calibration, nixon2019measuring}, struggle with noisy sensor observations \citep{berrada2021make}, require explicit safety verification \citep{wang2024empowering}, and perform poorly on long-tail edge cases \citep{hendrycks2021unsolved}. To address the lack of uncertainty estimation in regular DNNs, various methods have been proposed (e.g., \cite{neal2012bayesian,shafer2008tutorial,welling2011bayesian,hoffman2013stochastic,gal2016dropout, maddox2019simple, van2020uncertainty, wilson2020bayesian, franchi2024make}).  Hamiltonian Monte-Carlo (HMC) sampling is often regarded as the gold standard for uncertainty quantification and can yield top performance \citep{izmailov2021bayesian}. However, HMC can become too computationally expensive to estimate the uncertainty when scaling to bigger datasets and larger DNN architectures.

\begin{figure}[t]
    \centering
      \begin{subfigure}[b]{0.33\linewidth}
        \centering
        \includegraphics[width=\linewidth, trim=1.7cm 1.05cm 0.95cm 1.35cm, clip]{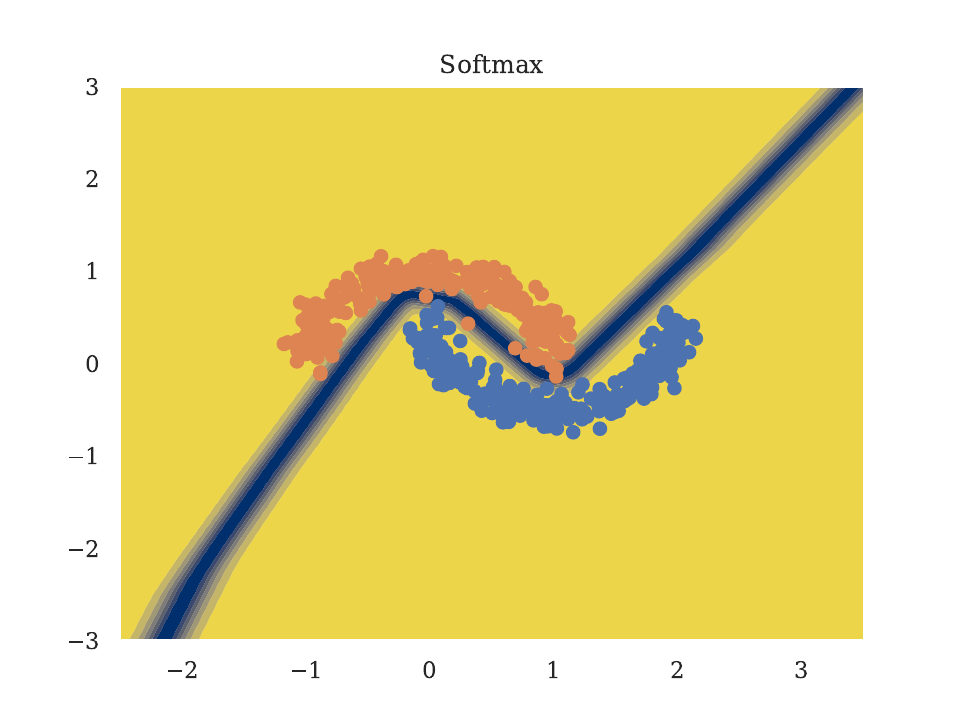} 
       \captionsetup{font=tiny}
        \caption{Softmax}
    \end{subfigure}
    \hfill
      \begin{subfigure}[b]{0.33\linewidth}
        \centering
        \includegraphics[width=\linewidth, trim=1.7cm 1.05cm 0.95cm 1.35cm, clip]{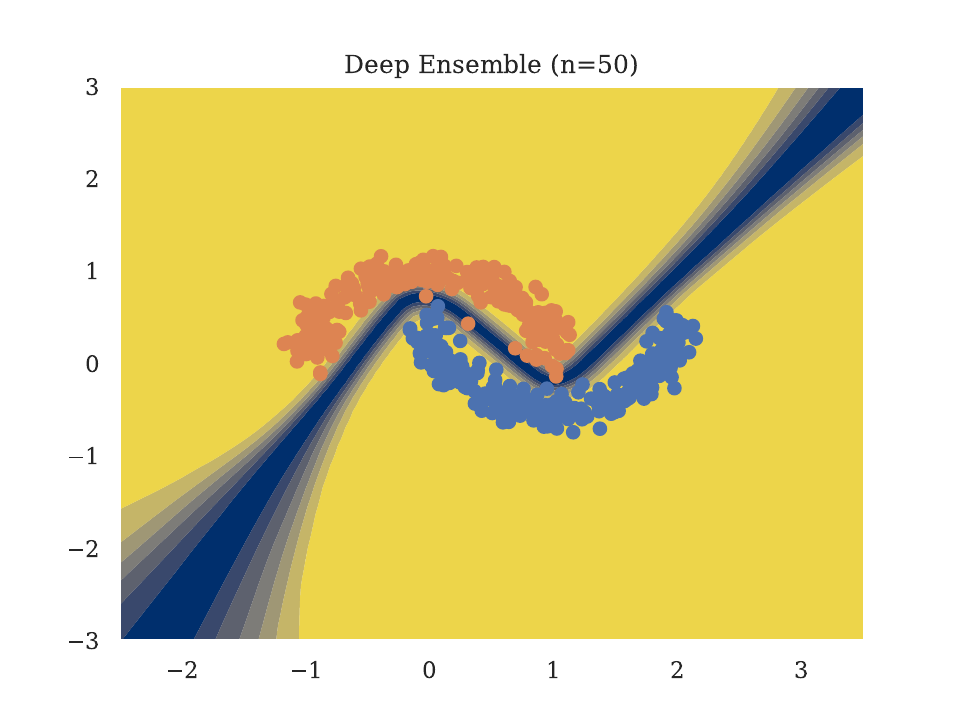} %
       \captionsetup{font=tiny}
        \caption{Deep Ensemble (N=50)}
    \end{subfigure}
    \hfill
      \begin{subfigure}[b]{0.33\linewidth}
        \centering
        \includegraphics[width=\linewidth, trim=1.7cm 1.05cm 0.95cm 1.35cm, clip]{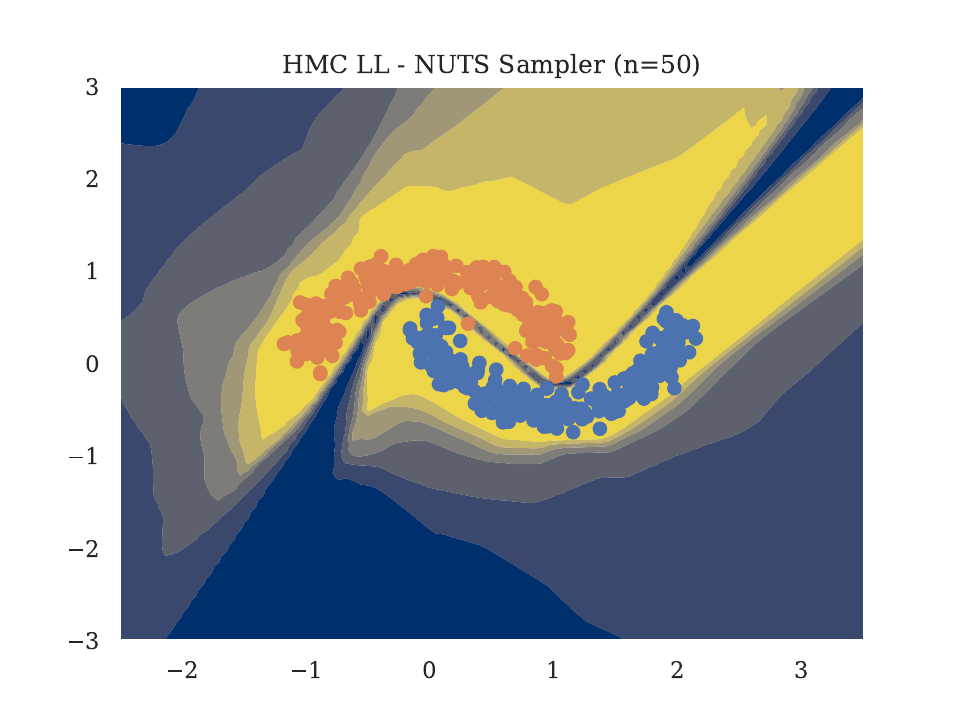} %
       \captionsetup{font=tiny}
        \caption{LL--HMC (N=50)}
    \end{subfigure}

    \caption{Uncertainty estimation example for the two moons toy dataset for a softmax, deep ensemble, and the last layer Hamiltonian Monte-Carlo methods. Yellow indicates certain areas, dark blue indicates uncertain areas (the color maps are normalized.).}
        \label{fig:example_sdllhmc}
\end{figure}

In machine learning it is common to decompose uncertainty into aleatoric uncertainty (randomness in the process) and epistemic uncertainty (the lack of knowledge) \citep{der2009aleatory}. Distinguishing between these two uncertainty types can help to identify the potentially reducible epistemic uncertainty \cite{hullermeier2021aleatoric}. In practice, it is challenging to decouple these uncertainty types \citep{mucsanyi2024benchmarking}. However, the predictive uncertainty (the aleatoric and epistemic uncertainty combined) can be used to evaluate, for example, whether a model expresses higher uncertainty for out-of-distribution (OOD) instances. The effectiveness of the predictive uncertainty depends on how well a probabilistic approach captures the underlying uncertainty. In classification tasks, the predictive distribution (e.g., softmax output) is sometimes assumed to be sufficient for identifying OOD instances. However, the softmax distribution is commonly only uncertain along the decision boundaries \citep{van2020uncertainty}, which limits its use for detecting unseen or ambiguous inputs (see Figure \ref{fig:example_sdllhmc}). Similarly, an ensemble improves the uncertainty estimations for regions where it has not observed any data compared to a single softmax-based uncertainty estimation. However, the ensemble mostly remains uncertain along the decision boundary. Additionally, the ensemble scales linearly with the number of ensemble members in terms of required computational resources to produce the estimations. 

Especially for safety-critical tasks that operate in a resource-constrained environment, we require efficient approaches to produce uncertainty estimates. Driving automation is an example safety-critical application domain with limited onboard computational resources, which requires continuous perception and decision-making under uncertainty. An example of a driving automation task is to accurately recognize the current actions and intentions of a driver to avoid future traffic conflicts. For such a human behavior application, estimating uncertainties and understanding the limitations of a model are essential. Therefore, in this paper we explore confining HMC sampling to the last layer of a DNN as an efficient probabilistic deep learning (PDL) approach. We evaluate last layer HMC (LL--HMC) on three driver action and intention recognition video datasets. We compare the classification performance and uncertainty-based OOD detection of LL--HMC to a deterministic softmax baseline, an ensemble, and five other probabilistic last layer approaches.

\section{Preliminaries}



\subsection{Probabilistic Deep Learning}
We provide a high-level explanation of a selection of PDL methods. For a comprehensive overview of Bayesian, PDL or stochastic approaches, refer to one of the following papers:  \cite{zhou2021survey,pitukbayesian,jospin2022hands,arbel2023primer,gawlikowski2023survey,papamarkou2024position,manchingal2025position}. \textcolor{black}{Regular training of DNNs commonly relies stochastic optimization methods (e.g., SGD, ADAM), which adjusts the model’s parameters by calculating the gradient of the loss function to find a local minimum. This results in a single point estimate of the parameters of a DNN. There exist multiple methods to estimate the uncertainty about the produced predictions. For example, a deep ensemble (DE) \citep{lakshminarayanan2017simple}  consists of multiple independently trained DNNs with random initializations, and can produce a diverse set of model parameters \citep{ashukhapitfalls,abe2022deep,laurentsymmetry}. However, a downside of a using a DE is that with each additional ensemble member we linearly increase the number of floating point operations (FLOPs) to produce an uncertainty estimate and a prediction (see Equation \ref{eq:de}). For example, if we take a driving automation scenario where we rely on processing 16 video frames with a 224 by 224 resolution through a the ViT-Base architecture \cite{tong2022videomae}, we estimated that this would require approximately $361 \times 10^9$ FLOPs. In this scenario, every additional ensemble member would increase the required FLOPs by $361 \times 10^9$ to produce a prediction for a single instance.}



\color{black}
\begin{equation}
\hat{y} = \frac{1}{N} \sum_{n=1}^N f_{\theta_n}(x),
\label{eq:de}
\end{equation}

where:
\begin{conditions}
$N$ &=&  number of ensemble members, and\\
f_{\theta_m}(x) &=&  output of the $n$-th DNN with parameters $\theta_m$. \\
\end{conditions}
\color{black}

\textcolor{black}{Packed-Ensembles (PE) \cite{laurent2023packed} aim to make DE more efficient by leveraging grouped convolutions to combine multiple smaller sub-networks into one larger network architecture. Grouped convolution is an operation that divides the input channels into separate groups and applies convolutions independently within each group, rather than across all channels together. PE uses grouped convolutions to combine multiple subnetworks into one network by associating each ensemble member with a separate group within the convolution layers. Each group processes a subset of the input channels independently, preserving the diversity of each subnetwork's learned representation.  As a more efficient alternative, multiple studies have considered restricting the stochastic component to the last layer of a DNN \citep{snoek2015scalable,watson2021latent}. In a sub-ensemble (SE) \cite{lee2015m, valdenegro2023sub} all layers of the DNN are shared except for the last layer (see Equation \ref{eq:se}). For our video-based driving automation scenario with a ViT-Base architecture, this means that for a downstream task with five classes, each additional sub-ensemble member would increase the number of FLOPS by $7.68 \times 10^3$. }
\color{black}

\begin{equation}
\hat{y} = \frac{1}{M} \sum_{m=1}^M h_{\theta_{LL_{m}}}( g_{\theta}(x)),
\label{eq:se}
\end{equation}

where:
\begin{conditions}
 M &=& the number of sub-ensemble members, \\
  g_{\theta} &=& the output of the penultimate layer (shared across all sub-ensemble members), and \\
 h_{\theta_{LL_{m}}} &=& the last layer parameters of the $m$-th sub-ensemble member.\\

\end{conditions}
\color{black}

\textcolor{black}{Deep Deterministic Uncertainty (DDU) \cite{mukhoti2023deep} is a sampling-free, distance-based uncertainty estimation method for DNNs. Unlike ensemble-based approaches, DDU does not rely on sampling multiple network parameters or multiple forward passes. Instead, it estimates uncertainty by modeling the density of learned feature representations in the feature space of a pre-trained DNN. The core idea of DDU is to apply Gaussian Discriminant Analysis (GDA) to this feature space post-training. GDA fits a multivariate Gaussian distribution to the features of each class, capturing their mean and covariance structure. DDU treats the latent representations extracted by the network as samples from class-conditional Gaussian distributions, with means and covariances computed from the training data. When using DDU to predict, the likelihood of a test instance’s latent representation under these Gaussian models reflects the uncertainty, where a low likelihood indicates OOD or unknown inputs where the model should express uncertainty.}

\textcolor{black}{Alternatively, Bayesian approaches let us learn about unknown parameters $\theta$ from observed data $x$ in an intuitive way. We start with a prior belief about $\theta$, then update this belief by considering how plausible the observed data is given different values of $\theta$. This plausibility is measured by the likelihood function $p(x|\theta)$, which quantifies how well the data would be expected given the parameters $\theta$. After observing the data, Bayes' rule combines the prior and the likelihood to give the posterior distribution $p(\theta|x)$, which describes what we know about $\theta$ after seeing the data. However, this posterior distribution is often intractable to compute exactly because it involves an integral over all possible parameter values in the denominator. This integral can be very complex or high-dimensional, making exact analytical solutions intractable. To overcome this, approximation methods, such as Variational Inference (VI) and Markov Chain Monte Carlo (MCMC) can be used to approximate the posterior \cite{salimans2015markov, blei2017variational, betancourt2017conceptual}.}

\textcolor{black}{Variational inference (VI) aims to approximate the intractable posterior distribution by replacing each parameter in a DNN with a variational distribution $q$. A common approach replaces each network parameter as a one-dimensional Gaussian distribution with learnable mean $\mu$ and standard deviation $\sigma$. To find the optimal variational distributions for all parameters, we compare how closely the variational distribution $q(\theta)$ approximates the true posterior $p(\theta | x)$. This comparison is done using the Kullback-Leibler (KL) divergence. However, since the true posterior is intractable, the evidence lower bound (ELBO) \cite{blei2017variational, jordan1999introduction} reformulates the KL divergence to separate the intractable term and define a tractable lower bound on the log marginal likelihood. Maximizing the ELBO thus provides a practical surrogate objective that implicitly minimizes the KL divergence between the variational distribution and the true posterior, yielding a tractable approximation. Backpropagating through randomly sampled variables from $q$ is not directly feasible. Bayes-by-Backprop (BBB, \cite{blundell2015weight}) addresses this by employing the reparameterization trick \cite{kingma2013auto}, which introduces an auxiliary noise variable $\epsilon \sim \mathcal{N}(0,1)$. This reparameterizes the sampling process so that a sample from $q$ can be expressed as a differentiable function of $\epsilon$, enabling gradient flow through the sampling step and thus allowing standard backpropagation.}


\textcolor{black}{To estimate uncertainty for a specific observation, BBB requires Monte Carlo sampling to draw multiple parameter samples. Similar to DE, this procedure proportionally increases the number of FLOPs for each additional prediction. However, similar to a SE, replacing only the parameters of the last layer of the DNN can significantly reduce the number of FLOPs to produce uncertainty estimations \cite{snoek2015scalable,watson2021latent}. Similar to the BBB last layer (BBB--LL) approach, Variational Bayes Last Layer (VBLL) \citep{harrison2024variational} adopts the principles of VI by applying them only to the last layer of a DNN. However, VBLL avoids the use of stochastic Monte Carlo sampling and relies on an approximate analytic expression for the ELBO of the marginal likelihood, which enables deterministic computation of both the variational objective and its gradients. As a result, VBLL preserves the ability to quantify epistemic uncertainty through last layer parameter distributions, while achieving computational efficiency comparable to regular DNNs.}

\textcolor{black}{Although ensemble or VI approaches can offer practical ways to estimate uncertainty, Markov Chain Monte Carlo (MCMC) methods have the advantage of being nonparametric and asymptotically exact when approximating the posterior distribution. The core idea behind MCMC is to use a Markov chain to explore the parameter space by iteratively applying transition steps that move points through the parameter space. Starting from an initial random state sampled from some distribution, the chain evolves through stochastic transitions, producing a sequence of samples (the DNN parameters $\theta$). After discarding an initial burn-in period to mitigate dependence on the starting state, sufficiently many subsequent samples enable the chain to effectively concentrate in high-probability regions and explore the posterior landscape \cite{salimans2015markov,betancourt2017conceptual,van2025mcmc}.}

\textcolor{black}{The Metropolis-Hastings (MH) \cite{metropolis1953equation,hastings1970monte} algorithm is a widely used MCMC method for generating samples from a posterior distribution. It works by proposing a new candidate point in the parameter space, typically via a random walk, and then deciding whether to accept or reject this proposal. If the proposed point has a  higher posterior probability (higher density) than the current set of DNN parameters, it is accepted; if it has a lower density, it is accepted with a probability proportional to the ratio of their densities. This acceptance rule ensures the Markov chain eventually converges to the target posterior distribution. A shortcoming of MH is that it can be inefficient for high-dimensional or complex posteriors because many proposals are rejected, which can lead to slow convergence.}

\textcolor{black}{Hamiltonian Monte Carlo (HMC) leverages gradient information from the log-posterior distribution to efficiently propose new states (DNN parameters in our scenario), increasing sampling efficiency in high-dimensional parameter spaces.  HMC draws inspiration from Hamiltonian dynamics by introducing an auxiliary momentum variable $m$ for the model parameters $\theta$. The total energy of this fictitious physical system is described by the Hamiltonian function (Equation \ref{eq:hamilt}), which decomposes into  the potential energy $U(\theta)$ and the kinetic energy $K(m)$. The potential energy $U(\theta)$ corresponds to the negative log-posterior and quantifies how unlikely a particular parameter value is particular parameter value is given the observed data. The kinetic energy $K(m)$ depends on the momentum variable and governs the sampler’s movement through parameter space. By assigning momentum to the parameters, HMC can use gradients of the posterior to guide the sampler’s trajectory, allowing it to move smoothly and efficiently through the energy landscape rather than performing slow, random-walk steps. The parameter values evolve according to the momentum, which itself is updated based on the  ``force" derived from the shape of the posterior distribution. The exact movement through the parameter space can be computed over the partial derivatives over the Hamiltonian function to determine how both parameter and momentum values evolve over time (Equation \ref{eq:dif1} and \ref{eq:dif2}).}  

\color{black}

\begin{equation}\label{eq:hamilt}
    H(\theta, m) = U(\theta) + K(m)
\end{equation}

where:
\begin{conditions}
    U(\theta) &=& the potential energy, attracts the sampler toward regions of high probability, \\
    K(m) &=& the kinetic energy, determines how fast the sampler moves through the parameter space, and \\
    H(\theta, m) &=& the Hamiltonian (total energy). 
\end{conditions}

\begin{equation}
    \frac{d\theta}{dt} = \frac{\partial H}{\partial m},
      \label{eq:dif1}
\end{equation}

\begin{equation}
    \frac{dm}{dt} = -\frac{\partial H}{\partial \theta},
    \label{eq:dif2}
\end{equation}\\

\color{black}

\textcolor{black}{The differential equations describe how the total energy evolves over time but typically do not have closed-form solutions \cite{van2025mcmc}, meaning the exact evolution of parameters and momentum under Hamiltonian dynamics cannot be computed analytically. To address this, HMC employs a numerical integration method called the leapfrog integrator, which approximates the sampler’s trajectory by discretizing time into small steps $ \varepsilon $ and iteratively updating the parameter and momentum values. After performing $ L $ leapfrog steps, a new proposal $ (\theta', m')$ is obtained. Because the leapfrog integration is an approximation, the total Hamiltonian energy is not perfectly conserved, so the proposed state is accepted or rejected using an acceptance criterion that compares the current and proposed states. The step size $ \varepsilon $  and trajectory length $L$ control the trajectory and hence the sampler’s efficiency. An inappropriate choice of these hyperparameters can cause poor exploration or numerical instability. The No-U-Turn Sampler (NUTS) \cite{hoffman2014no} overcomes the need to manually tune $L$ by adaptively building a binary tree of candidate states and halting the trajectory when it starts to reverse direction, thus maintaining the detailed balance property while enhancing the robustness and efficiency of HMC in complex, high-dimensional posterior distributions.}

\textcolor{black}{Once the chain has converged, the samples produced by the Markov Chain can be stored and used for making predictions on the test data \citep{geyer2011introduction,betancourt2017conceptual,jospin2022hands}. However,  upfront it unknown how many samples are required before reaching convergence. Additionally, performing full-HMC on dataset that for example consist of visual data is computationally expensive. For example, Izmailov et al. \cite{izmailov2021bayesian} selected a random subset of 40,000 images from the CIFAR dataset, and required a compute cluster of 512 TPUs to perform HMC. This is required, because the entire dataset is used to compute to evaluate the current state to the previous. While there have been examples of either using a pre-trained starting position  \cite{benker2021utilizing,vellenga2024pthmc}, chain-splitting \cite{cobb2021scaling}, it remains a challenge to use HMC sampling for larger video datasets.   }

\textcolor{black}{Once the Markov chain has converged, the subsequent samples can be stored and used for predictions on test data \citep{geyer2011introduction,betancourt2017conceptual,jospin2022hands}, where each sample corresponds to a set of DNN parameters. Upfront it is unknown how many samples are required to reach convergence, which depends on the complexity of the posterior landscape. To perform standard full-HMC, it is practically required to have the entire dataset is required in memory during the HMC sampling because each iteration computes gradients over all data points to evaluate the posterior. This makes HMC computationally expensive and memory-intensive for large datasets or complex models. For instance, Izmailov et al. \cite{izmailov2021bayesian} applied HMC to a random subset of 40,000 images from the CIFAR dataset and used a cluster of 512 TPUs to perform sampling. Such resource demands highlight the challenges of scaling full-HMC to larger datasets, models, and problems. While recent approaches employing pre-trained starting points \cite{benker2021utilizing,vellenga2024pthmc} or chain-splitting techniques \cite{cobb2021scaling} can partially reduce the required computational burden, effectively scaling HMC to, for example, video datasets or large-scale deep learning architectures remains an open challenge.}

\textcolor{black}{As an alternative to full MCMC methods, stochastic gradient variants have been developed to address computational challenges. For example, Stochastic Gradient Hamiltonian Monte Carlo (SGHMC) \cite{chen2014stochastic} uses mini-batches to approximate the full gradient, thereby enabling scalable posterior inference. Since mini-batches yield noisy gradient estimates, SGHMC incorporates a friction term that compensates for stochastic fluctuations, stabilizing the sampling dynamics. Similar to full HMC, SGHMC leverages momentum to explore the posterior geometry. Other Stochastic Gradient MCMC methods, such as Stochastic Gradient Langevin Dynamics (SGLD) \cite{welling2011bayesian}, do not include momentum, which tends to make them less effective than SGHMC for Bayesian inference in DNNs \cite{chen2015convergence,li2019stochastic}.}

\subsection{Driver action and intention recognition}
\textcolor{black}{Building upon the PDL methods, driving automation exemplifies an application domain where uncertainty estimations are both critical and challenging.} Given the continued presence of human-driven vehicles, understanding human behavior in traffic scenarios remains essential for driving automation.  \textcolor{black}{Human driving behavior is inherently stochastic \cite{siebinga2023uncovering}, which highlights the importance of advanced driver assistance systems (ADAS) components that can express when the produced predictions are uncertain or identify OOD scenarios. However, practical applications require PDL methods that balance predictive performance, uncertainty reliability, and computational efficiency, given the real-time constraints and hardware limitations of automotive systems.} An example of an ADAS is to assess whether the intended maneuvers of a driver are safe to execute for a particular driving scene. Central components of such an ADAS are driver action and intention recognition. Actions refer to the observable behaviors a driver performs, such as steering, braking, or changing lanes, which can be directly captured through various sensors on a vehicle. Intention recognition focuses on identifying what a driver plans to do in the immediate future, before the actual maneuver takes place \citep{pereira2013state, sadri2011logic}. For example, before changing lanes, a driver might form the intention several seconds before executing any observable actions of that intended driving maneuver. Understanding both actions and intentions is crucial for developing an ADAS that can anticipate and respond to driver behavior \citep{vellenga2022driver}.

Previous studies on driving action recognition (DAR) and driver intention recognition (DIR) have explored various approaches to model the sequential  spatio-temporal relationships captured by various sensors. For example, Yang et al. (2023) \cite{yang2023aide} demonstrated DAR performance improvements when fusing extracted facial landmarks of the ego-vehicle driver and end-to-end representations from three exterior video streams and one in-cabin video streams.  Noguchi and Tanizawa (2023) \cite{noguchi2023ego} constructed spatial-temporal graphs based on exterior-based  object detection and tracking, employing a semi-supervised contrastive learning framework to train a graph convolutional network. For end-to-end DIR, Gebert et al. (2019) \cite{gebert2019end} extracted the optical flow of the in-cabin videos, whereas Rong et al. (2020) \cite{rong2020driver} extracted and predicted the optical flow for the exterior cameras. Vellenga et al. (2024) \cite{vellenga2024evaluation} fine-tuned video masked auto-encoders and attention fusion to improve DAR and DIR performance. Although these DAR and DIR approaches have incrementally improved the recognition performance, they have not considered the uncertainty-based OOD detection, which is a critical aspect for safe driving automation.

\begin{figure}[t]
           \centering
        \includegraphics[trim={0cm 8.5cm 16.6cm 0cm},clip, width=0.70\textwidth, keepaspectratio]{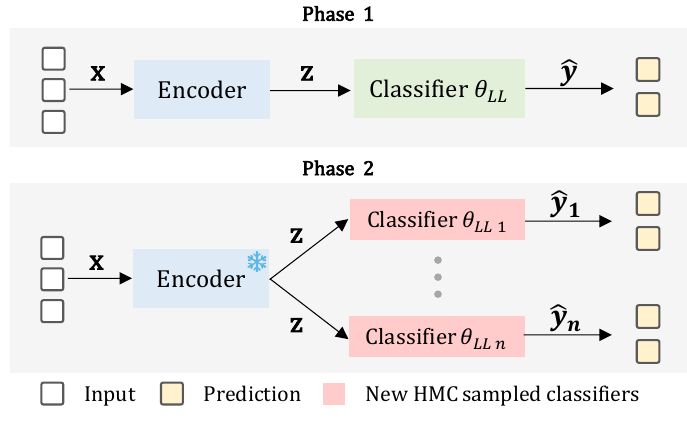}
        \caption{Schematic overview of the two last layer HMC phases. Phase 1 covers the regular optimization-based  training phase. During phase 2, the frozen pre-trained encoder is responsible for extracting the latent representations $z$. Then, the HMC is applied to sample the parameters of the last layer $\theta_{LL}$.}\label{fig:concept}
    \end{figure}


\begin{algorithm}[t]
\color{black}
\caption{Last Layer Hamiltonian Monte Carlo}\label{algo:llhmc}
\begin{flushleft} \small{\textbf{Hyperparameters --} $B$: Burn-in period per chain, $E$: Number of epochs, $C$: Number of chains, $S$: Total number of $\theta_{LL}$ samples
}
\begin{flushleft} \small{\textbf{Parameters -- }$\theta$: Network parameters, $P$: Parameter storage matrix  \\}
\end{flushleft}

\end{flushleft}
\begin{algorithmic}

\State Randomly initialize network parameters $\theta$
\State \textbf{Phase 1 -- Optimization-based pre-training:}
\For{$e$ $\leftarrow 1,\ldots, E$}
\State Update $\theta^{(e)}$ using a stochastic optimizer (e.g., SGD/ADAM)
\EndFor
\State Remove last layer ($\theta_{LL}$) to obtain latent features Z from input data
\State \textbf{Phase 2 -- HMC Sampling for $\theta_{LL}$:}
\State Compute number of samples per chain: $S_c = S / C$ \Comment{Total samples divided evenly across chains}
\State Initialize the last layer's parameters $\theta_{\text{LL}}$
\For{$c$ $\leftarrow 1,\ldots, C$}
\For{$s$ $\leftarrow 1,\ldots, S_c$}
\State Perform HMC to generate a new sample $\theta_{\text{LL}}'$
\If{$s > B$} \Comment{Discard burn-in samples per chain}
\State Add $\theta_{\text{LL}}'$ to $P \lbrack c,s \rbrack $
\EndIf
\EndFor
\EndFor
\end{algorithmic}
\end{algorithm}
\color{black}

\section{Last layer Hamiltonian Monte Carlo}

\textcolor{black}{To mitigate the computational demands of standard full HMC (e.g., \cite{izmailov2021bayesian}) and the limitations of mini-batch SGMCMC approaches, we propose last-layer Hamiltonian Monte Carlo (LL--HMC). LL--HMC confines the HMC sampling procedure to the final layer of the network, making it feasible for data-intensive applications or for larger pre-trained models. This method consists of two phases: regular optimization-based training and full HMC sampling of the last layer parameters $\theta_{LL}$ (see \cref{algo:llhmc} and \cref{fig:concept}). In Phase 1, the DNN is trained or fine-tuned with standard optimizers such as SGD or ADAM to update all parameters $\theta$. Subsequently, the last layer $\theta_{LL}$ is replaced, and latent representations $\mathbf{z}$ from the penultimate layer are extracted for training, test, and OOD datasets.}


\textcolor{black}{During phase 2, we estimate the posterior over $\theta_{LL}$ conditioned on $\mathbf{z}$ rather than the raw observed data using HMC. Similar to full HMC with a NUTS sampler, we must tune several hyperparameters, which include the number of burn-in samples, number of to be collected  $\theta_{LL}$ parameter samples, the scale of the prior distribution over initialized parameters, the target acceptance probability that influences step size adaptation, and the number of chains used during sampling. The number of leapfrog steps steps is determined adaptively by NUTS \citep{hoffman2014no}.}

\textcolor{black}{The sampling procedure begins by initializing $\theta_{LL}$ values according to the chosen prior standard deviation scale. For each chain $C$, HMC explores the posterior given $\mathbf{z}$, treating $\theta_{LL}$ as positions in a dynamical system. The leapfrog integrator proposes new parameter states $\theta'_{LL}$ by simulating Hamiltonian dynamics, while the NUTS sampler adjusts leapfrog steps to efficiently traverse the parameter space. The  acceptance ratio between current $(\theta_{LL}, \mathbf{p})$ and proposed $(\theta'_{LL}, \mathbf{p}')$ states determines acceptance of proposals.}

\textcolor{black}{From a computational complexity perspective, the efficiency gain of LL-–HMC is comparable to that of using a sub-ensemble compared to a regular ensemble. In contrast to full HMC, which produces multiple samples of all network parameters and entails substantial computational overheads \cite{izmailov2021bayesian}, LL–-HMC only samples the parameters of the final layer. For our driving automation example, which processes video inputs with a ViT-Base backbone, every additional full-network parameter sample requires $361 \times 10^9$ FLOPs, whereas each additional last-layer parameter sample adds only $7.68 \times 10^3$ FLOPs for a five-class scenario. Hence, the computational cost of additional last-layer parameter samples is negligible compared to the number of operations required to produce the latent feature representations or when leveraging multiple full-network parameter samples.}

\begin{figure}[t!]
      \begin{subfigure}[b]{0.33\linewidth}
        \includegraphics[width=\linewidth, trim=1.7cm 1.05cm 0.95cm 1.35cm, clip]{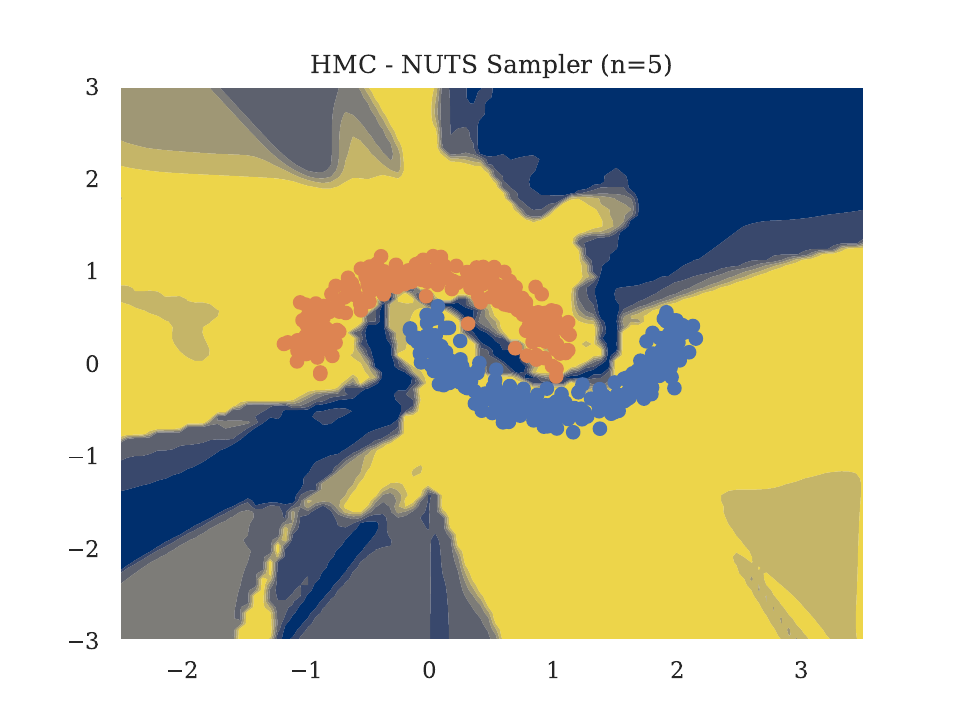} 
       \captionsetup{font=tiny}
        \caption{HMC NUTS (N=5)}
    \end{subfigure}
    \hfill
      \begin{subfigure}[b]{0.33\linewidth}
        \centering
        \includegraphics[width=\linewidth, trim=1.7cm 1.05cm 0.95cm 1.35cm, clip]{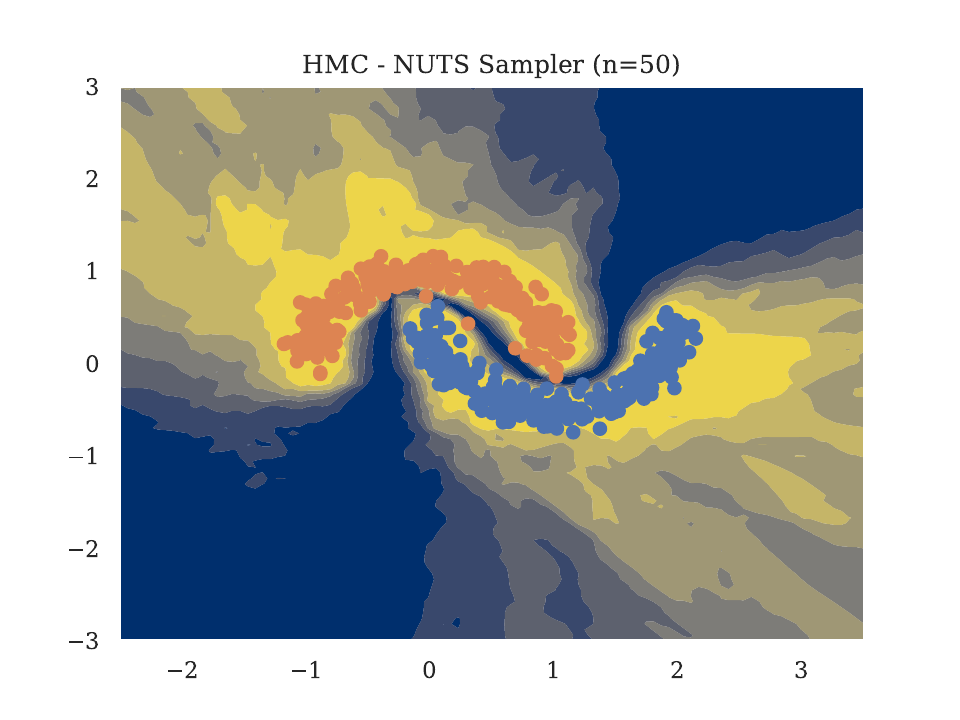} %
       \captionsetup{font=tiny}
        \caption{HMC NUTS (N=50)}
    \end{subfigure}
    \hfill
      \begin{subfigure}[b]{0.33\linewidth}
        \centering
        \includegraphics[width=\linewidth, trim=1.7cm 1.05cm 0.95cm 1.35cm, clip]{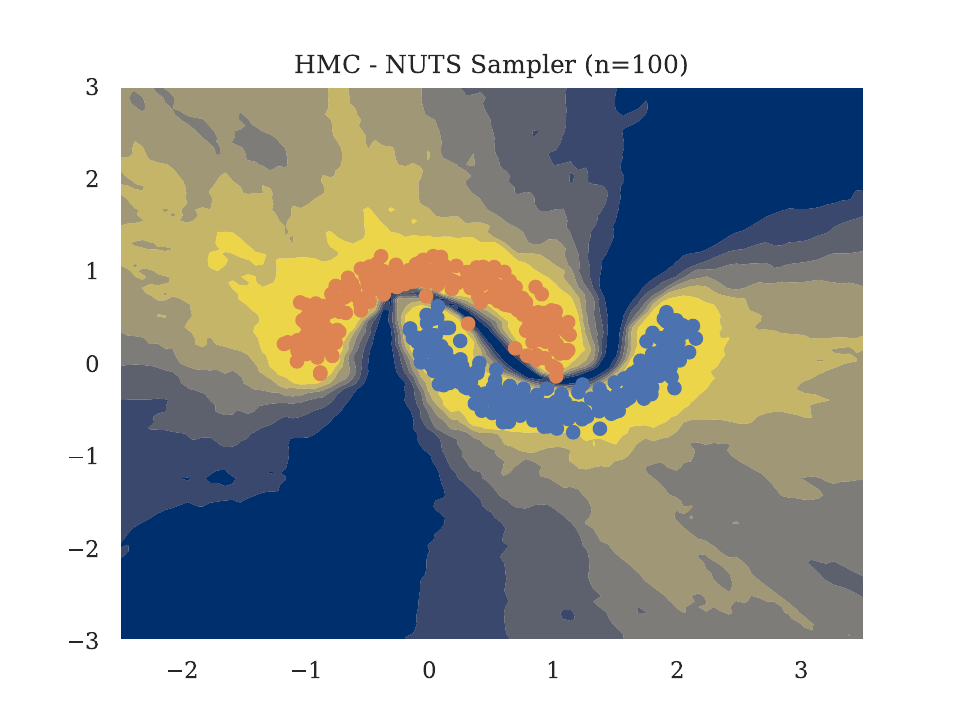} %
       \captionsetup{font=tiny}
        \caption{HMC NUTS (N=100)}
    \end{subfigure}
    
    \vskip\baselineskip 
    
      \begin{subfigure}[b]{0.33\linewidth}
        \centering
        \includegraphics[width=\linewidth, trim=1.7cm 1.05cm 0.95cm 1.35cm, clip]{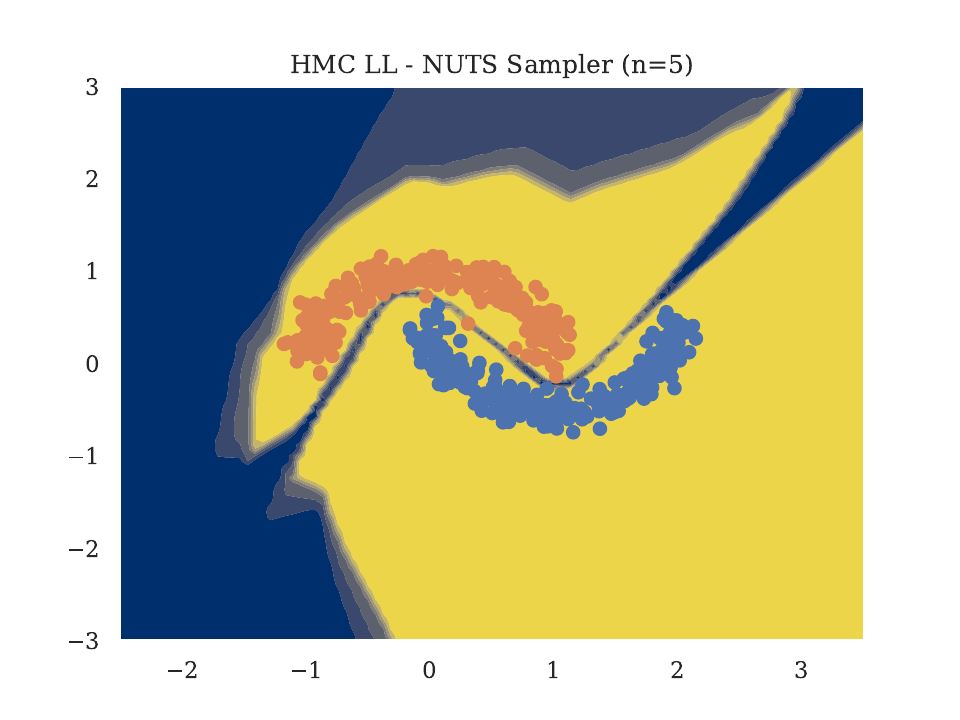} %
       \captionsetup{font=tiny}
        \caption{LL--HMC NUTS (N=5)}
    \end{subfigure}
    \hfill
      \begin{subfigure}[b]{0.33\linewidth}
        \centering
        \includegraphics[width=\linewidth, trim=1.7cm 1.05cm 0.95cm 1.35cm, clip]{toy_class/hmc_ll_50.pdf} 
       \captionsetup{font=tiny}
        \caption{LL--HMC NUTS (N=50)}
    \end{subfigure}
    \hfill
      \begin{subfigure}[b]{0.33\linewidth}
        \centering
        \includegraphics[width=\linewidth, trim=1.7cm 1.05cm 0.95cm 1.35cm, clip]{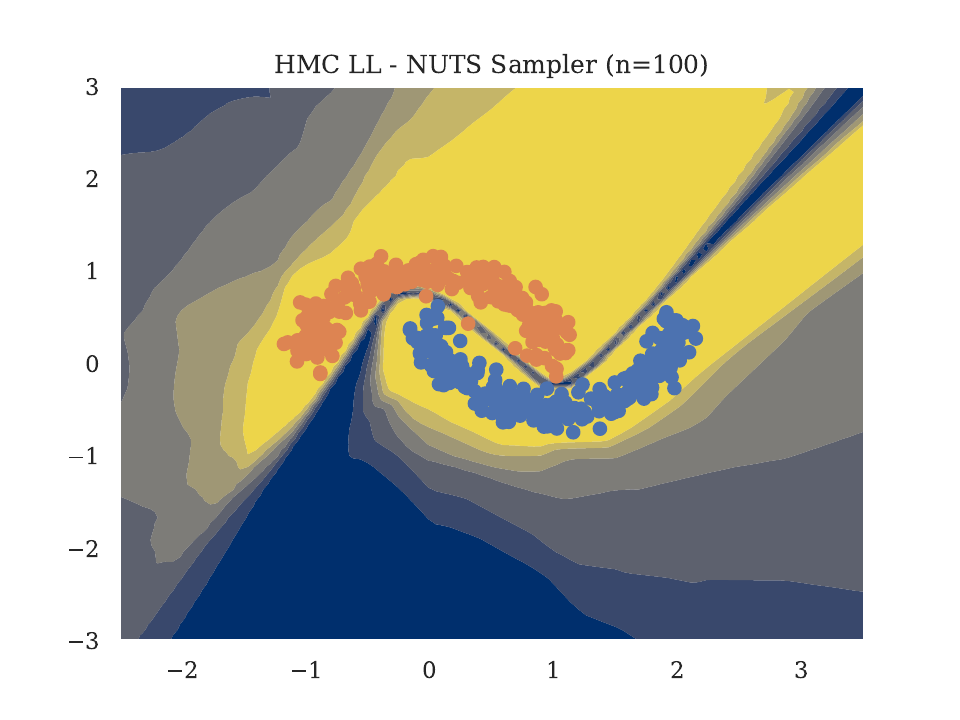} 
       \captionsetup{font=tiny}
        \caption{LL--HMC NUTS (N=100)}
    \end{subfigure}
    
    \vskip\baselineskip
    
      \begin{subfigure}[b]{0.33\linewidth}
        \centering
        \includegraphics[width=\linewidth, trim=1.7cm 1.05cm 0.95cm 1.35cm, clip]{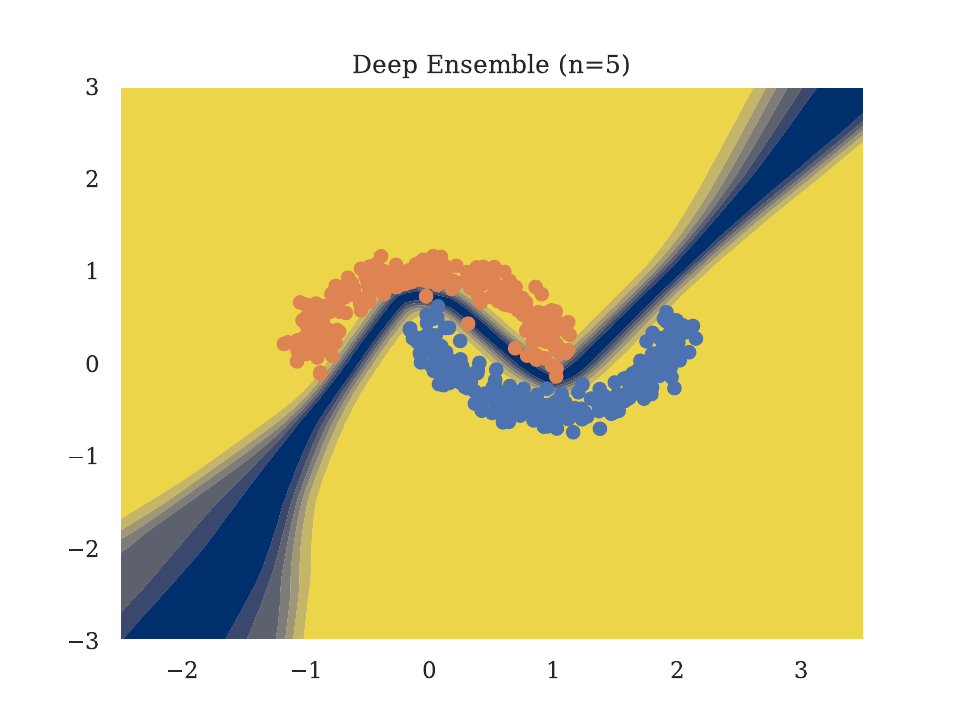} 
       \captionsetup{font=tiny}
        \caption{DE (N=5)}
    \end{subfigure}
    \hfill
      \begin{subfigure}[b]{0.33\linewidth}
        \centering
        \includegraphics[width=\linewidth, trim=1.7cm 1.05cm 0.95cm 1.35cm, clip]{toy_class/ensemble_50.pdf} 
       \captionsetup{font=tiny}
        \caption{DE (N=50)}
    \end{subfigure}
    \hfill
      \begin{subfigure}[b]{0.33\linewidth}
        \centering
        \includegraphics[width=\linewidth, trim=1.7cm 1.05cm 0.95cm 1.35cm, clip]{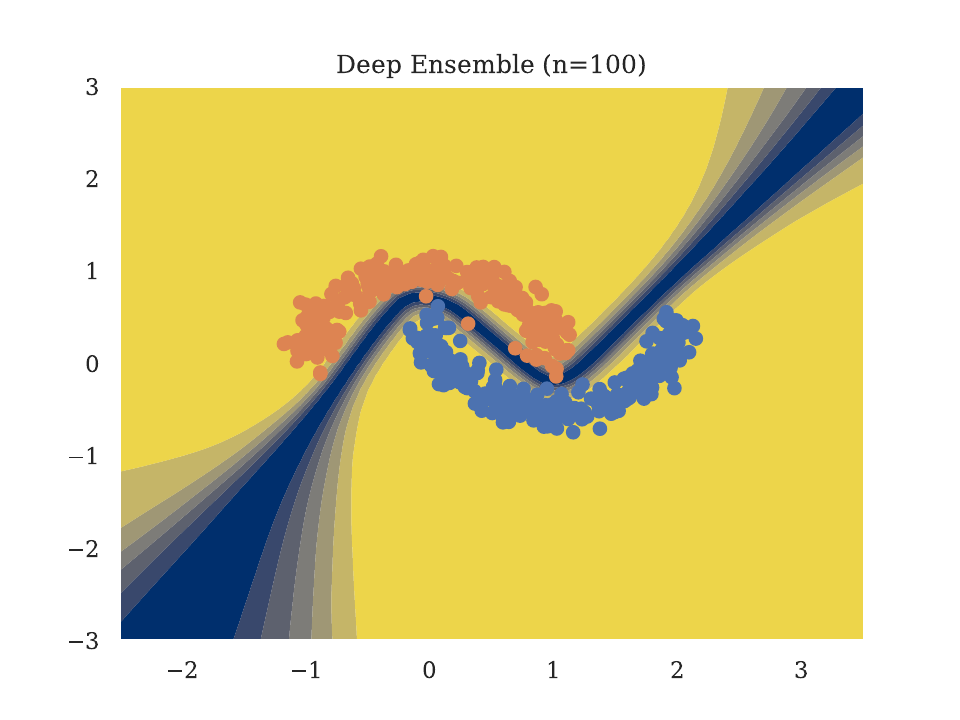} 
        \captionsetup{font=tiny}
        \caption{DE (N=100)}
    \end{subfigure}

    \caption{Toy two moons classification example results. First row consists of full HMC with 5, 50, and 100 samples, the second row shows the LL--HMC results, and the third row shows the results for a deep ensemble. The yellow area denotes certain predictions, whereas the dark blue area indicates uncertain predictions. The certain (yellow) vs uncertain (dark blue) color map is normalized.}
    \label{fig:toy_class}
\end{figure}

\begin{figure}[t!]
    \centering 
       \begin{subfigure}[b]{0.33\linewidth}
        \centering
        \includegraphics[width=\linewidth, trim=2.85cm 1.2cm 1.5cm 1.4cm, clip]{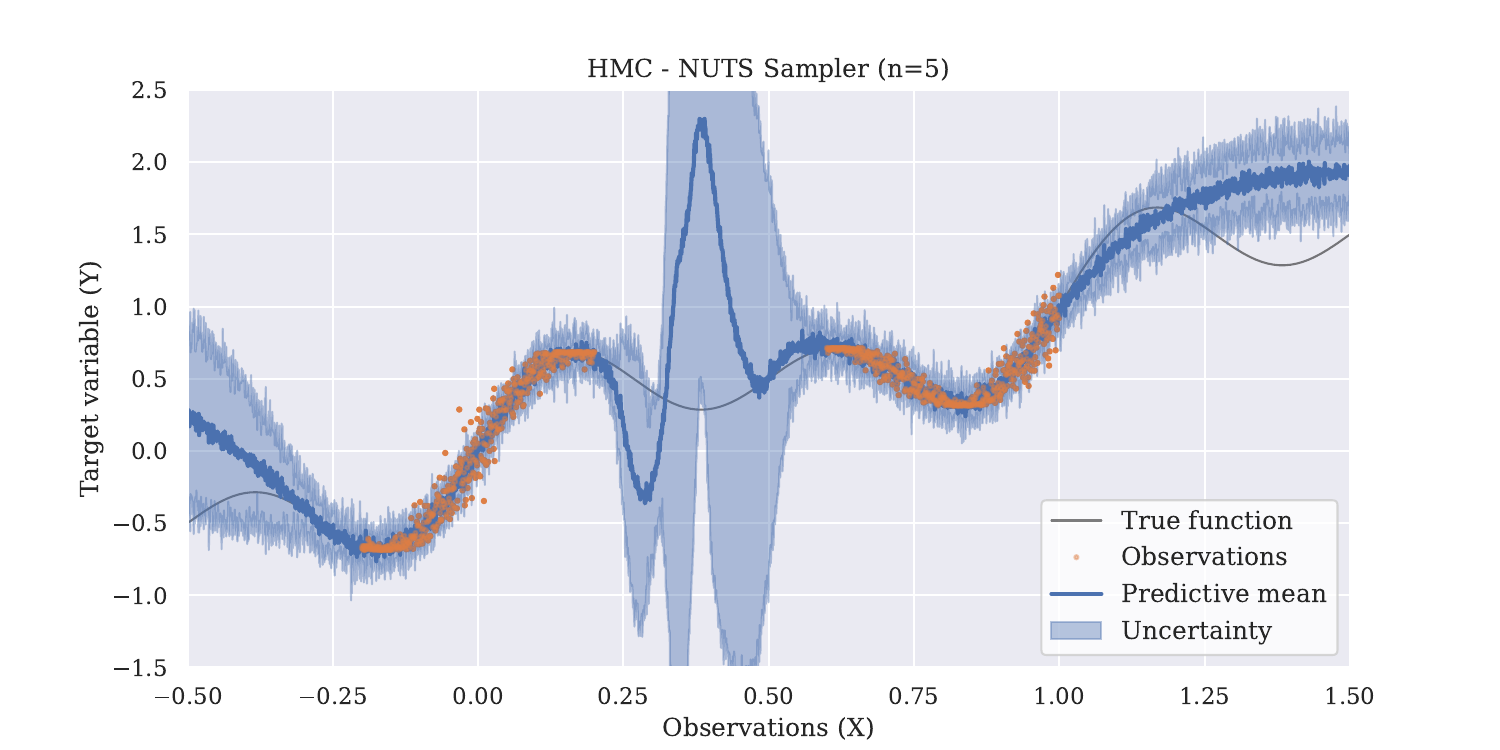} 
       \captionsetup{font=tiny}
             \caption{HMC NUTS (N=5)}
    \end{subfigure}
    \hfill
      \begin{subfigure}[b]{0.33\linewidth}
        \centering
        \includegraphics[width=\linewidth, trim=2.85cm 1.2cm 1.5cm 1.4cm, clip]{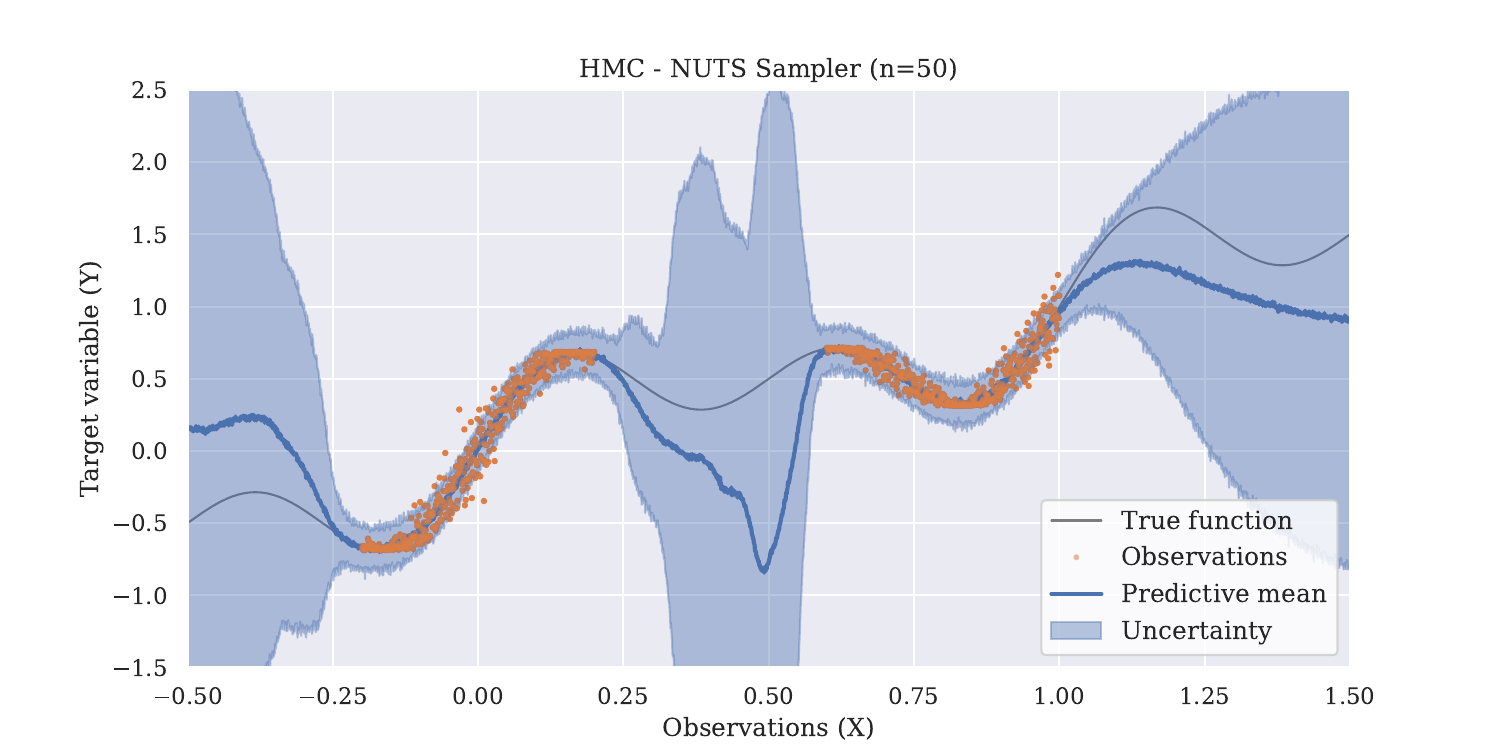} 
           \captionsetup{font=tiny}
             \caption{HMC NUTS (N=50)}
    \end{subfigure}
    \hfill
      \begin{subfigure}[b]{0.33\linewidth}
        \centering
        \includegraphics[width=\linewidth, trim=2.85cm 1.2cm 1.5cm 1.4cm, clip]{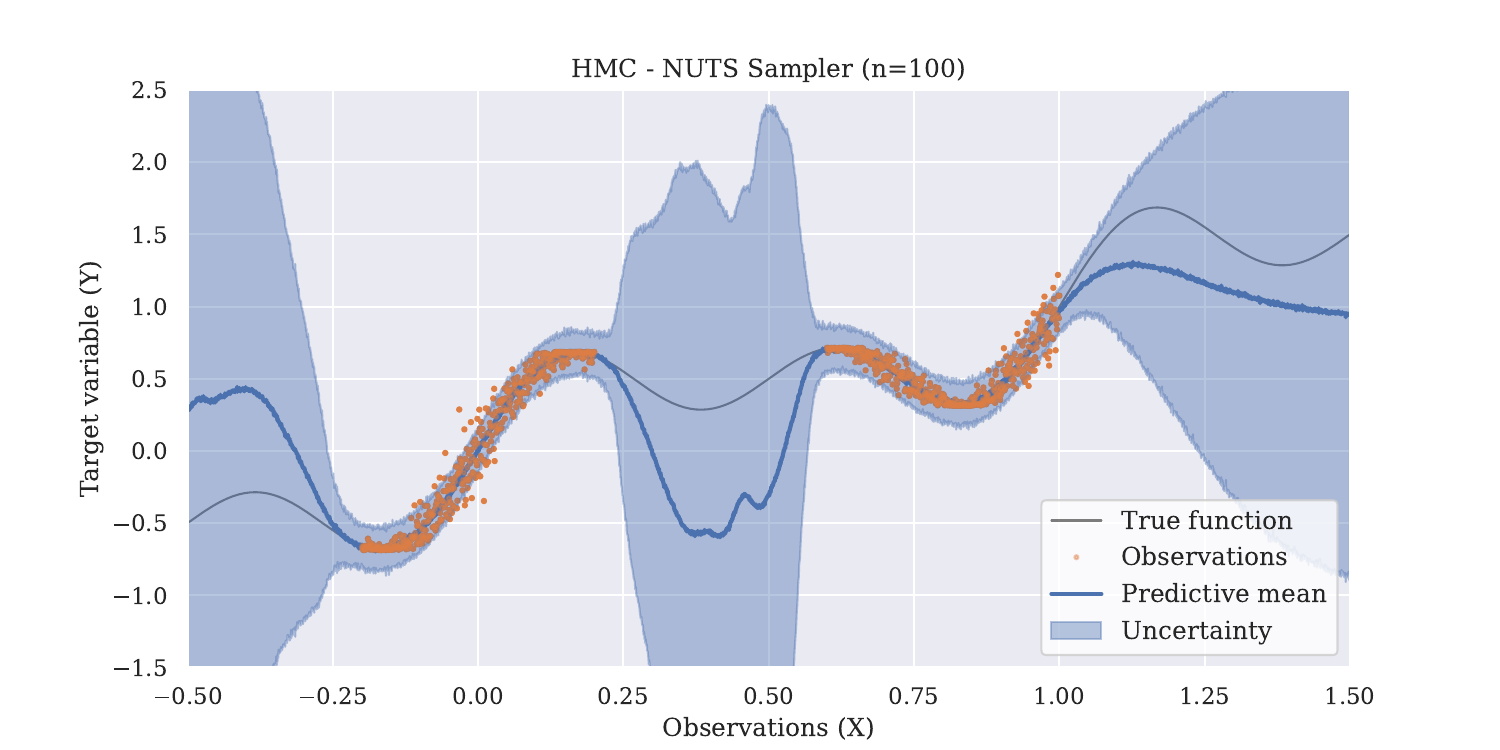} 
       \captionsetup{font=tiny}
             \caption{HMC NUTS (N=100)}
    \end{subfigure}
    
    \vskip\baselineskip 
    
      \begin{subfigure}[b]{0.33\linewidth}
        \centering
        \includegraphics[width=\linewidth, trim=2.85cm 1.2cm 1.5cm 1.4cm, clip]{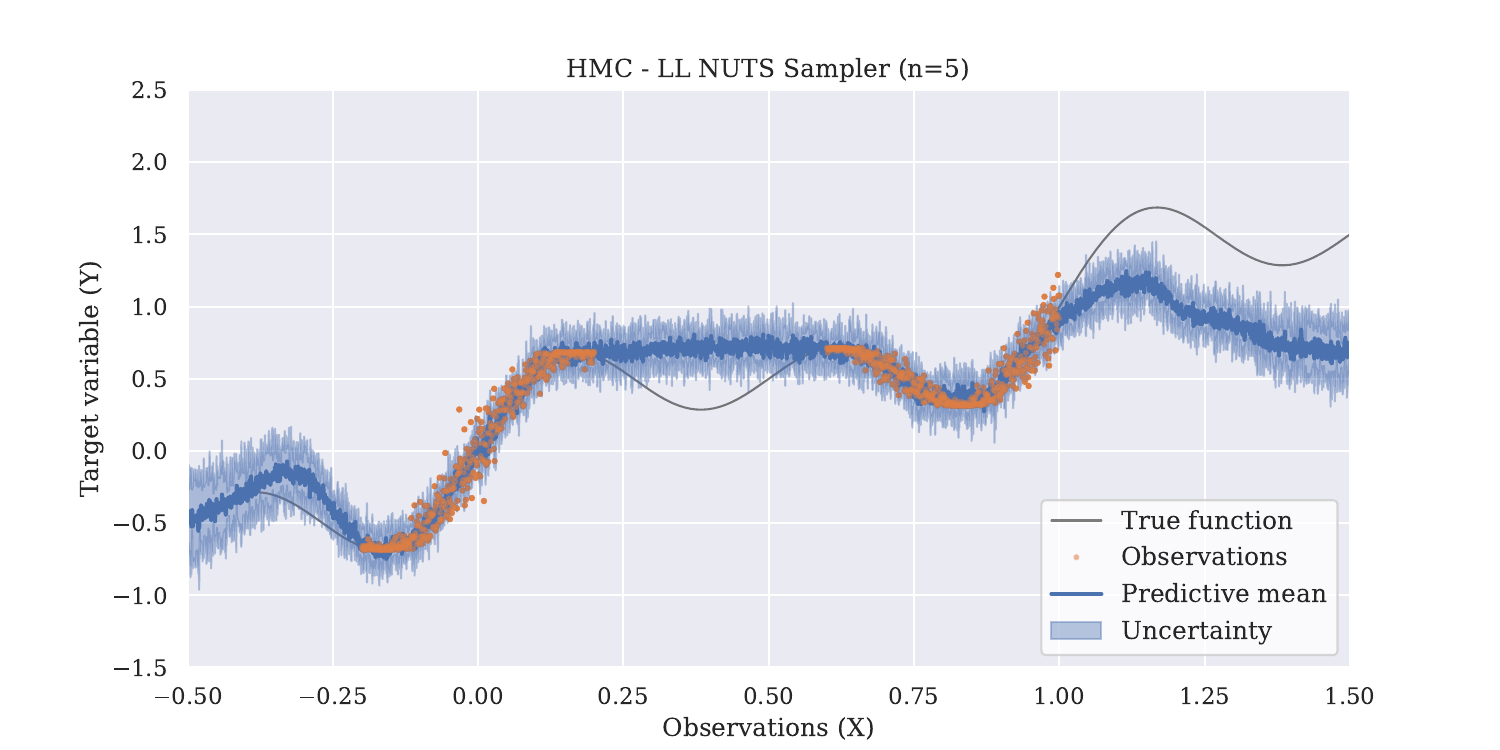} 
        \captionsetup{font=tiny}
         \caption{LL--HMC NUTS (N=5)}
    \end{subfigure}
    \hfill
      \begin{subfigure}[b]{0.33\linewidth}
        \centering
        \includegraphics[width=\linewidth, trim=2.85cm 1.2cm 1.5cm 1.4cm, clip]{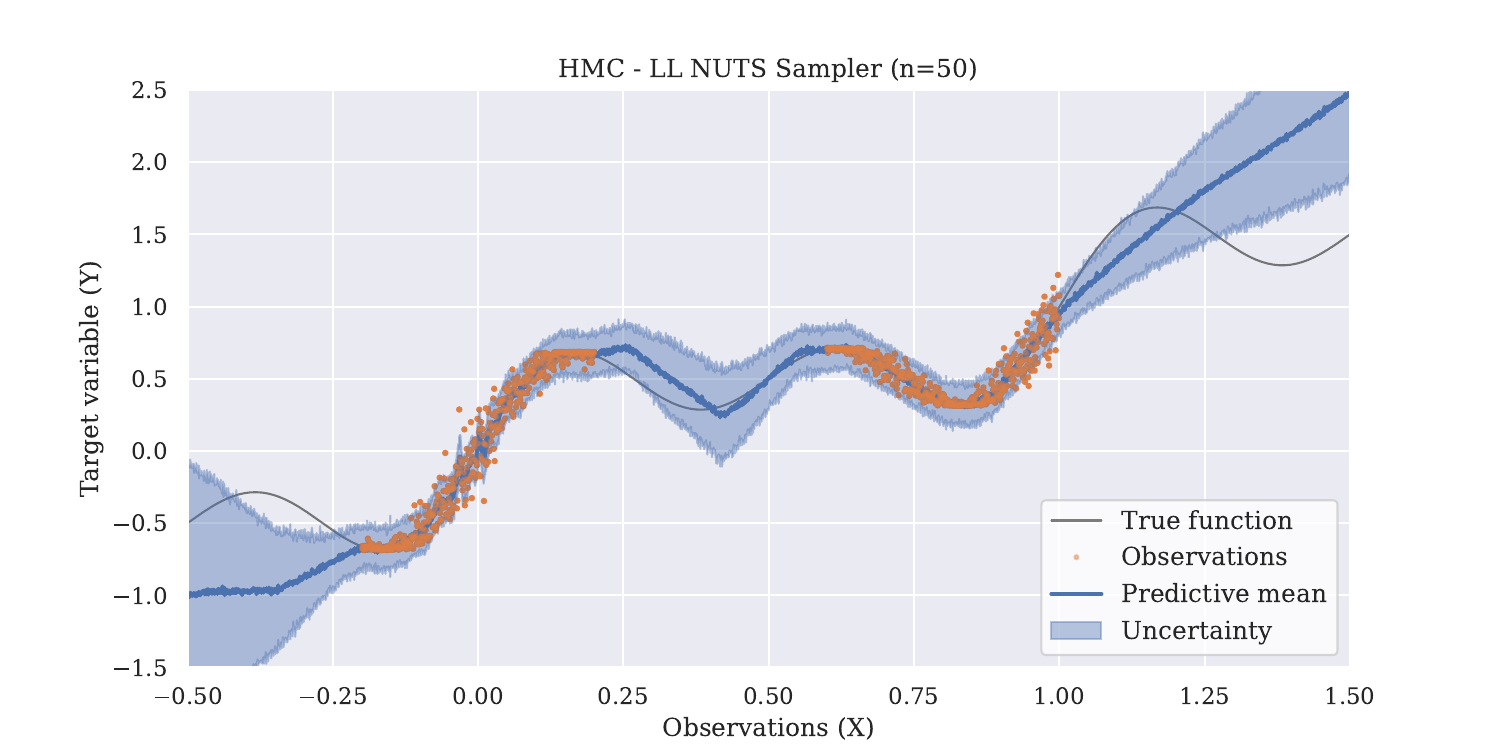} %
       \captionsetup{font=tiny}
         \caption{LL--HMC NUTS (N=50)}
    \end{subfigure}
    \hfill
      \begin{subfigure}[b]{0.33\linewidth}
        \centering
        \includegraphics[width=\linewidth, trim=2.85cm 1.2cm 1.5cm 1.4cm, clip]{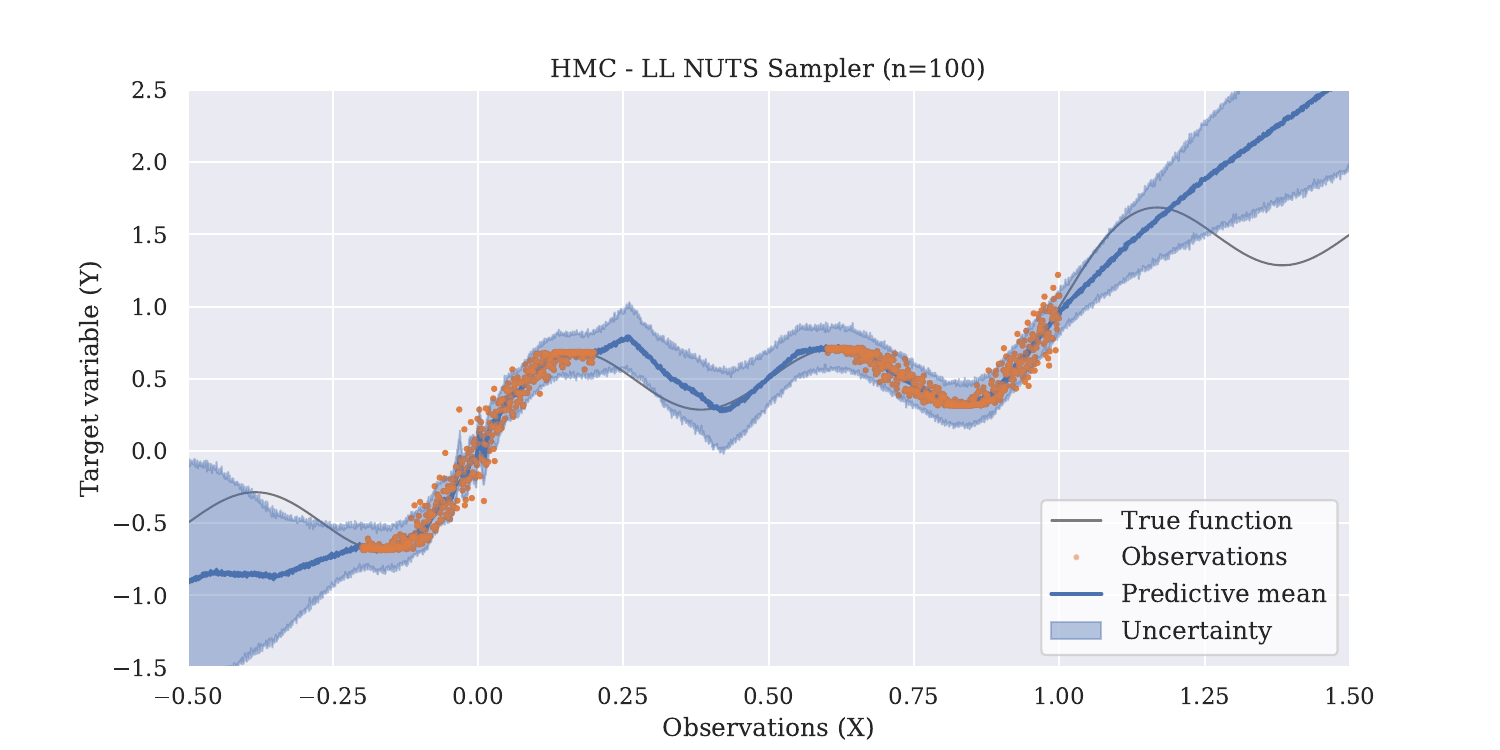} %
       \captionsetup{font=tiny}
         \caption{LL--HMC NUTS (N=100)}
    \end{subfigure}
    
    \vskip\baselineskip
    
      \begin{subfigure}[b]{0.33\linewidth}
        \centering
        \includegraphics[width=\linewidth, trim=2.85cm 1.2cm 1.5cm 1.4cm, clip]{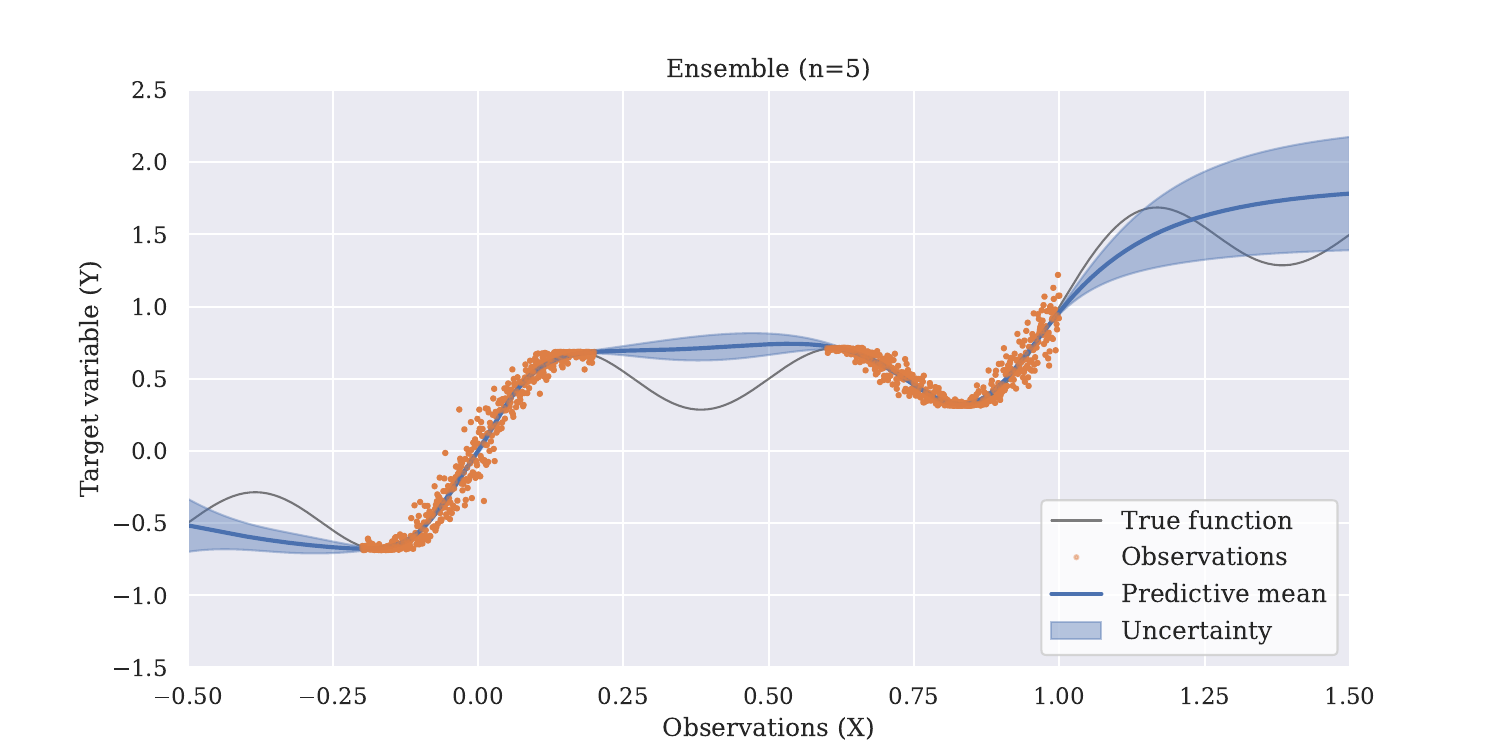} %
       \captionsetup{font=tiny}
         \caption{DE (N=5)}
    \end{subfigure}
    \hfill
      \begin{subfigure}[b]{0.33\linewidth}
        \centering
        \includegraphics[width=\linewidth, trim=2.85cm 1.2cm 1.5cm 1.4cm, clip]{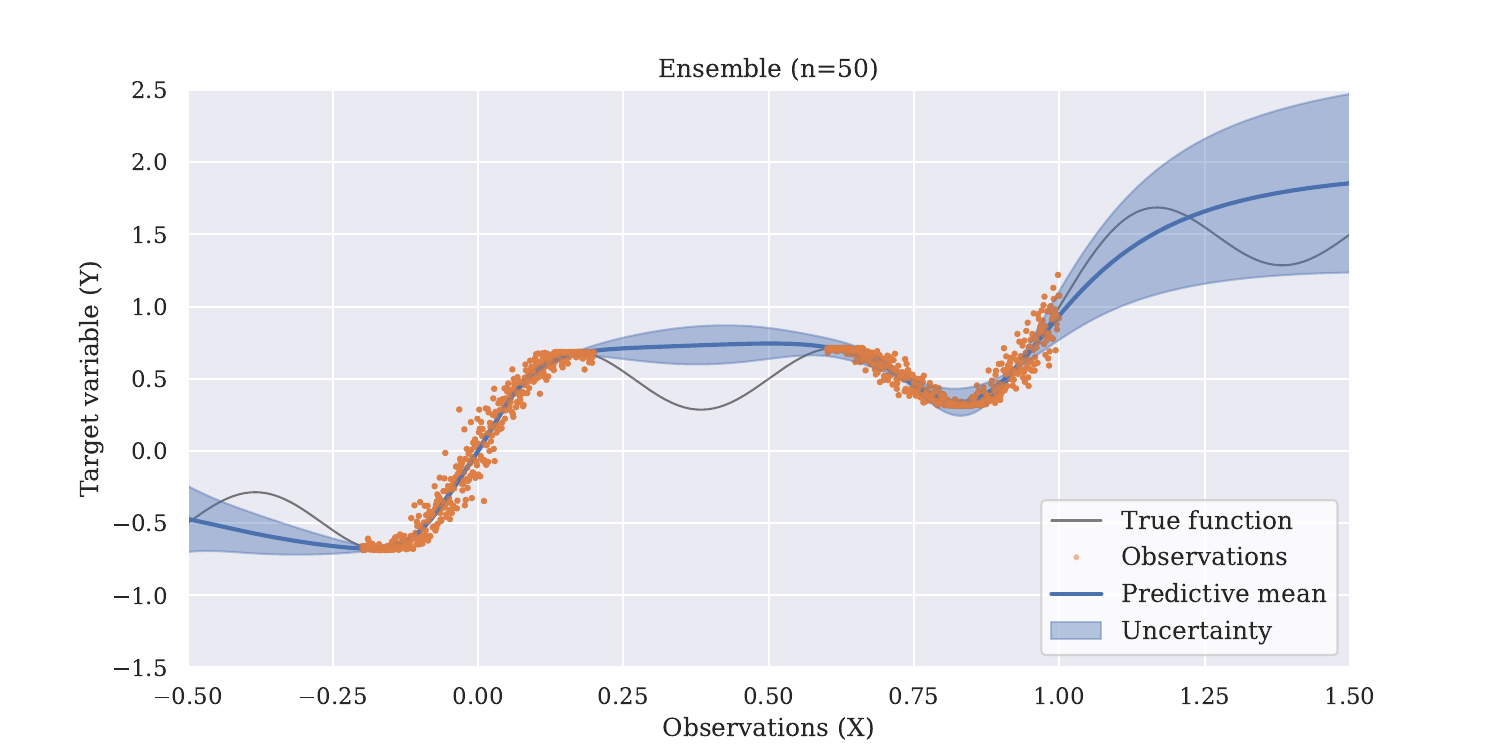} %
       \captionsetup{font=tiny}
         \caption{DE (N=50)}
    \end{subfigure}
    \hfill
      \begin{subfigure}[b]{0.33\linewidth}
        \centering
        \includegraphics[width=\linewidth, trim=2.85cm 1.2cm 1.5cm 1.4cm, clip]{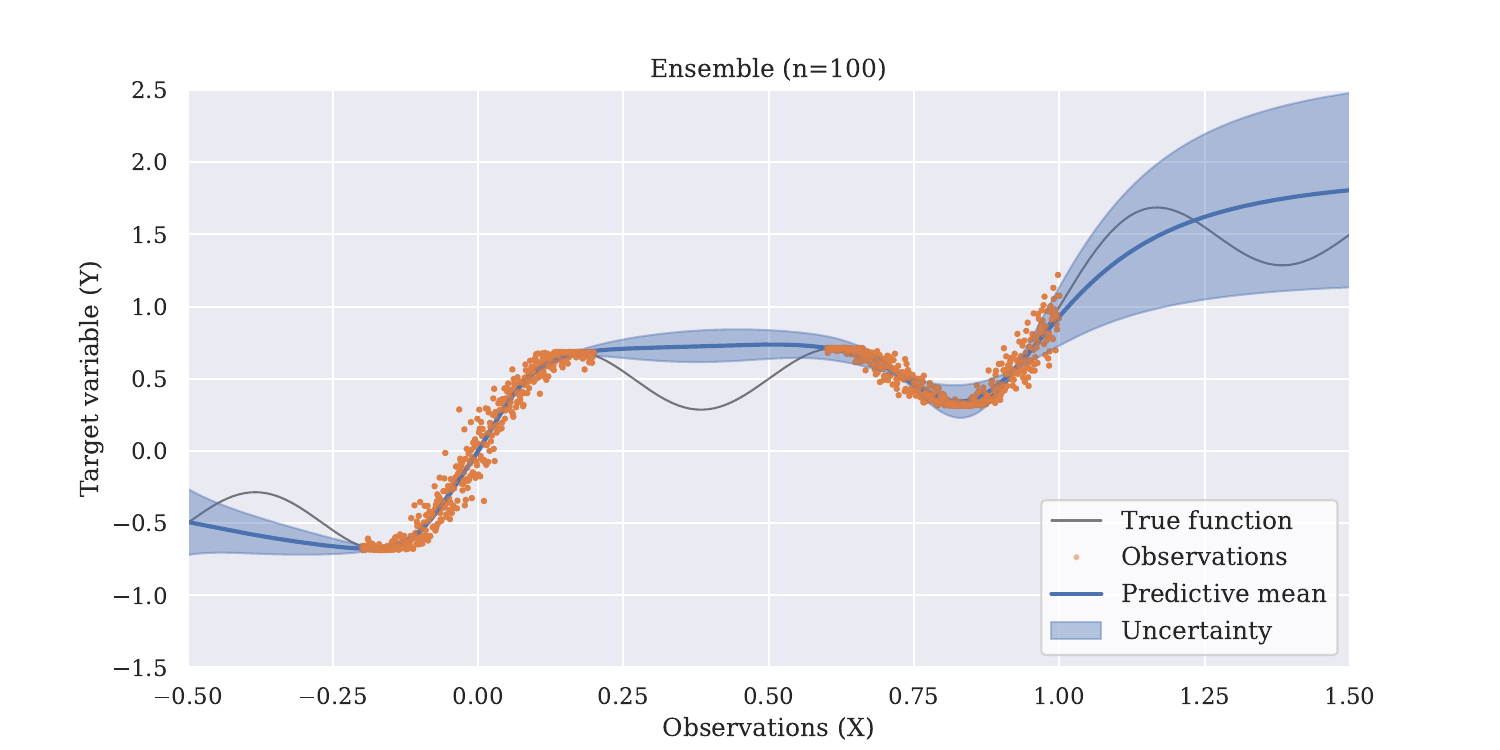} %
       \captionsetup{font=tiny}
         \caption{DE (N=100)}
    \end{subfigure}

    \caption{Toy Regression Example results. First row consists of full HMC with 5, 50, and 100 samples, the second row shows the LL--HMC results, and the third row shows the results for a deep ensemble. The y-axis is the target variable, the x-axis the observations. The gray line is the original function, the orange dots represent the noisy observations, the blue line the mean prediction, and the light blue area the uncertainty estimation.}
      \label{fig:toy_reg}
\end{figure}

\subsection{Toy-examples} \label{sec:3.1}
We visualize the uncertainty estimation capabilities of LL–HMC using a classification and a regression toy example. Specifically, we compare LL--HMC to HMC sampling on the entire network and a DE that uses a similar number of samples. We use the two moons dataset \citep{pedregosa2011scikit} with a multi-layer perceptron with two hidden layers of 20 units, following the implementation from van Amersfoort et al. \cite{van2020uncertainty}. Figure \ref{fig:toy_class} shows where the different methods are certain (yellow) and uncertain (dark blue) for the two moons dataset. With more samples, full HMC gradually becomes more uncertain in areas for which no data has been observed. LL--HMC becomes more uncertain in unobserved areas as the number of samples increases from 5 to 50. However, the uncertainty estimations do not improve further with additional samples, indicating the limitations of the LL--HMC approach. In congruence with van Amersfoort et al. 
\cite{van2020uncertainty}, we observe that the DE mostly remains uncertain along the decision boundary  and only improves slightly with more ensemble members. 

To evaluate the uncertainty estimations for a regression toy example, we use the implementation of Lippe \cite{lippe2022uva} to generate noisy observations from a sinusoidal function. \textcolor{black}{The orange dots in Figure \ref{fig:toy_reg} show the observed data points, whereas the gray line visualizes the original sinusoidal function. The goal for this regression toy example is to produce high uncertainty estimates in regions without data observations (i.e., where no orange dots are present). To achieve this, we use an MLP with five hidden layers and 50 hidden units for all methods as the underlying architecture. In Figure \ref{fig:toy_reg}, the blue line denotes the mean prediction, and the light blue area denotes the uncertainty estimation, where a larger area indicates higher uncertainty. When performing HMC on the entire DNN, we observe higher uncertainty estimations for regions without observations, which is the desired outcome. With an increasing number of samples, the uncertainty estimation becomes more reliable, the predictive mean and the corresponding uncertainty estimation area appear smoother and less noisy. The apparent increase in the width of the uncertainty area, particularly in regions without observed data, does not indicate greater overall uncertainty but rather reflects improved estimation of uncertainty due to the additional network parameter samples or ensemble members. For both LL–HMC and deep ensembles, the uncertainty estimates for the area in between the observed data points used for training, remains relatively stable. This highlights that applying full HMC on the entire network results in better uncertainty estimates for the areas without observed data points.} 



\section{Experimental setup}

\subsection{Datasets}
For the DAR task, we use the AssIstive Driving pErception (AIDE) dataset \citep{yang2023aide} that consists of 2898 short videos recorded in China, with the following labels: 185 \textit{ backward moving}, 1522 \textit{forward moving}, 138 \textit{lane changing}, 727 \textit{parking}, and 326 \textit{turning}. Additionally, we use annotations from the ROad event Awareness Dataset (ROAD) \citep{singh2022road}, which extend a subset of 22 videos (122.000 frames) of the Oxford RobotCar Dataset \citep{RobotCarDatasetIJRR} with ego-vehicle and road user driving action labels. Since the ROAD data was collected with manual driving, the vehicle action labels reflect human driving maneuvers \citep{RobotCarDatasetIJRR}. To capture temporal patterns, we transform the frame-level annotations into video-level labels by selecting only driving actions with at least 8 consecutive frames. This threshold ensures that each clip provides sufficient temporal context for distinguishing maneuvers while reducing ambiguous or noisy labels.  As a result, the offline ROAD DAR dataset consists of 75 \textit{stopping},  162 \textit{move forward}, 31 \textit{turn right}, {41} \textit{turn left}, 12 \textit{move right}, 12 \textit{move left} and 8 \textit{overtake} maneuvers. 

For DIR, we use the open-source Brain4Cars (B4C) dataset \citep{jain2015car} collected in the United States. The B4C dataset consists of 124 \textit{left lane changes}, 123 \textit{right lane changes}, 58 \textit{left turns}, 55 \textit{right turns}, and 234 \textit{driving straight} maneuvers. We retain the natural class imbalance present in all datasets to reflect the real-world distribution of driving behaviors. For all datasets, we use the original train-test splits. For the ROAD and B4C datasets, we use the first fold from the provided splits.


\subsection{Metrics}
\paragraph{\textbf{Performance, and uncertainty.}} 

\textcolor{black}{To assess in-distribution classification performance, the accuracy ($\uparrow$, see Equation \ref{eq:ac}) measures the proportion of correct predictions among all predictions, offering an overall indication of the model’s effectiveness across classes. Given the presence of class imbalance in the datasets, we also report the F1 score ($\uparrow$, Equation \ref{eq:f1}). The F1 score is the harmonic mean of precision (the proportion of instances correctly assigned to a class among all predictions for that class) and recall (the proportion of instances of a given class that the model correctly identifies). This balanced view of precision and recall is important in applications where both missed and false detections may lead to significant consequences.}



\color{black}
\begin{equation}
   Accuracy = \frac{TP+TN}{TP+TN+FP+FN},
    \label{eq:ac}
\end{equation}

\begin{equation*}
   Precision = \frac{TP}{TP+FP},
   \label{eq:prec}
\end{equation*} 

\begin{equation*}
   Recall = \frac{TP}{TP+FN},
   \label{eq:rec}
\end{equation*}

\begin{equation}
   F_{1} = 2 \cdot  \frac{Precision \cdot Recall}{Precision+Recall},
   \label{eq:f1}
\end{equation}

where:
\begin{conditions} 
   TP & = &  True Positives, instances correctly classified as belonging to a specific class, \\
   TN & = &  True Negatives, instances correctly classified as not belonging to a class, \\
   FP & = &  False Positives, instances incorrectly classified as belonging to the class (false alarms), and\\
   FN & = & False Negatives, instances belonging to the class but incorrectly classified as not  belonging to it (misses).\\
\end{conditions}

\color{black}

To understand the differences between the PDL methods, we use two times the standard error of the mean (2SEM \cite{barde2012use},  Equation \ref{eq:2sem}) instead of the standard deviation (the variability across random seeds). 2SEM roughly corresponds to the 95\% confidence interval of the mean performance of each PDL method. To estimate uncertainty, we use predictive entropy \citep{smith2018understanding} (Equation \ref{eq:uc_it}), which is maximized when the probabilities are uniformly distributed among all classes. The more confident the model is in its predictions for a particular class, the lower the predictive entropy will be, indicating less uncertainty.   

\begin{equation}
2SEM = 2 \frac{\sigma}{\sqrt{n}},
\label{eq:2sem}
\end{equation}

where:
\begin{conditions}
\sigma & = & standard deviation, and \\
n &=& number of observations.
\end{conditions}

\begin{equation}
H[P(y|x)] = - \sum_{y \in \mathcal{Y}} P(y|x) \log P(y|x),
\label{eq:uc_it}
\end{equation}

where:
\begin{conditions} 
H[P(y|x)]  & = & the entropy of the predictive distribution, and \\ 
P(y|x) & = & conditional probability distribution, over some discrete set of outcomes $\mathcal{Y}$.\\
\end{conditions}

\paragraph{\textbf{Calibration.}} \textcolor{black}{Calibration metrics evaluate how well the predicted probabilistic or uncertainty estimates correspond to the true likelihood of outcomes. For example, if a model predicts a class with 70\% confidence, that class should be correct in roughly 70\% of such predictions. A model is considered overconfident when it assigns high probabilities to incorrect outcomes and conservative when it assigns low probabilities to correct outcomes. } \textcolor{black}{The expected calibration error (ECE) \cite{naeini2015obtaining} partitions predictions into a fixed number of bins, grouping samples based on their confidence values. Each bin’s average predicted probability is then compared to the empirical fraction of correct predictions within that bin. However, the fixed binning approach can lead to unequal sample counts across bins, causing poorly populated intervals to dominate the error estimation. To mitigate unequal bins, the adaptive calibration error (ACE  $\downarrow$, \cite{nixon2019measuring}, Equation \ref{eq:ACE}) replaces the fixed binning with adaptive bin intervals, ensuring that each bin contains a similar number of instances. This yields a more balanced assessment of confidence–accuracy alignment by reducing the bias introduced by unevenly distributed predictions. In our evaluation, we compute the ACE over an adaptive range of 10 bins, similar to \cite{vellenga2024pthmc}. } To assess the calibration of the uncertainty estimates relative to the model performance, we use the relative Area Under the Lifted Curve (rAULC $\uparrow$, \cite{postels2022practicality}, Equation \ref{eq:raulc}). The rAULC evaluates how well uncertainty scores are calibrated. Ideally, confident and correct predictions should correspond to low uncertainty, while incorrect or low-confidence predictions should yield high uncertainty.


\begin{equation}
ACE =  \frac{1}{KR} \sum_{k=1}^{K} \sum_{r=1}^{R} | acc(r,k) - conf(r,k) |,
\label{eq:ACE}
\end{equation}
where:
\begin{conditions}
  k & = & class label,\\
  r & = & adaptive range for class label,\\
  acc(r,k) & = & accuracy within a bin, and\\
  conf(r,k) & = & predicted confidence within a bin.\\
\end{conditions}

\begin{equation}
     \text{rAULC} = -1 + \sum_{i=1}^{S} \frac{s \cdot A(q_i)}{R(q_i)},
     \label{eq:raulc}
\end{equation}

where:
\begin{conditions}
q_i & = & uncertainty quantiles $i\in [1, ..., S]$, \\
s & = &  quantile step width based on the number of predictions in  the test dataset ($1/{N}$), \\
A(q_i) & = & accuracy of samples with uncertainty below quantile $q_i$, \\
R(q_i) & = & expected accuracy for random guessing at quantile $q_i$, and \\
S & = & total number of quantiles. \\
\end{conditions}

\paragraph{\textbf{MCMC diagnostics.}}We evaluate the quality of posterior samples using the Effective Sample Size (ESS, \cite{geyer1992practical}, Equation \ref{eq:ess}). A higher ESS implies better exploration of the posterior distribution. For LL--HMC with multiple chains, we use Gelman’s $\hat{R}$ statistic \citep{gelman1992inference} to assess convergence (Equation \ref{eq:rhat}). $\hat{R}$ compares the variance between chains to the variance within each chain. Values close to 1 suggest that the chains are exploring the same distribution \citep{izmailov2021bayesian}.

\begin{equation}\label{eq:ess}
ESS(N) = N / \left(1 + 2 \times \sum_{k=1}^{N-1} (1 - \frac{k}{N}) \times R_k \right),
\end{equation}
where:
\begin{conditions}
N &=& total number of samples, and \\
R_k &=& the auto-correlation at lag $k$.
\end{conditions}

\begin{equation}
\label{eq:rhat}
\hat{R} = \frac{E[\text{Var}[X | \text{chain}]] + \text{Var}[E[X | \text{chain}]]}{E[\text{Var}[X | \text{chain}]]},
\end{equation}
where:
\begin{conditions}
E[\text{Var}[X | \text{chain}]]&=& the expected value of the variance of X given the chain, and\\

Var[E[X | \text{chain}]] & = & the variance of the expected value of X given the chain.\\

\end{conditions}

\paragraph{\textbf{OOD detection.}} \textcolor{black}{To create realistic OOD instances for the included datasets, we removed either the most or least occurring maneuvers. We avoided using instances from other datasets as OOD samples due to differences in sensor setups and video resolutions. The detection of OOD instances relative to in-distribution samples, based on their uncertainty estimates, can be formulated as a binary classification problem \cite{malinin2018predictive, durasov2024zigzag}. To formalize this, we assign a ground-truth label of \textit{0} to all in-distribution instances and \textit{1} to all OOD instances. The predictive uncertainty for each instance is then used as a continuous value to distinguish between the two classes. By comparing the ranked uncertainty scores against the binary ground-truth labels across all possible decision thresholds, we compute the Area Under the Receiver Operating Characteristic curve (ROC-AUC, $\uparrow$) and the Area Under the Precision-Recall curve (PR-AUC, $\uparrow$). Higher values for these metrics indicate better separability between uncertainty distributions of in-distribution and OOD data. Furthermore, we report the False Positive Rate at 95\% True Positive Rate (FPR95, $\downarrow$), which measures the proportion of OOD instances incorrectly classified as in-distribution when the model correctly identifies 95\% of in-distribution samples.}





\subsection{Underlying model architecture}
Given the relatively small size of the DAR and DIR datasets, we use a Kinetics-400 \citep{kay2017kinetics} self-supervised pre-trained Video Masked Autoencoder (VMAE, \cite{tong2022videomae}). As a backbone, the VMAE uses a Vision Transformer (ViT) Base architecture \citep{arnab2021vivit} that consists of about 94 million parameters. The ViT-Base expects videos of 16 frames, has a depth of 12 blocks and a hidden latent dimension of 768. For the datasets with multiple video streams, we combine the latent modality representations via attentional feature fusion \citep{dai2021attentional}.

\subsection{Implementation details \textcolor{black}{and hyperparameters}}
 We fine-tune the ViT-Base model for 20 epochs on each dataset using the AdamW optimizer \citep{loshchilov2018decoupled} with a weight decay of 0.05 \textcolor{black}{similar to \cite{vellenga2024evaluation}}. We apply early stopping with a patience of 5 epochs and set the learning rate to $5\mathrm{e}{-5}$. The experiments are run on an Ubuntu 20.04 server with two 16GB NVIDIA Tesla T4 GPUs. A gradient accumulation module from the Accelerate library (Gugger et al., \citeyear{accelerate}) is applied to simulate a batch size of 8.  \textcolor{black}{To create the deep ensemble, we perform this procedure 5 times, which results in five different ViT-Base models. For the included probabilistic last layer-based approaches (BBB, PE, SE, VBLL, LL--HMC), the latent representations are produced by the same ViT-Base models that yield the highest F1-score. In practice this means that Phase 1 is similar for these PDL methods to enable us to compare the effects of different PDL methods for Phase 2. For the DDU approach, we use the implementation of Mukhoti et al., \cite{mukhoti2023deep}, which applies GDA to quantify the separability of learned representations for each class. For the remaining PDL methods, Table \ref{tab:hyperparam_summary} shows the hyperparameters ranges for each method.}

  The LL--HMC experiments are run on a separate Ubuntu 20.04 server (CPU-only, 60GB RAM) using the NUTS sampler from Pyro \citep{bingham2018pyro}. \textcolor{black}{To tune the hyperparameters for LL--HMC,} we perform a grid search across 5 random seeds for different prior standard deviation (0.01, 0.1, 1, 2.5, 5, 10), number of burn-in samples (10, 25, 50, 100, 200), target acceptance probabilities (0.6, 0.7, 0.8), multiple chains (1, 2) and numbers of collected samples (2, 5, 10, ..., 50) for the classification layer $\theta_{LL}$. When we use more than a single chain, we divide the number of collected samples evenly over the chains. For the LL--HMC with multiple starting positions (latent representations from different independently fine-tuned models), we use the top-performing ensemble models to produce the latent representations.

\begin{table}[t]
\color{black}
\centering
\caption{Hyperparameters ranges for the LL--HMC, BBB--LL, PE--LL, SE, and VBLL methods. SD=Standard Deviation.}
\resizebox{0.65\linewidth}{!}{\begin{tabular}{l||c|c|c|c|c}
\toprule
\textbf{} & \textbf{LL--HMC} &  \textbf{BBB--LL}&  \textbf{PE--LL} &  \textbf{SE} &   \textbf{VBLL}  \\
\midrule
\begin{tabular}{@{}c@{}} Num Preds/Samples \end{tabular}  & \multicolumn{1}{c|}{2, 5, 10, ..., 50 } &  \multicolumn{3}{c|}{5, 10, ..., 50} & -- \\ \cline{1-6}
Batch Size  & \multirow{3}{*}{--}  & \multicolumn{4}{c}{16, 32, 64} \\ \cline{1-1} \cline{3-6}
Learning Rate    &  & \multicolumn{4}{c}{$1\mathrm{e}{-2}, 1\mathrm{e}{-3}, 1\mathrm{e}{-4}$} \\ \cline{3-6} \cline{1-1}
Epochs           &  & \multicolumn{4}{c}{5, 10, 15, 20, 25}\\ \cline{1-6}
Prior SD     & 0.01, 0.1, 1, 2.5, 5, 10&  \multicolumn{4}{c}{  \multirow{4}{*}{--}} \\ \cline{1-2} 
Burn-in Samples  & 10, 25, 50, 100, 200& \multicolumn{3}{c}{}\\ \cline{1-2}
Target Accept.   & 0.6, 0.7, 0.8& \multicolumn{3}{c}{} \\ \cline{1-2}
Num Chains       & 1, 2& \multicolumn{3}{c}{} \\ \cline{1-6}
KL Regularization& \multicolumn{4}{c|}{--}  &   1, 5, 10 \\  
\bottomrule
\end{tabular}}
\label{tab:hyperparam_summary}
\end{table}

For the VBLL \cite{harrison2024variational} and PE \cite{laurent2023packed} approaches, we use the original implementations, and for BBB--LL we use the Linear Reparameterization  layer from BayesianTorch \cite{krishnan2022bayesiantorch}. For the last-layer PDL (LL--PDL) methods, we perform a grid search across the same random seeds  \textcolor{black}{to find the optimal hyperparameter configurations. We} evaluate the number of predictions (5--50), batch size (16, 32, 64), learning rate ($1\mathrm{e}{-2}, 1\mathrm{e}{-3}, 1\mathrm{e}{-4}$) for different numbers of epochs (5 -- 25).  \textcolor{black}{While} VBLL does not produce multiple predictions, \textcolor{black}{we do evaluate the effects of the KL regularization parameter (1, 5, 10). This acts as a constraint that prevents the last layer's variational posterior from deviating excessively from the prior, thereby controlling model complexity and improving uncertainty calibration during training.}

\subsection{Experiments}

\subsubsection{Video-based DAR and DIR performance}
To compare LL--HMC to other PDL methods, we use the fine-tuned ViT-Base models as a regular baseline. The baseline models also produce the latent representations on which the other LL--PDL methods rely. Additionally, we also evaluate an ensemble of 5 independently fine-tuned ViT-Base models.  For the  LL--PDL methods, we include the following approaches that do not require any form of post-hoc calibration: SE  \citep{valdenegro2023sub},  DDU \citep{mukhoti2023deep}, BBB \citep{blundell2015weight}, PE \citep{laurent2023packed}, and VBLL \citep{harrison2024variational}.

\subsubsection{Uncertainty-based out-of-distribution detection} 
OOD detection aims to identify inputs that do not belong to the distribution of the training data \citep{yang2024generalized}. To construct realistic OOD instances for DAR and DIR, we remove the most and least prevalent classes from each dataset to simulate OOD instances for common or rare maneuvers. The removed classes from both the train and test datasets serve as the unseen OOD instances. We fine-tune a ViT-Base model on each reduced training set to obtain latent representations to use for the PDL methods. For the ROAD dataset, removing the most occurring maneuver would leave a subset of fewer than 200 videos. Therefore, for the ROAD dataset, we only apply the OOD-min setting, where we remove the maneuvers for which we have less than 20 videos (\textit{Move Right, Move Left, and Overtake}).

\begin{table*}[t]
\centering
\caption{Average performance and two standard errors of the mean of the top-performing hyperparameter configurations per random seed for the AIDE, B4C and ROAD datasets. Each PDL method, except for the deep ensemble, uses the latent representations produced by the \textit{Regular} model. Acc=Accuracy, ACE=Adaptive Calibration Error with 10 bins, rAULC=relative Area under the lifted curve. Best results are highlighted in \colorbox{best}{Blue}, and the second best in \colorbox{2best}{Red}.} 
\vspace{-0.2cm}

\resizebox{\linewidth}{!}{
\begin{tabular}{l|c|c|c|c||c|c|c|c||c|c|c|c}
\toprule
& \multicolumn{4}{c||}{\textbf{AIDE} \citep{yang2023aide}} & \multicolumn{4}{c||}{\textbf{B4C} \citep{jain2015car}} & \multicolumn{4}{c}{\textbf{ROAD} \citep{singh2022road}} \\
& \multicolumn{1}{c|}{\textbf{Acc ($\uparrow$)}} & \multicolumn{1}{c|}{\textbf{F1 ($\uparrow$)}} & \multicolumn{1}{c|}{\textbf{ACE ($\downarrow$)}} & \multicolumn{1}{c||}{\textbf{rAULC ($\uparrow$)}}
& \multicolumn{1}{c|}{\textbf{Acc ($\uparrow$)}} & \multicolumn{1}{c|}{\textbf{F1 ($\uparrow$)}} & \multicolumn{1}{c|}{\textbf{ACE ($\downarrow$)}} & \multicolumn{1}{c||}{\textbf{rAULC ($\uparrow$)}} & \multicolumn{1}{c|}{\textbf{Acc ($\uparrow$)}} & \multicolumn{1}{c|}{\textbf{F1 ($\uparrow$)}} & \multicolumn{1}{c|}{\textbf{ACE ($\downarrow$)}} & \multicolumn{1}{c}{\textbf{rAULC ($\uparrow$)}}  \\
\midrule
 \textbf{Regular}   & \multicolumn{1}{c|}{87.19}  & \multicolumn{1}{c|}{81.88}  & \multicolumn{1}{c|}{\cellcolor{2best}0.027} & \multicolumn{1}{c||}{0.77} & \multicolumn{1}{c|}{92.31} & \multicolumn{1}{c|}{93.71} & \multicolumn{1}{c|}{0.050} & \multicolumn{1}{c||}{0.53}  &    \multicolumn{1}{c|}{71.64}  &  \multicolumn{1}{c|}{45.36}   &   \multicolumn{1}{c|}{\cellcolor{best}0.063} &   \multicolumn{1}{c}{\cellcolor{2best}0.61}  \\
  \textbf{DE (N=5)} &  \multicolumn{1}{c|}{89.66\cellcolor{best}}  & \multicolumn{1}{c|}{82.24}  & \multicolumn{1}{c|}{\cellcolor{best}0.019} & \multicolumn{1}{c||}{\cellcolor{2best}0.80} &  \multicolumn{1}{c|}{90.88} & \multicolumn{1}{c|}{91.49} & \multicolumn{1}{c|}{0.059} &  \multicolumn{1}{c||}{\cellcolor{best}0.73}   &   \multicolumn{1}{c|}{\cellcolor{best}74.24}  &  \multicolumn{1}{c|}{44.57}   &   \multicolumn{1}{c|}{0.066} &  \multicolumn{1}{c}{\cellcolor{best}0.70}  \\
  \textbf{DDU} & \multicolumn{1}{c|}{57.31}  & \multicolumn{1}{c|}{19.12}  & \multicolumn{1}{c|}{0.088} & \multicolumn{1}{c||}{\cellcolor{best}0.85} &     \multicolumn{1}{c|}{76.07} & \multicolumn{1}{c|}{74.72} & \multicolumn{1}{c|}{0.124} & \multicolumn{1}{c||}{0.61} &  \multicolumn{1}{c|}{59.70}  &  \multicolumn{1}{c|}{23.85}   &   \multicolumn{1}{c|}{0.138} &  \multicolumn{1}{c}{0.42}  \\ \midrule
\textbf{BBB--LL} & 88.67$\pm$0.36 &  83.13$\pm$0.26 &  0.032$\pm$0.005 & 0.73$\pm$0.02&   \cellcolor{2best} 93.16$\pm$0.94 &94.28$\pm$0.87 &0.025$\pm$0.007 & 0.65$\pm$0.08  & 71.64$\pm$3.13 & \cellcolor{2best}56.14$\pm$4.37 & 0.068$\pm$0.003 &  0.50$\pm$0.13   \\

\textbf{PE--LL} & 88.64$\pm$0.32 & \cellcolor{2best} 83.37$\pm$0.02 &  0.030$\pm$0.003 & 0.74$\pm$0.01 &\cellcolor{best} 94.02$\pm$2.24  &   \cellcolor{best} 95.02$\pm$0.02 & \cellcolor{best}  0.022$\pm$0.005 & 0.67$\pm$0.03 & 68.66$\pm$2.11 & \cellcolor{best} 59.28$\pm$1.38 & 0.077$\pm$0.008 &  0.36$\pm$0.09   \\

\textbf{SE} & 88.31$\pm$0.19 &  82.69$\pm$0.21 &  0.029$\pm$0.000 & 0.75$\pm$0.01 & \cellcolor{2best} 93.16$\pm$0.18 &  \cellcolor{2best} 94.38$\pm$0.04 & \cellcolor{2best}  0.023$\pm$0.001 & \cellcolor{2best}  0.72$\pm$0.00 &  68.06$\pm$2.23 & 45.63$\pm$2.45 & \cellcolor{2best} 0.065$\pm$0.002 &  0.57$\pm$0.01 \\

\textbf{VBLL} & 88.67$\pm$0.36 &  83.13$\pm$0.26 &  0.032$\pm$0.005 &  0.73$\pm$0.02 &  92.14$\pm$1.26 &  93.47$\pm$1.13 & 0.116$\pm$0.038 &  0.55$\pm$0.10 & \cellcolor{2best}72.54$\pm$2.60 &  47.25$\pm$2.08 & 0.084$\pm$0.004 &  0.48$\pm$0.05 \\


\textbf{LL--HMC} & \cellcolor{2best} 89.00$\pm$0.16 & \cellcolor{best} 83.87$\pm$0.08 & 0.029$\pm$0.002 & 0.68$\pm$0.06 & \cellcolor{best} 94.02$\pm$0.42 & \cellcolor{2best}  95.00$\pm$0.29 & 0.027$\pm$0.007 & 0.66$\pm$0.10 &   70.15$\pm$5.05 &   54.18$\pm$4.43 & 0.073$\pm$0.014 & 0.41$\pm$0.12  \\







\bottomrule
\end{tabular}
}
\label{tab:performance}
\end{table*}

\begin{table*}[t!]
\centering
\caption{Average grid search results and two standard errors of the mean for the best average hyperparameter configuration across five random seeds. 
}
\vspace{-0.2cm}

\resizebox{\linewidth}{!}{
\begin{tabular}{l|c|c|c|c||c|c|c|c||c|c|c|c}
\toprule
& \multicolumn{4}{c||}{\textbf{AIDE} \citep{yang2023aide}} & \multicolumn{4}{c||}{\textbf{B4C} \citep{jain2015car}} & \multicolumn{4}{c}{\textbf{ROAD} \citep{singh2022road}} \\
& \multicolumn{1}{c|}{\textbf{Acc ($\uparrow$)}} & \multicolumn{1}{c|}{\textbf{F1 ($\uparrow$)}} & \multicolumn{1}{c|}{\textbf{ACE ($\downarrow$)}} & \multicolumn{1}{c||}{\textbf{rAULC ($\uparrow$)}}
& \multicolumn{1}{c|}{\textbf{Acc ($\uparrow$)}} & \multicolumn{1}{c|}{\textbf{F1 ($\uparrow$)}} & \multicolumn{1}{c|}{\textbf{ACE ($\downarrow$)}} & \multicolumn{1}{c||}{\textbf{rAULC ($\uparrow$)}} & \multicolumn{1}{c|}{\textbf{Acc ($\uparrow$)}} & \multicolumn{1}{c|}{\textbf{F1 ($\uparrow$)}} & \multicolumn{1}{c|}{\textbf{ACE ($\downarrow$)}} & \multicolumn{1}{c}{\textbf{rAULC ($\uparrow$)}}  \\
\midrule

\textbf{BBB--LL} &  \cellcolor{best} 88.47$\pm$0.41 &  \cellcolor{2best} 82.49$\pm$0.41 & \cellcolor{2best} 0.030$\pm$0.001 &0.70$\pm$0.05 & 90.09$\pm$1.16 &  91.26$\pm$1.70 & 0.028$\pm$0.005 & 0.80$\pm$0.02    & 63.58$\pm$5.05 &  \cellcolor{best} 34.51$\pm$1.72 & \cellcolor{best} 0.077$\pm$0.007 & 0.62$\pm$0.04 \\

\textbf{PE--LL} & 87.85$\pm$0.63 & 80.61$\pm$2.12 &0.034$\pm$0.004 & \cellcolor{best} 0.79$\pm$0.02 & \cellcolor{2best} 91.97$\pm$0.42 & \cellcolor{2best} 91.99$\pm$0.45 &  \cellcolor{best}  0.021$\pm$0.002 &  \cellcolor{2best} 0.84$\pm$0.01  &  65.37$\pm$1.12 & 29.72$\pm$1.12 & 0.092$\pm$0.002 & \cellcolor{best} 0.67$\pm$0.02 \\

\textbf{SE} & 88.11$\pm$0.13 &  81.61$\pm$0.11 & \cellcolor{best}  0.028$\pm$0.001 & 0.77$\pm$0.01 &  \cellcolor{best} 92.14$\pm$0.34 & \cellcolor{best} 92.21$\pm$0.26 & \cellcolor{2best} 0.022$\pm$0.002 & 0.82$\pm$0.00 & \cellcolor{2best} 65.97$\pm$0.60 &  29.45$\pm$0.32 & 0.103$\pm$0.002 & \cellcolor{best} 0.67$\pm$0.02 \\

\textbf{VBLL} & 87.95$\pm$0.17 & 78.62$\pm$3.17 & 0.034$\pm$0.003 &  \cellcolor{2best} 0.78$\pm$0.02 & 86.32$\pm$3.01 & 84.92$\pm$3.17 & 0.081$\pm$0.007 & 0.80$\pm$0.07 & \cellcolor{best} 66.27$\pm$1.52 & 31.29$\pm$1.71 & 0.108$\pm$0.005 &  \cellcolor{2best} 0.63$\pm$0.03 \\ 


\textbf{LL--HMC} &   \cellcolor{2best} 88.44$\pm$0.22 & \cellcolor{best} 82.69$\pm$0.26 & \cellcolor{best} 0.028$\pm$0.000 &  0.76$\pm$0.01 & 90.26$\pm$1.39 & 90.17$\pm$2.46 &  0.027$\pm$0.004 & \cellcolor{best} 0.85$\pm$0.03 & 61.79$\pm$4.07 &  \cellcolor{2best} 32.03$\pm$4.00 &   \cellcolor{2best} 0.083$\pm$0.010 & \cellcolor{2best} 0.63$\pm$0.15 \\


\bottomrule
\end{tabular}
}
\label{tab:performance_avghp}
\end{table*}

\section{Results}
To avoid reliance on a single random initialization \cite{bouthillier2019unreproducible}, we examine classification and OOD detection performance differences across multiple random seeds. For both the in-distribution and OOD experiments, we first analyze the grid search results across five random seeds. We present the top performance averaged across random seeds to illustrate classification and OOD detection results for a single grid search. Next, we report the performance of the best average hyperparameter configuration across the random seeds. To better understand the differences between the included LL--PDL methods, as well as the effects of additional predictions or sampled last-layer parameters, we use the best average hyperparameter configuration and conduct further evaluations with more random seeds. For LL--HMC, we also analyze the influence of the prior, additional chains, additional starting positions, parameter trajectories, and MCMC diagnostics.


\subsection{In-distribution action and intention recognition performance}  
\paragraph{\textbf{Grid search results.}} Table  \ref{tab:performance} reports the average performance of the top hyperparameter configuration per random seed. Table \ref{tab:performance_avghp} shows the performance of the single hyperparameter configuration that achieved the best average across the random seeds. The regular DNN serves as the baseline, and the DE and DDU were only evaluated once per dataset. For the DDU, the lower performance is most likely due to the absence of spectral normalization. The variability in DE performance may be attributed to the small ensemble size (N=5), which limits its robustness. Given the computational cost of adding more members, DE may not be practical for resource-constrained operating environments.

The top-performing LL--HMC configurations in Table \ref{tab:performance} consistently achieve performance comparable to the best-performing PDL methods. From a confidence-performance calibration perspective (ACE), the BBB--LL, PE--LL, SE and LL--HMC yield a lower ACE for the B4C and ROAD datasets compared to the regular model, but not for the AIDE dataset. Additionally, no single method optimally balances both classification performance and calibration across all datasets. The regular softmax-based model is better calibrated than the LL--HMC approach, except for the B4C dataset. Additionally, we do not observe that the higher classification performance of LL--HMC in Table \ref{tab:performance} also results in an improved rAULC (the ability to distinguish between correctly and incorrectly predicted in-distribution instances based on the uncertainty estimations). 

While Table  \ref{tab:performance} presents the potential top-performance we could observe when we would perform a single grid search, Table \ref{tab:performance_avghp}  reveals that using the average hyperparameter configuration across seeds leads to significantly lower average in-distribution classification performance. More specifically, the F1-scores dropped on average with 2.1, 4.3 and 21.1 for the AIDE, B4C and ROAD datasets respectively. The larger performance drop on the ROAD dataset suggests that LL--PDL methods may be more sensitive to observation noise and initialization instability, particularly in smaller or noisier datasets.

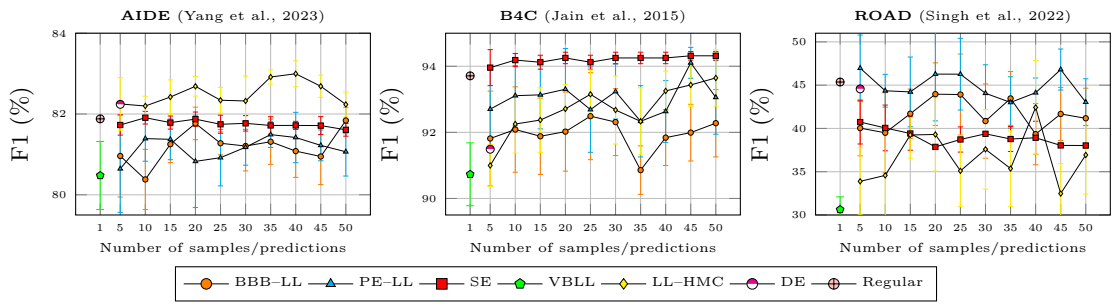
\begin{figure*}[!t]
\begin{tikzpicture}
\begin{groupplot}[
 group style={
group size=3 by 1,
horizontal sep=1.0cm,
 vertical sep=1.0cm,
 },
width=5.5cm,
height=4.0cm,
grid=major,
 xlabel={Number of samples/predictions},
 xtick={1,5,10,15,20,25,30,35,40,45,50},
tick label style={font=\small},
label style={font=\small},
 y tick label style={
 /pgf/number format/fixed,
 /pgf/number format/precision=0,
 /pgf/number format/fixed zerofill=true,
 },
 tick label style={font=\fontsize{4}{4}\selectfont},
xlabel style={yshift=0.1cm}, 
]

\nextgroupplot[
title={{\textbf{AIDE} \citep{yang2023aide}}},
title style={yshift=-0.2cm},
ylabel={\small{F1 (\%)}},
 ymin=79.5,
ymax=84
]

\addplot[mark=*, thin, line width=0.25pt, mark options={fill=orange}, mark size=1.0pt, error bars/.cd,
y dir=both,
 y explicit,
error bar style={line width=0.1pt, color=orange}] coordinates {
(5,80.95996846938037) += (0,1.0157) -=(0,1.0157)
(10,80.38051179888186) += (0,0.7448) -=(0,0.7448)
(15,81.2522205585229) += (0,0.4554) -=(0,0.4554)
(20,81.76684349204947) += (0,0.4032) -=(0,0.4032)
(25,81.27572973546255) += (0,0.2846) -=(0,0.2846)
(30,81.20717689275352) += (0,0.6158) -=(0,0.6158)
(35,81.31343389831275) += (0,0.5597) -=(0,0.5597)
(40,81.08118064051627) += (0,0.6506) -=(0,0.6506)
(45,80.94812310733515) += (0,0.6945) -=(0,0.6945)
(50,81.83969173721589) += (0,0.3263) -=(0,0.3263)
};

\addplot[mark=triangle*, thin, line width=0.25pt, mark options={fill=cyan}, mark size=1pt,error bars/.cd,
y dir=both,
 y explicit,
error bar style={line width=0.1pt, color=cyan}] coordinates {
(5,80.64427423685513) += (0,1.0801) -=(0,1.0801)
(10,81.39108033133715) += (0,0.5604) -=(0,0.5604)
(15,81.37295306180285) += (0,0.5042) -=(0,0.5042)
(20,80.82737023929269) += (0,1.1425) -=(0,1.1425)
(25,80.91886472940277) += (0,0.6976) -=(0,0.6976)
(30,81.17688066908337) += (0,0.4442) -=(0,0.4442)
(35,81.49046003528815) += (0,0.3197) -=(0,0.3197)
(40,81.41813206619261) += (0,0.6226) -=(0,0.6226)
(45,81.22341562264936) += (0,0.3816) -=(0,0.3816)
(50,81.064988976834) += (0,0.6010) -=(0,0.6010)
};

\addplot[mark=square*, thin, line width=0.25pt, mark options={fill=red}, mark size=1.0pt,
error bars/.cd,
y dir=both,
y explicit,
error bar style={line width=0.1pt, color=red}] coordinates {
(5,81.72796641439183) += (0,0.2508) -=(0,0.2508)
(10,81.90801885064703) += (0,0.1565) -=(0,0.1565)
(15,81.78739214969264) += (0,0.1620) -=(0,0.1620)
(20,81.87270775771678) += (0,0.1763) -=(0,0.1763)
(25,81.74547366757284) += (0,0.2244) -=(0,0.2244)
(30,81.76892921540839) += (0,0.1918) -=(0,0.1918)
(35,81.71655273815813) += (0,0.2177) -=(0,0.2177)
(40,81.7176956093715) += (0,0.1000) -=(0,0.1000)
(45,81.71353922074951) += (0,0.2220) -=(0,0.2220)
(50,81.60908470361797) += (0,0.1585) -=(0,0.1585)
};

\addplot[mark=diamond*, thin, line width=0.25pt, mark options={fill=yellow}, mark size=1.0pt,
error bars/.cd,
y dir=both,
 y explicit,
error bar style={line width=0.1pt, color=yellow}] coordinates {
(5,82.24926660396423) += (0,0.6520) -=(0,0.6520)
(10,82.19827942316209) += (0,0.2412) -=(0,0.2412)
(15,82.42064618224457) += (0,0.4252) -=(0,0.4252)
(20,82.68614557647774) += (0,0.2421) -=(0,0.2421)
(25,82.34071008535416) += (0,0.3180) -=(0,0.3180)
(30,82.3229477909996) += (0,0.6125) -=(0,0.6125)
(35,82.91580828836082) += (0,0.1741) -=(0,0.1741)
(40,82.99625162343334) += (0,0.3213) -=(0,0.3213)
(45,82.68709960581347) += (0,0.2738) -=(0,0.2738)
(50,82.23586915620874) += (0,0.3072) -=(0,0.3072)
};

\addplot[mark=halfcircle*, thin, line width=0.25pt, mark options={fill=magenta}, mark size=1.5pt] coordinates {
(5,82.24)
};

\addplot[mark=oplus*, thin, line width=0.25pt, mark options={fill=pink}, mark size=1.5pt] coordinates {
(1,81.88)
};


\addplot[mark=pentagon*, thin, line width=0.25pt, mark options={fill=green}, mark size=1.5pt,
error bars/.cd,
y dir=both,
 y explicit,
error bar style={line width=0.1pt, color=green}] coordinates {
(1,80.478158) += (0,0.8418) -=(0,0.8418)
};

\nextgroupplot[
title={{\textbf{B4C} \citep{jain2015car}}},
title style={yshift=-0.2cm},
ylabel={\small{F1 (\%)}},
ymax=95,
ymin=89.5,
]

\addplot[mark=*, thin, line width=0.25pt, mark options={fill=orange}, mark size=1.0pt,error bars/.cd,
y dir=both,
y explicit,
error bar style={line width=0.1pt, color=orange}] coordinates {
(5,91.81428553594436) += (0,1.4334) -=(0,1.4334)
(10,92.08634408387084) += (0,1.2917) -=(0,1.2917)
(15,91.88171836718176) += (0,1.1540) -=(0,1.1540)
(20,92.02217647893363) += (0,1.1959) -=(0,1.1959)
(25,92.4902186633233) += (0,1.3156) -=(0,1.3156)
(30,92.3112903345926) += (0,1.0125) -=(0,1.0125)
(35,90.85696435247613) += (0,0.7344) -=(0,0.7344)
(40,91.84051123373924) += (0,0.8450) -=(0,0.8450)
(45,91.99210892719971) += (0,0.8574) -=(0,0.8574)
(50,92.27768581229043) += (0,1.0191) -=(0,1.0191)
};

\addplot[mark=triangle*, thin, line width=0.25pt, mark options={fill=cyan}, mark size=1pt,error bars/.cd,
y dir=both,
y explicit,
error bar style={line width=0.1pt, color=cyan}] coordinates {
(5,92.7037411936547) += (0,1.2931) -=(0,1.2931)
(10,93.11025366083305) += (0,0.9090) -=(0,0.9090)
(15,93.13511115961856) += (0,1.0287) -=(0,1.0287)
(20,93.30473881052755) += (0,1.2327) -=(0,1.2327)
(25,92.68852537845807) += (0,1.2925) -=(0,1.2925)
(30,93.23527829587042) += (0,0.8315) -=(0,0.8315)
(35,92.33493213991402) += (0,1.0786) -=(0,1.0786)
(40,92.63245413657515) += (0,0.9345) -=(0,0.9345)
(45,94.10309815824014) += (0,0.4679) -=(0,0.4679)
(50,93.05837606672831) += (0,1.1152) -=(0,1.1152)
};

\addplot[mark=square*, thin, line width=0.25pt, mark options={fill=red}, mark size=1.0pt,
error bars/.cd,
y dir=both,
y explicit,
error bar style={line width=0.1pt, color=red}] coordinates {
(5.0,93.9590185161521) += (0,0.5442) -=(0,0.5442)
(10.0,94.18708528392156) += (0,0.1981) -=(0,0.1981)
(15.0,94.12224595632553) += (0,0.2117) -=(0,0.2117)
(20.0,94.2519246115176) += (0,0.1729) -=(0,0.1729)
(25.0,94.12224595632551) += (0,0.2117) -=(0,0.2117)
(30.0,94.2519246115176) += (0,0.1729) -=(0,0.1729)
(35.0,94.2519246115176) += (0,0.1729) -=(0,0.1729)
(40.0,94.2519246115176) += (0,0.1729) -=(0,0.1729)
(45.0,94.31676393911363) += (0,0.1297) -=(0,0.1297)
(50.0,94.31676393911363) += (0,0.1297) -=(0,0.1297)
};

\addplot[mark=diamond*, thin, line width=0.25pt,
mark options={fill=yellow},
mark size=1.0pt,
error bars/.cd,
y dir=both,
y explicit,
error bar style={line width=0.1pt, color=yellow}] coordinates {
(5,90.99785760859635) += (0,0.6351) -=(0,0.6351)
(10,92.25213285534451) += (0,0.8798) -=(0,0.8798)
(15,92.3731837998395) += (0,0.9760) -=(0,0.9760)
(20,92.71090059658137) += (0,0.7351) -=(0,0.7351)
(25,93.1521381369241) += (0,0.7345) -=(0,0.7345)
(30,92.67648931096207) += (0,1.0303) -=(0,1.0303)
(35,92.33309758977552) += (0,1.0145) -=(0,1.0145)
(40,93.25363425383188) += (0,0.6015) -=(0,0.6015)
(45,93.43567253225955) += (0,0.6504) -=(0,0.6504)
(50,93.64543100334956) += (0,0.8540) -=(0,0.8540)
};

\addplot[mark=halfcircle*, thin, line width=0.25pt, mark options={fill=magenta}, mark size=1.5pt] coordinates {
(5,91.49)
};

\addplot[mark=oplus*, thin, line width=0.25pt, mark options={fill=pink}, mark size=1.5pt] coordinates {
(1,93.71)
};

\addplot[mark=pentagon*, thin, line width=0.25pt, mark options={fill=green}, mark size=1.5pt,
error bars/.cd,
y dir=both,
y explicit,
error bar style={line width=0.1pt, color=green}] coordinates {
(1,90.7315) += (0,0.9509) -=(0,0.9509)
};

\nextgroupplot[
title={{\textbf{ROAD} \citep{singh2022road}}},
title style={yshift=-0.2cm},
ylabel={\small{F1 (\%)}},
ymin=30,
ymax=51
]

\addplot[mark=*, thin, line width=0.25pt, mark options={fill=orange}, mark size=1.0pt, error bars/.cd,
y dir=both,
y explicit,
error bar style={line width=0.1pt, color=orange}] coordinates {
(5,40.03344690170473) += (0,3.2007) -=(0,3.2007)
(10,39.497117850555775) += (0,4.6328) -=(0,4.6328)
(15,41.70045043815289) += (0,2.8485) -=(0,2.8485)
(20,43.95679573963744) += (0,3.6190) -=(0,3.6190)
(25,43.91655890502581) += (0,4.6789) -=(0,4.6789)
(30,40.84849670921552) += (0,4.2924) -=(0,4.2924)
(35,43.46600586280709) += (0,3.1102) -=(0,3.1102)
(40,39.33604574991489) += (0,3.5332) -=(0,3.5332)
(45,41.67154706867009) += (0,3.0697) -=(0,3.0697)
(50,41.170672005887695) += (0,3.4761) -=(0,3.4761)
};

\addplot[mark=triangle*, thin, line width=0.25pt, mark options={fill=cyan}, mark size=1pt,error bars/.cd,
y dir=both,
y explicit,
error bar style={line width=0.1pt, color=cyan}] coordinates {
(5,46.95914301867057) += (0,3.8279) -=(0,3.8279)
(10,44.35103292152759) += (0,1.8950) -=(0,1.8950)
(15,44.21816213450458) += (0,4.0279) -=(0,4.0279)
(20,46.27727705113644) += (0,5.4009) -=(0,5.4009)
(25,46.26570675100138) += (0,4.1432) -=(0,4.1432)
(30,44.06082062870729) += (0,3.2937) -=(0,3.2937)
(35,43.00217105321485) += (0,2.9669) -=(0,2.9669)
(40,44.123030934571204) += (0,1.7077) -=(0,1.7077)
(45,46.803628670707965) += (0,2.3713) -=(0,2.3713)
(50,43.03308469196917) += (0,2.7090) -=(0,2.7090)
};

\addplot[mark=square*, thin, line width=0.25pt, mark options={fill=red}, mark size=1.0pt,
error bars/.cd,
y dir=both,
y explicit,
error bar style={line width=0.1pt, color=red}] coordinates {

(5,40.73743947791) += (0,2.5367) -=(0,2.5367)
(10,40.056428478757425) += (0,2.6397) -=(0,2.6397)
(15,39.41337466881583) += (0,1.9312) -=(0,1.9312)
(20,37.86591958163472) += (0,0.1186) -=(0,0.1186)
(25,38.723007197626444) += (0,1.4574) -=(0,1.4574)
(30,39.378180178826824) += (0,1.7709) -=(0,1.7709)
(35,38.776521032744974) += (0,1.4436) -=(0,1.4436)
(40,38.91944120501935) += (0,1.9113) -=(0,1.9113)
(45,38.03843460972704) += (0,0.0723) -=(0,0.0723)
(50,38.02956311592404) += (0,0.0777) -=(0,0.0777)
};

\addplot[mark=diamond*, thin, line width=0.25pt, mark options={fill=yellow}, mark size=1.0pt, error bars/.cd,
y dir=both, y explicit,
error bar style={line width=0.1pt, color=yellow}] coordinates {
(5,33.887826567916576) += (0,3.8856) -=(0,3.8856)
(10,34.574903713756974) += (0,5.5229) -=(0,5.5229)
(15,39.24376817031906) += (0,2.6870) -=(0,2.6870)
(20,39.30141891898767) += (0,4.2289) -=(0,4.2289)
(25,35.09617883452023) += (0,4.1437) -=(0,4.1437)
(30,37.58546499845445) += (0,4.5910) -=(0,4.5910)
(35,35.363634910564876) += (0,4.3608) -=(0,4.3608)
(40,42.42697302678621) += (0,5.4215) -=(0,5.4215)
(45,32.4746292995812) += (0,3.4674) -=(0,3.4674)
(50,36.93272588550002) += (0,4.5218) -=(0,4.5218)
};

\addplot[mark=halfcircle*, thin, line width=0.25pt, mark options={fill=magenta}, mark size=1.5pt] coordinates {
(5,44.57)
};

\addplot[mark=oplus*, thin, line width=0.25pt, mark options={fill=pink}, mark size=1.5pt] coordinates {
(1,45.36)
};

\addplot[mark=pentagon*, thin, line width=0.25pt, mark options={fill=green}, mark size=1.5pt,
error bars/.cd,
y dir=both,
y explicit,
error bar style={line width=0.1pt, color=green}] coordinates {
(1,30.63016) += (0,1.4655) -=(0,1.4655)
};

\end{groupplot}
\end{tikzpicture}

\begin{tikzpicture}
\vspace{0.3cm} 
\begin{axis}[
hide axis,
xmin=0, xmax=1, ymin=0, ymax=1,
legend style={at={(0.0,0.0)}, anchor=center, legend columns=7},
legend entries={\tiny{BBB--LL},\tiny{PE--LL},\tiny{SE},\tiny{VBLL}, \tiny{LL--HMC},\tiny{DE},\tiny{Regular},}
]

\addlegendimage{mark=*, thin, line width=0.25pt, mark options={fill=orange}, mark size=2.0pt}
\addlegendimage{mark=triangle*, thin, line width=0.25pt, mark options={fill=cyan}, mark size=2.0pt}
\addlegendimage{mark=square*, thin, line width=0.25pt, mark options={fill=red}, mark size=2pt}
\addlegendimage{mark=pentagon*, thin, line width=0.25pt, mark options={fill=green}, mark size=2.0pt}
\addlegendimage{mark=diamond*, thin, line width=0.25pt, mark options={fill=yellow}, mark size=2.0pt}
\addlegendimage{mark=halfcircle*, thin, line width=0.25pt, mark options={fill=magenta}, mark size=2.0pt}
\addlegendimage{mark=oplus*, thin, line width=0.25pt, mark options={fill=pink}, mark size=2.0pt}

\end{axis}
\end{tikzpicture}

\caption{Influence of the number of predictions \textcolor{black}{(e.g., ensemble members, or last layer parameter} samples) for independent random seeds for the average best-performing hyperparameter configurations of the included PDL methods. The regular model (single prediction), DE (N=5), and VBLL (single prediction) are only included once in each figure. For each number of samples, the whiskers indicate two standard  errors of the mean  for 10 runs for different random seeds for every number of samples to avoid dependencies.}
\label{fig:method_comparison}

\end{figure*}


\definecolor{steelblue31119180}{RGB}{31,119,180} 


\definecolor{steelblue31119180}{RGB}{31,119,180} 

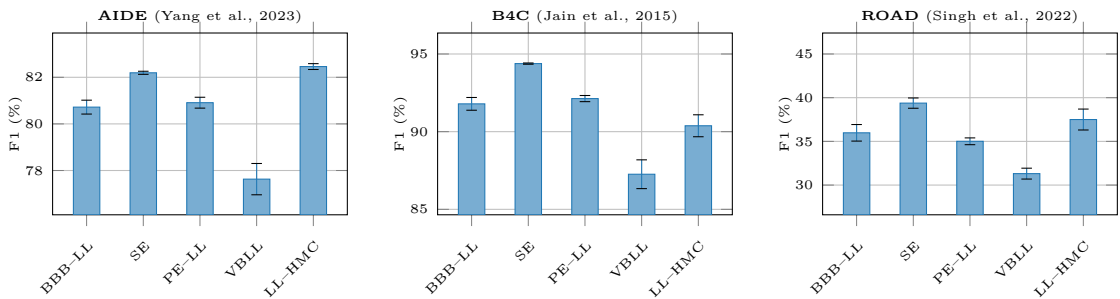
\begin{figure}[t!]
    \centering
    \begin{tikzpicture}
        \begin{groupplot}[
            group style={
                group size=3 by 1, 
                horizontal sep=1.2cm,
                vertical sep=1.0cm,
            },
            width=5.5cm,
            height=4.0cm,
            ybar,
            enlargelimits=0.15,
            ylabel={\small {F1 (\%)}},
            symbolic x coords={BBB--LL, SE, PE--LL, VBLL, LL--HMC},
            xtick=data,
            ymin=0, ymax=100,
            grid=major,
            ymajorgrids=true,
            tick label style={font=\tiny},
            every node near coord/.append style={
                yshift=3pt,
                font=\footnotesize,
    label style={font=\small},
    y tick label style={
        /pgf/number format/fixed,
        /pgf/number format/precision=0,
        /pgf/number format/fixed zerofill=true, 
    },
    tick label style={font=\fontsize{4}{4}\selectfont},
    xlabel style={yshift=0.1cm},  
            },
        ]

        \nextgroupplot[title={ \textbf{AIDE} \citep{yang2023aide}},
         title style={yshift=-0.2cm},
         ylabel style={yshift=-0.2cm},
           ymin=77, ymax=83,]
        \addplot[
            fill=steelblue31119180!60,
            draw=steelblue31119180,
            error bars/.cd,
            y dir=both,
            y explicit,
        ] coordinates {
        (SE, 82.1926) +- (0.06622,0.06622)

                (BBB--LL, 80.72) +- (0.29784,0.29784)
(PE--LL, 80.91) +- (0.23318,0.23318)
(VBLL, 77.63) +- (0.67146,0.67146)
(LL--HMC, 82.46) +- (0.12536,0.12536)

        };

        \nextgroupplot[title={ \textbf{B4C}  \citep{jain2015car}},
   title style={yshift=-0.2cm},
   ylabel style={yshift=-0.2cm},
     ymin=86, ymax=95,]
        \addplot[
            fill=steelblue31119180!60,
            draw=steelblue31119180,
            error bars/.cd,
            y dir=both,
            y explicit,
        ] coordinates {

(SE, 94.3816) +- (0.04,0.04)
(BBB--LL, 91.79) +- (0.41068,0.41068)
(PE--LL, 92.13) +- (0.19792,0.19792)
(VBLL, 87.26) +- (0.92628,0.92628)
(LL--HMC, 90.38) +- (0.70676,0.70676)

        };

     \nextgroupplot[title={ \textbf{ROAD}  \citep{singh2022road}},
   title style={yshift=-0.2cm},  ymin=29, ymax=45, ylabel style={yshift=-0.2cm},]
        \addplot[
            fill=steelblue31119180!60,
            draw=steelblue31119180,
            error bars/.cd,
            y dir=both,
            y explicit,
        ] coordinates {
(SE, 39.379) +- (0.58896,0.58896)
(BBB--LL, 35.98) +- (0.94728,0.94728)
(PE--LL,35.01) +- (0.38868,0.38868)
(VBLL, 31.31) +- (0.62208,0.62208)
(LL--HMC, 37.5) +- (1.19686,1.19686)

        };
        \end{groupplot}
    \end{tikzpicture}
    \caption{Average F1-scores and two standard errors of the mean across the included LL--PDL methods for the best performing grid search hyperparameter configurations across 100 random seeds.}
    \label{fig:id_performance}
\end{figure}

\paragraph{\textbf{Influence of additional predictions or samples.}} 
For the ViT-Base architecture, processing a single 16-frame video requires approximately $361 \times 10^9$ FLOPs. The majority of the FLOPs are required to produce the latent representation on which the LL--PDL methods rely. Every additional last layer sample or prediction requires $7.68$ or $9.22 \times 10^3$ FLOPs for a classifier with 5 or 6 classes respectively. Considering the $361\times 10^9$ FLOPs that are required to produce the latent representation $\mathbf{z}$, the computational impact of the additional predictions are relatively negligible. To examine the influence of additional predictions or sampled last layers, we evaluated the top configurations from Table \ref{tab:performance_avghp} for 10 different random seeds for each number of samples. Figure \ref{fig:method_comparison} illustrates the average performance and two standard errors of the mean for each number of samples. For the LL--PDL methods, no clear trend indicates that additional samples (or predictions) consistently improves the performance. Additionally, we observe that for each dataset another LL--PDL method performs best on average.

\paragraph{\textbf{Comparing the best hyperparameter configurations. }}

To evaluate whether any of the included LL--PDL methods consistently outperforms others, we tested the average best performing hyperparameter configuration from Table \ref{tab:performance_avghp} across 100 different random seeds.  Figure \ref{fig:id_performance} illustrates the average DAR and DIR classification performance for each dataset and method. Consistent with the results in Figure \ref{fig:method_comparison}, no method consistently achieves the best average performance across the datasets. VBLL exhibits higher standard errors across datasets, suggesting greater sensitivity to random seed variations. As also observed in Table \ref{tab:performance_avghp}, the performance degradation from top-performing configurations is more pronounced in the smaller ROAD dataset. 

\begin{figure*}[t]
\begin{tikzpicture}
\begin{groupplot}[
    group style={
        group size=3 by 1,
        horizontal sep=1.2cm,
        vertical sep=1.5cm,
    },
    width=5.5cm,
    height=4.0cm,
    grid=major,
    xlabel={Number of samples},
    xtick={2,5,10,15,20,25,30,35,40,45,50},
    tick label style={font=\small},
    label style={font=\small},
    y tick label style={
        /pgf/number format/fixed,
        /pgf/number format/precision=0,
        /pgf/number format/fixed zerofill=true, 
    },
    tick label style={font=\fontsize{4}{4}\selectfont},
    xlabel style={yshift=0.1cm},  
]

\nextgroupplot[
    title={{\textbf{AIDE} \citep{yang2023aide}}},
    ylabel={\small{F1 (\%)}},
          title style={yshift=-0.2cm},
    ymin=0,
    ymax=100 
]

\addplot[mark=*, thin, line width=0.25pt, mark options={fill=orange}, mark size=1.0pt] coordinates {
   (2,14.0021344717182)
(5,14.017094017094)
(10,14.0021344717182)
(15,14.0021344717182)
(20,14.0021344717182)
(25,14.0021344717182)
(30,14.0021344717182)
(35,14.0021344717182)
(40,14.0021344717182)
(45,14.0021344717182)
(50,14.0021344717182)

};

\addplot[mark=triangle*, thin, line width=0.25pt, mark options={fill=cyan}, mark size=1.0pt] coordinates {
  (2,44.0785617720902)
(5,45.5579271170791)
(10,44.619027696427)
(15,45.6406287596942)
(20,45.6524267426279)
(25,46.6828965371792)
(30,45.1265108186647)
(35,45.294367914765)
(40,45.294367914765)
(45,45.294367914765)
(50,45.294367914765)
};

\addplot[mark=square*, thin, line width=0.25pt, mark options={fill=red}, mark size=1pt] coordinates {
     (2,81.1265996341357)
(5,81.4880528149863)
(10,80.6886033388472)
(15,80.8240769818706)
(20,80.8240769818706)
(25,81.7497542395929)
(30,81.147775318895)
(35,80.6886033388472)
(40,81.2682541885595)
(45,81.2682541885595)
(50,80.8240769818706)
};

\addplot[mark=pentagon*, thin, line width=0.25pt, mark options={fill=green}, mark size=1.0pt] coordinates {
  (2,81.9988673273403)
(5,81.0725353796292)
(10,83.2210460012779)
(15,82.8653507807344)
(20,82.6611086890146)
(25,82.5995157384987)
(30,82.2820554210384)
(35,82.932999137567)
(40,83.5649615920802)
(45,82.3797680158336)
(50,82.8254570288468)
};

\addplot[mark=diamond*, thin, line width=0.25pt, mark options={fill=pink}, mark size=1.0pt] coordinates {
  (2,82.5713665407654)
(5,81.3040212030037)
(10,82.3251434729353)
(15,82.3119087170357)
(20,82.6568972522956)
(25,83.0308431010185)
(30,82.5466618076418)
(35,82.2070251182043)
(40,83.3504293628523)
(45,82.5884467876544)
(50,82.7713852772888)
};

\addplot[mark=oplus*, thin, line width=0.25pt, mark options={fill=yellow,}, mark size=1.0pt] coordinates {
   (2,80.7062045482558)
(5,81.8924768551738)
(10,81.6863324033022)
(15,81.7573352754438)
(20,82.0510107716231)
(25,81.8958601891935)
(30,82.1463683907081)
(35,81.4699359624406)
(40,81.894579329894)
(45,81.9912178082785)
(50,82.3647639021652)
};

\nextgroupplot[
    title={{\textbf{B4C} \citep{jain2015car}}},
    ylabel={\small{F1 (\%)}},
          title style={yshift=-0.2cm},
    ymax=100,
    ymin=0,
]

\addplot[mark=*, thin, line width=0.25pt, mark options={fill=orange, }, mark size=1.0pt] coordinates {
    (2,11.0045871559633)
(2,11.0045871559633)
(5,6.80851063829787)
(10,11.6363636363636)
(15,11.6363636363636)
(20,16.1375661375661)
(25,12.1518987341772)
(30,11.6363636363636)
(35,11.6363636363636)
(40,11.7791411042944)
(45,11.6363636363636)
(50,11.6363636363636)
};

\addplot[mark=triangle*, thin, line width=0.25pt, mark options={fill=cyan, }, mark size=1.0pt] coordinates {
  (2,11.6363636363636)
(5,11.6363636363636)
(10,11.6363636363636)
(15,11.6363636363636)
(20,11.6363636363636)
(25,11.6363636363636)
(30,11.6363636363636)
(35,11.6363636363636)
(40,11.6363636363636)
(45,11.6363636363636)
(50,11.6363636363636)

};

\addplot[mark=square*, thin, line width=0.25pt, mark options={fill=red, }, mark size=1.0pt] coordinates {
   (2,92.612915759661)
(5,89.2186741363212)
(10,90.3246376811594)
(15,90.2082047003198)
(20,90.2082047003198)
(25,90.3246376811594)
(30,90.3246376811594)
(35,90.3246376811594)
(40,90.2374331550802)
(45,90.3246376811594)
(50,91.4892504519538)

};

\addplot[mark=pentagon*, thin, line width=0.25pt, mark options={fill=green, }, mark size=1.0pt] coordinates {
   (2,93.1248656780571)
(5,91.9724578203374)
(10,92.0444444444444)
(15,93.5018046485499)
(20,92.882146160962)
(25,91.991304347826)
(30,93.7147086031452)
(35,92.6623462346234)
(40,92.882146160962)
(45,92.0444444444444)
(50,94.3356005165734)

};

\addplot[mark=diamond*, thin, line width=0.25pt, mark options={fill=pink, }, mark size=1.0pt] coordinates {
  (2,89.0888888888888)
(5,89.452427184466)
(10,91.0268584403197)
(15,93.546728151076)
(20,94.3356005165734)
(25,94.3356005165734)
(30,92.8344673265824)
(35,92.8777777777777)
(40,95.0027752081406)
(45,94.3816032667096)
(50,92.882146160962)

};

\addplot[mark=oplus*, thin, line width=0.25pt, mark options={fill=yellow, }, mark size=1.0pt] coordinates {
   (2,85.2579288323969)
(5,90.2374331550802)
(10,91.7308282160585)
(15,95.0027752081406)
(20,93.7332099907493)
(25,90.8346405228758)
(30,94.3356005165734)
(35,92.8777777777777)
(40,92.7064935064935)
(45,92.8777777777777)
(50,90.7626538987688)
};

\nextgroupplot[
    title={{\textbf{ROAD} \citep{singh2022road}}},
    ylabel={\small{F1 (\%)}},
          title style={yshift=-0.2cm},
    ymin=0,
    ymax=60
]

\addplot[mark=*, thin, line width=0.25pt, mark options={fill=orange}, mark size=1.0pt] coordinates {
    (2,8.83652430044182)
(5,4.64285714285714)
(10,13.5233444616074)
(15,8.83652430044182)
(20,20.1298701298701)
(25,19.5238095238095)
(30,8.83652430044182)
(35,8.83652430044182)
(40,8.83652430044182)
(45,8.83652430044182)
(50,10.969387755102)
};

\addplot[mark=triangle*, thin, line width=0.25pt, mark options={fill=cyan}, mark size=1.0pt] coordinates {
     (2,12.8320802005012)
(5,18.4417086672725)
(10,17.2532781228433)
(15,12.8320802005012)
(20,15.9392789373814)
(25,14.4756838905775)
(30,15.9392789373814)
(35,10.969387755102)
(40,14.4756838905775)
(45,17.2532781228433)
(50,14.4756838905775)
};

\addplot[mark=square*, thin, line width=0.25pt, mark options={fill=red}, mark size=1pt] coordinates {
   (2,32.6147576147576)
(5,36.200555737169)
(10,38.2060502426635)
(15,32.9069049435182)
(20,32.4157002417872)
(25,31.4532453297524)
(30,36.4112285230297)
(35,32.7231121281464)
(40,34.1060337178349)
(45,31.4532453297524)
(50,31.3196033562166)
};

\addplot[mark=pentagon*, thin, line width=0.25pt, mark options={fill=green}, mark size=1.0pt] coordinates {
(2,32.5569358178053)
(5,41.3946922642574)
(10,39.881051011954)
(15,34.1855520116389)
(20,36.4604292421193)
(25,37.4110223799664)
(30,37.3990175780457)
(35,38.7693309616151)
(40,37.3990175780457)
(45,45.1818230079099)
(50,38.7693309616151)
};

\addplot[mark=diamond*, thin, line width=0.25pt, mark options={fill=pink}, mark size=1.0pt] coordinates {
  (2,36.1268193195513)
(5,38.6976638921718)
(10,31.7531179138321)
(15,40.382975932848)
(20,41.1269889530759)
(25,37.3990175780457)
(30,40.0163452108532)
(35,36.6398733406406)
(40,38.494528246081)
(45,41.4699324545871)
(50,41.1269889530759)
};

\addplot[mark=oplus*, thin, line width=0.25pt, mark options={fill=yellow,}, mark size=1.0pt] coordinates {
  (2,23.8758067329495)
(5,39.196334345588)
(10,47.3923943568408)
(15,37.8867588635307)
(20,45.2957704821058)
(25,40.3683120845592)
(30,39.7268204940839)
(35,38.1645285279289)
(40,39.2487545925163)
(45,39.8514719620189)
(50,39.5121435301809)
};

\end{groupplot}
\end{tikzpicture}
\begin{tikzpicture}
\vspace{0.3cm} 
\begin{axis}[
    hide axis,
    xmin=0, xmax=1, ymin=0, ymax=1,
    legend style={at={(0.0,0.0)}, anchor=center, legend columns=7},
    legend entries={\tiny{0.01},\tiny{0.1},\tiny{1},\tiny{2.5},\tiny{5.0},\tiny{10.0},\tiny{50}}
]

\addlegendimage{mark=*,  thin, line width=0.25pt, mark options={fill=orange}, mark size=2.0pt}
\addlegendimage{mark=triangle*, thin, line width=0.25pt, mark options={fill=cyan}, mark size=2.0pt}
\addlegendimage{mark=square*,  thin, line width=0.25pt, mark options={fill=red}, mark size=2pt}
\addlegendimage{mark=pentagon*, thin, line width=0.25pt, mark options={fill=green}, mark size=2.0pt}
\addlegendimage{mark=diamond*, thin,line width=0.25pt, mark options={fill=pink}, mark size=2.0pt}
\addlegendimage{mark=oplus*,  thin,line width=0.25pt, mark options={fill=yellow}, mark size=2.0pt}

\end{axis}
\end{tikzpicture}

\caption{Effect of different prior standard deviation values and the number of samples on the LL--HMC in-distribution performance with a target acceptance of 0.7 and 100 burn-in samples for a single random seed.}
\label{fig:hmc_perf_samples}

\end{figure*}
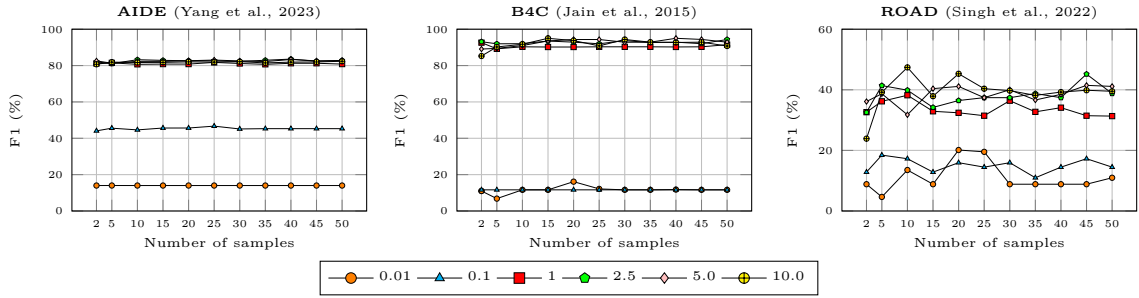

\paragraph{\textbf{Prior scale influence.}}
To investigate the impact of prior scale on LL--HMC, Figure \ref{fig:hmc_perf_samples}  shows the relationship between F1-score and the number of HMC samples, for multiple prior standard deviations. Because the last layer is randomly initialized (cold start) with uninformative priors, smaller prior scales (i.e., stricter priors) tend to degrade performance. For the AIDE and B4C datasets, performance plateaus after 5–10 samples, suggesting that further samples yield negligible performance gains. This implies that LL--HMC achieves stable predictive performance with relatively few samples, which is beneficial for computational efficiency. The ROAD dataset exhibits greater sensitivity to prior scale and the number of samples, with more variation in F1-score across configurations.

\definecolor{steelblue31119180}{RGB}{31,119,180} 


\definecolor{steelblue31119180}{RGB}{31,119,180} 

\begin{figure}[t!]
    \begin{tikzpicture}
    \begin{groupplot}[
        group style={
        group size=3 by 1,
        horizontal sep=1.2cm,
        vertical sep=1.0cm,
    },
    width=5.5cm,
    height=4.0cm,
    grid=major,
    xlabel={Number of samples},
    xtick={1,5,10,15,20,25,30,35,40,45,50},
    tick label style={font=\small},
    label style={font=\small},
    y tick label style={
        /pgf/number format/fixed,
        /pgf/number format/precision=0,
        /pgf/number format/fixed zerofill=true, 
    },
    tick label style={font=\fontsize{4}{4}\selectfont},
    xlabel style={yshift=0.1cm},  
]
    \nextgroupplot[
        title={{\textbf{AIDE} \citep{yang2023aide}}},
        ylabel={\small{F1 (\%)}},
          title style={yshift=-0.2cm},
          ymax=100,
          ymin=0
    ]
    \path [fill=steelblue31119180, fill opacity=0.2]
(axis cs:1,2.4074074074074)
--(axis cs:1,2.4074074074074)
--(axis cs:2,25.474620560273)
--(axis cs:3,36.3674667925979)
--(axis cs:4,42.51325638209)
--(axis cs:5,46.6554748348895)
--(axis cs:6,49.747791819033)
--(axis cs:7,52.3759854381015)
--(axis cs:8,54.4028424588609)
--(axis cs:9,55.9209589301261)
--(axis cs:10,57.2453624582765)
--(axis cs:11,58.5348946957579)
--(axis cs:12,59.560576578204)
--(axis cs:13,60.4014316685476)
--(axis cs:14,61.2272377480546)
--(axis cs:15,62.0077534585458)
--(axis cs:16,62.5394693599207)
--(axis cs:17,63.2542963942399)
--(axis cs:18,63.788592250409)
--(axis cs:19,64.2778232465613)
--(axis cs:20,64.7268921009604)
--(axis cs:21,64.9933289964728)
--(axis cs:22,65.6312297849877)
--(axis cs:23,66.0002645342677)
--(axis cs:24,66.4217424160643)
--(axis cs:25,66.6605724419808)
--(axis cs:26,66.9852816769292)
--(axis cs:27,67.3434718662061)
--(axis cs:28,67.7023653610644)
--(axis cs:29,67.8897825833578)
--(axis cs:30,68.3337000581644)
--(axis cs:31,68.6067687334276)
--(axis cs:32,68.8531898855806)
--(axis cs:33,69.0005682843235)
--(axis cs:34,69.2553123777573)
--(axis cs:35,69.3835844541345)
--(axis cs:36,69.5052932853387)
--(axis cs:37,69.6413262263697)
--(axis cs:38,69.9221789807699)
--(axis cs:39,70.1410243390448)
--(axis cs:40,70.300281447589)
--(axis cs:41,70.4300286526485)
--(axis cs:42,70.5788842775389)
--(axis cs:43,70.6350495479844)
--(axis cs:44,71.085236983809)
--(axis cs:45,71.0570364024075)
--(axis cs:46,71.2729827035557)
--(axis cs:47,71.3526935006528)
--(axis cs:48,71.3552715499531)
--(axis cs:49,71.4340747546225)
--(axis cs:50,71.4354491242607)
--(axis cs:50,94.1786508585519)
--(axis cs:50,94.1786508585519)
--(axis cs:49,94.4073920109395)
--(axis cs:48,94.5650672343655)
--(axis cs:47,94.8060944357168)
--(axis cs:46,94.9769607166573)
--(axis cs:45,95.0195435497572)
--(axis cs:44,95.3157190345896)
--(axis cs:43,95.1412055075163)
--(axis cs:42,95.3730772181638)
--(axis cs:41,95.5219328430542)
--(axis cs:40,95.7008795433275)
--(axis cs:39,95.8616812758951)
--(axis cs:38,95.9753495283022)
--(axis cs:37,96.040852880551)
--(axis cs:36,96.2665274286223)
--(axis cs:35,96.5212919940758)
--(axis cs:34,96.7851925727096)
--(axis cs:33,96.93919918045)
--(axis cs:32,97.2200153268321)
--(axis cs:31,97.4212690202365)
--(axis cs:30,97.6181646755939)
--(axis cs:29,97.6685872019126)
--(axis cs:28,98.004412171389)
--(axis cs:27,98.1960607900218)
--(axis cs:26,98.4209996436117)
--(axis cs:25,98.7148563481743)
--(axis cs:24,99.133124270034)
--(axis cs:23,99.4089550241137)
--(axis cs:22,99.7855739501644)
--(axis cs:21,99.9452464078881)
--(axis cs:20,100.540954174571)
--(axis cs:19,101.016770640648)
--(axis cs:18,101.527592003416)
--(axis cs:17,102.07948275049)
--(axis cs:16,102.550316303429)
--(axis cs:15,103.324530418874)
--(axis cs:14,103.981307650733)
--(axis cs:13,104.756382415227)
--(axis cs:12,105.712538604354)
--(axis cs:11,106.718015323327)
--(axis cs:10,107.752898669926)
--(axis cs:9,109.136768768901)
--(axis cs:8,110.811668475449)
--(axis cs:7,112.621760156812)
--(axis cs:6,114.750741388895)
--(axis cs:5,117.80588027108)
--(axis cs:4,121.963209752486)
--(axis cs:3,127.876375982594)
--(axis cs:2,136.852978239871)
--(axis cs:1,2.4074074074074)
--cycle;

\addplot [semithick, steelblue31119180]
table {%
1 2.4074074074074
2 81.1637994000723
3 82.1219213875962
4 82.238233067288
5 82.2306775529845
6 82.2492666039642
7 82.4988727974568
8 82.6072554671548
9 82.5288638495136
10 82.4991305641013
11 82.6264550095427
12 82.6365575912792
13 82.5789070418873
14 82.604272699394
15 82.6661419387101
16 82.5448928316749
17 82.6668895723652
18 82.6580921269123
19 82.6472969436046
20 82.6339231377655
21 82.4692877021805
22 82.708401867576
23 82.7046097791907
24 82.7774333430492
25 82.6877143950775
26 82.7031406602705
27 82.769766328114
28 82.8533887662267
29 82.7791848926352
30 82.9759323668791
31 83.014018876832
32 83.0366026062064
33 82.9698837323867
34 83.0202524752334
35 82.9524382241052
36 82.8859103569805
37 82.8410895534604
38 82.948764254536
39 83.0013528074699
40 83.0005804954583
41 82.9759807478514
42 82.9759807478514
43 82.8881275277503
44 83.2004780091993
45 83.0382899760824
46 83.1249717101065
47 83.0793939681848
48 82.9601693921593
49 82.920733382781
50 82.8070499914063
};

    \nextgroupplot[
        title={{\textbf{B4C} \citep{jain2015car}}},
        ylabel={\small{F1 (\%)}},
          title style={yshift=-0.2cm},
          ymin=0,
          ymax=100
    ]
\path [fill=steelblue31119180, fill opacity=0.2]
(axis cs:1,11.6363636363636)
--(axis cs:1,11.6363636363636)
--(axis cs:2,33.2313373683485)
--(axis cs:3,45.2246665353119)
--(axis cs:4,52.042157305909)
--(axis cs:5,55.8944647155769)
--(axis cs:6,59.3894499308765)
--(axis cs:7,61.5966515078833)
--(axis cs:8,64.5848047719954)
--(axis cs:9,65.8611485395994)
--(axis cs:10,66.5052033211166)
--(axis cs:11,67.7210659063231)
--(axis cs:12,69.081696920578)
--(axis cs:13,70.0248797370222)
--(axis cs:14,70.9859630677586)
--(axis cs:15,71.3825239118339)
--(axis cs:16,72.5432304063399)
--(axis cs:17,73.4558336272192)
--(axis cs:18,73.9464435736174)
--(axis cs:19,74.5250671931756)
--(axis cs:20,74.9495867623111)
--(axis cs:21,75.5987525582482)
--(axis cs:22,75.838532064086)
--(axis cs:23,76.3104282366841)
--(axis cs:24,76.8770199961402)
--(axis cs:25,76.9157529537468)
--(axis cs:26,77.7018668086424)
--(axis cs:27,77.7150922528999)
--(axis cs:28,78.1266069263336)
--(axis cs:29,78.4636466636392)
--(axis cs:30,78.7042764853169)
--(axis cs:31,79.0148008006084)
--(axis cs:32,78.9074912541829)
--(axis cs:33,79.3249514693347)
--(axis cs:34,79.3629089965501)
--(axis cs:35,80.0455279788005)
--(axis cs:36,79.840823848513)
--(axis cs:37,80.3325944151017)
--(axis cs:38,80.3764324238401)
--(axis cs:39,80.6641448835065)
--(axis cs:40,80.776560973099)
--(axis cs:41,80.8457035189497)
--(axis cs:42,81.0764746662266)
--(axis cs:43,81.4398079136612)
--(axis cs:44,81.3706820201598)
--(axis cs:45,81.9669509098113)
--(axis cs:46,81.6148925944012)
--(axis cs:47,81.8247957557394)
--(axis cs:48,81.6396921468075)
--(axis cs:49,81.7369481942302)
--(axis cs:50,82.1366349951082)
--(axis cs:50,105.385915559779)
--(axis cs:50,105.385915559779)
--(axis cs:49,105.21394235306)
--(axis cs:48,105.353723185802)
--(axis cs:47,105.782852517169)
--(axis cs:46,105.821317896431)
--(axis cs:45,106.430512374853)
--(axis cs:44,106.094330764408)
--(axis cs:43,106.438015869588)
--(axis cs:42,106.355641147629)
--(axis cs:41,106.41903074269)
--(axis cs:40,106.655338991235)
--(axis cs:39,106.858202699132)
--(axis cs:38,106.897199697058)
--(axis cs:37,107.194579490296)
--(axis cs:36,107.055163814962)
--(axis cs:35,107.630860061592)
--(axis cs:34,107.329002433587)
--(axis cs:33,107.696083684298)
--(axis cs:32,107.698520919979)
--(axis cs:31,108.248868288834)
--(axis cs:30,108.395513100277)
--(axis cs:29,108.636142921955)
--(axis cs:28,108.806325765594)
--(axis cs:27,108.929548743278)
--(axis cs:26,109.481582940541)
--(axis cs:25,109.287554271572)
--(axis cs:24,109.887528564127)
--(axis cs:23,109.991327117343)
--(axis cs:22,110.238550729354)
--(axis cs:21,110.768829610754)
--(axis cs:20,110.936355158095)
--(axis cs:19,111.400172166044)
--(axis cs:18,111.777903190313)
--(axis cs:17,112.326983877402)
--(axis cs:16,112.541953393202)
--(axis cs:15,112.628025986446)
--(axis cs:14,113.622001019176)
--(axis cs:13,114.191465600902)
--(axis cs:12,114.969112301717)
--(axis cs:11,115.552305306608)
--(axis cs:10,116.584606203721)
--(axis cs:9,118.54222554112)
--(axis cs:8,120.241132877498)
--(axis cs:7,120.74573289365)
--(axis cs:6,122.997534773672)
--(axis cs:5,125.106823462113)
--(axis cs:4,128.805248708613)
--(axis cs:3,132.504587547441)
--(axis cs:2,137.501094294046)
--(axis cs:1,11.6363636363636)
--cycle;

\addplot [semithick, steelblue31119180]
table {%
1 11.6363636363636
2 85.3662158311972
3 88.8646270413765
4 90.4237030072611
5 90.5006440888451
6 91.1934923522742
7 91.1711922007669
8 92.4129688247467
9 92.2016870403596
10 91.544904762419
11 91.6366856064658
12 92.0254046111473
13 92.1081726689622
14 92.3039820434675
15 92.00527494914
16 92.5425918997708
17 92.8914087523107
18 92.8621733819652
19 92.9626196796096
20 92.9429709602032
21 93.1837910845011
22 93.0385413967201
23 93.1508776770136
24 93.3822742801337
25 93.1016536126595
26 93.5917248745918
27 93.3223204980889
28 93.4664663459637
29 93.5498947927971
30 93.5498947927971
31 93.6318345447213
32 93.3030060870809
33 93.5105175768162
34 93.3459557150687
35 93.8381940201965
36 93.4479938317373
37 93.7635869526986
38 93.6368160604492
39 93.7611737913194
40 93.715949982167
41 93.6323671308198
42 93.716057906928
43 93.9389118916245
44 93.7325063922837
45 94.1987316423322
46 93.718105245416
47 93.803824136454
48 93.4967076663045
49 93.4754452736453
50 93.7612752774434
};

\nextgroupplot[
        title={{\textbf{ROAD} \citep{singh2022road}}},
        ylabel={\small{F1 (\%)}},
          title style={yshift=-0.2cm},
          ymax=60,
          ymin=0
    ]
\path [fill=steelblue31119180, fill opacity=0.2]
(axis cs:1,4.64285714285714)
--(axis cs:1,4.64285714285714)
--(axis cs:2,13.7444005665719)
--(axis cs:3,17.7905619071176)
--(axis cs:4,19.8908775703713)
--(axis cs:5,20.0353509975758)
--(axis cs:6,21.244676573495)
--(axis cs:7,22.6726533959784)
--(axis cs:8,23.1457818409577)
--(axis cs:9,23.7296709742352)
--(axis cs:10,24.5079919979849)
--(axis cs:11,24.9780474068055)
--(axis cs:12,25.7144547609977)
--(axis cs:13,25.8248159860842)
--(axis cs:14,27.5208373031441)
--(axis cs:15,27.6721986791789)
--(axis cs:16,27.8843029851824)
--(axis cs:17,28.367417821544)
--(axis cs:18,28.6858842086104)
--(axis cs:19,28.6619933248124)
--(axis cs:20,28.8195138523721)
--(axis cs:21,29.1022618506936)
--(axis cs:22,28.4764867004226)
--(axis cs:23,28.3634034525499)
--(axis cs:24,28.4987243284675)
--(axis cs:25,28.6579351109257)
--(axis cs:26,28.7784635847482)
--(axis cs:27,28.8466558797843)
--(axis cs:28,28.9409754815265)
--(axis cs:29,28.828071976118)
--(axis cs:30,28.8861408660453)
--(axis cs:31,28.5756777864016)
--(axis cs:32,28.9299866051721)
--(axis cs:33,29.0413463529665)
--(axis cs:34,29.1223091022094)
--(axis cs:35,29.239503867429)
--(axis cs:36,29.3453299291526)
--(axis cs:37,29.4545014171028)
--(axis cs:38,29.4793736986124)
--(axis cs:39,29.390857157841)
--(axis cs:40,29.4070263695667)
--(axis cs:41,29.5062655170007)
--(axis cs:42,29.8293144138247)
--(axis cs:43,29.8840078379749)
--(axis cs:44,29.7823280779212)
--(axis cs:45,29.5250653096515)
--(axis cs:46,29.8133747473016)
--(axis cs:47,29.9106720548669)
--(axis cs:48,29.9580799256032)
--(axis cs:49,30.0385585583976)
--(axis cs:50,30.2242211935527)
--(axis cs:50,39.5413827066097)
--(axis cs:50,39.5413827066097)
--(axis cs:49,39.4424443308492)
--(axis cs:48,39.4531055629992)
--(axis cs:47,39.5005134337355)
--(axis cs:46,39.5008944673221)
--(axis cs:45,39.3139648583047)
--(axis cs:44,39.6756760500227)
--(axis cs:43,39.8853884524678)
--(axis cs:42,39.940081876618)
--(axis cs:41,39.7300245354178)
--(axis cs:40,39.7498986034867)
--(axis cs:39,39.8571699837793)
--(axis cs:38,40.0731598609598)
--(axis cs:37,40.1847467422023)
--(axis cs:36,40.2180472108852)
--(axis cs:35,40.2603812091167)
--(axis cs:34,40.2976355558866)
--(axis cs:33,40.3785983051295)
--(axis cs:32,40.4363714008718)
--(axis cs:31,40.2597717749872)
--(axis cs:30,40.7605836944945)
--(axis cs:29,40.8974417198148)
--(axis cs:28,41.2155716618331)
--(axis cs:27,41.3389469536479)
--(axis cs:26,41.5003122518907)
--(axis cs:25,41.6208407257132)
--(axis cs:24,41.7167384369853)
--(axis cs:23,41.852059312903)
--(axis cs:22,42.2528224243902)
--(axis cs:21,43.1789833334106)
--(axis cs:20,43.1835316700738)
--(axis cs:19,43.3410521976335)
--(axis cs:18,43.6991322840418)
--(axis cs:17,43.7145477149107)
--(axis cs:16,43.6018761508319)
--(axis cs:15,43.8139804568354)
--(axis cs:14,44.1157851115632)
--(axis cs:13,42.8891815206056)
--(axis cs:12,43.4785505104141)
--(axis cs:11,43.5253937402596)
--(axis cs:10,43.9764283868407)
--(axis cs:9,44.2724539282949)
--(axis cs:8,44.9676334589388)
--(axis cs:7,46.0468979469108)
--(axis cs:6,46.5309765623381)
--(axis cs:5,47.7403021382574)
--(axis cs:4,50.8756180563043)
--(axis cs:3,53.6838854445184)
--(axis cs:2,57.6905397106913)
--(axis cs:1,4.64285714285714)
--cycle;

\addplot [semithick, steelblue31119180]
table {%
1 4.64285714285714
2 35.7174701386316
3 35.737223675818
4 35.3832478133378
5 33.8878265679166
6 33.8878265679166
7 34.3597756714446
8 34.0567076499482
9 34.001062451265
10 34.2422101924128
11 34.2517205735326
12 34.5965026357059
13 34.3569987533449
14 35.8183112073537
15 35.7430895680072
16 35.7430895680072
17 36.0409827682274
18 36.1925082463261
19 36.001522761223
20 36.001522761223
21 36.1406225920521
22 35.3646545624064
23 35.1077313827264
24 35.1077313827264
25 35.1393879183194
26 35.1393879183194
27 35.0928014167161
28 35.0782735716798
29 34.8627568479664
30 34.8233622802699
31 34.4177247806944
32 34.6831790030219
33 34.709972329048
34 34.709972329048
35 34.7499425382729
36 34.7816885700189
37 34.8196240796525
38 34.7762667797861
39 34.6240135708102
40 34.5784624865267
41 34.6181450262092
42 34.8846981452214
43 34.8846981452214
44 34.729002063972
45 34.4195150839781
46 34.6571346073119
47 34.7055927443012
48 34.7055927443012
49 34.7405014446234
50 34.8828019500812
};
    \end{groupplot}
    \end{tikzpicture}
    \caption{Effects of additional dependent samples from the same chain for the top-performing LL--HMC configuration across 100 random seeds. The light blue area illustrates the standard deviation of the cummulative F1-score for the  additional samples.}
    \label{fig:dependent_samples}
\end{figure}
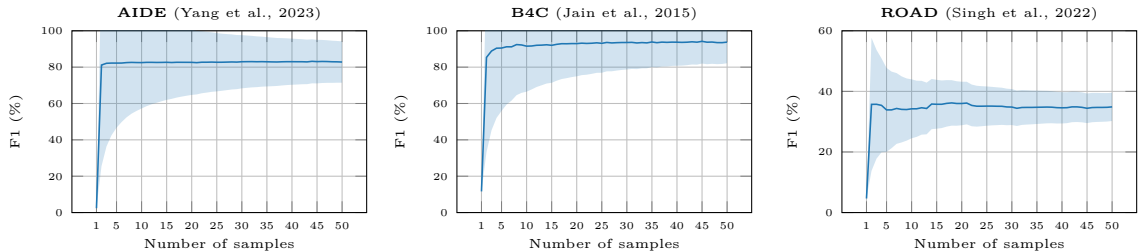

\paragraph{\textbf{Influence of additional dependent LL--HMC samples.}}
To examine \textcolor{black}{effect of the additional $\theta_{LL}$ parameter samples} from the same LL--HMC chain\textcolor{black}{, which increases the computational complexity, on the} DAR and DIR performance, Figure \ref{fig:dependent_samples} presents the average F1 score and the standard deviation as more samples are cumulatively drawn across 100 random seeds.  The results show that standard deviation decreases with more samples, but performance converges quickly. This suggests that beyond a small number of posterior samples, further samples from the LL--HMC chain yield diminishing returns in classification performance.



\definecolor{darkgray176}{RGB}{176,176,176}
\definecolor{steelblue31119180}{RGB}{31,119,180}

\begin{figure}[t]
    \centering
    \begin{minipage}[t]{0.32\linewidth}
        \centering
        \includegraphics[width=0.99\linewidth, keepaspectratio]{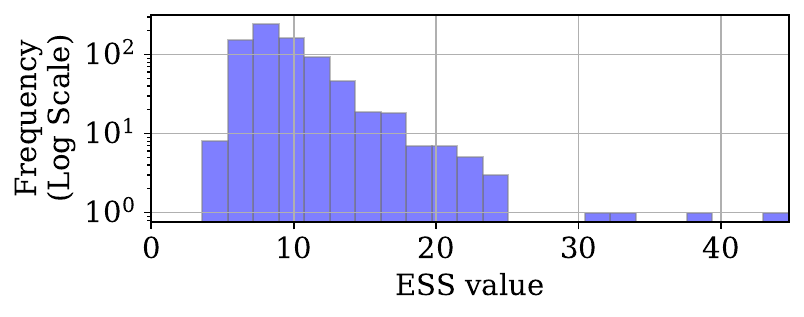}
              \captionsetup{font=tiny}
        \caption*{ESS per last layer parameter for the AIDE dataset with a single chain for 50 samples.}
    \end{minipage}
    \hfill
    \begin{minipage}[t]{0.32\linewidth}
        \centering
        \includegraphics[width=0.99\linewidth, keepaspectratio]{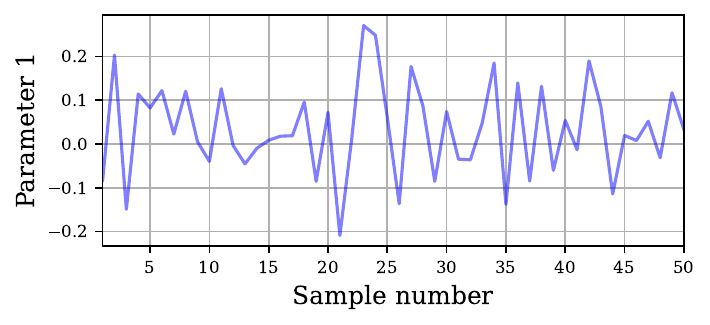}
              \captionsetup{font=tiny}
        \caption*{Value for the first weight parameter of the last layer for each sample.}
    \end{minipage}
    \hfill
    \begin{minipage}[t]{0.32\linewidth}
        \centering
        \includegraphics[width=0.99\linewidth, keepaspectratio]{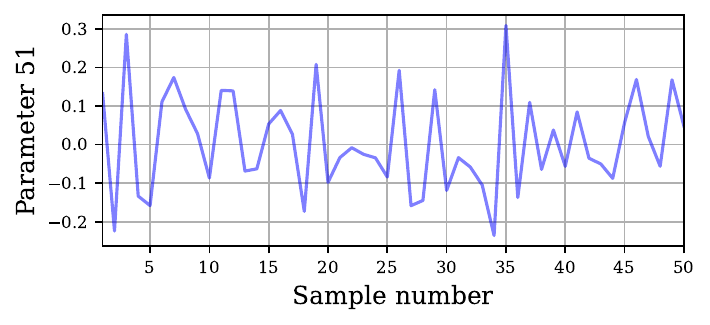}
              \captionsetup{font=tiny}
        \caption*{Value for the fifty-first weight parameter of the last layer for each sample.}
    \end{minipage}

    \caption{Visualization of the ESS and the parameter values for 50 samples for LL--HMC for the AIDE dataset with a single chain, target acceptance of 0.7, prior standard deviation of 1, and 100 burn in samples.}
    \label{fig:ess}
\end{figure}

\begin{figure}[t!]
    \centering
    \begin{minipage}[t]{0.32\linewidth}
        \centering
        \includegraphics[width=0.99\linewidth, keepaspectratio]{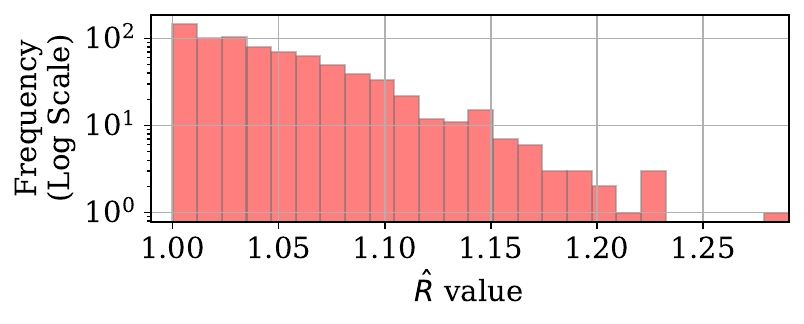}
        \vspace{-0.6cm}
              \captionsetup{font=tiny}
        \caption*{AIDE - Two Chains, One Starting Position.}
        \label{fig:nuscenes_rhat_aide1}
    \end{minipage}
    \hfill
    \begin{minipage}[t]{0.32\linewidth}
        \centering
        \includegraphics[width=0.99\linewidth, keepaspectratio]{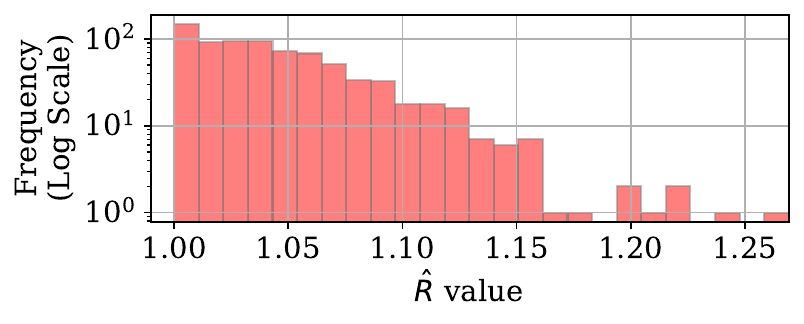}
        \vspace{-0.6cm}
              \captionsetup{font=tiny}
        \caption*{B4C - Two Chains, One Starting Position.}
        \label{fig:nuscenes_rhat_b4c1}
    \end{minipage}
    \hfill
    \begin{minipage}[t]{0.32\linewidth}
        \centering
        \includegraphics[width=0.99\linewidth, keepaspectratio]{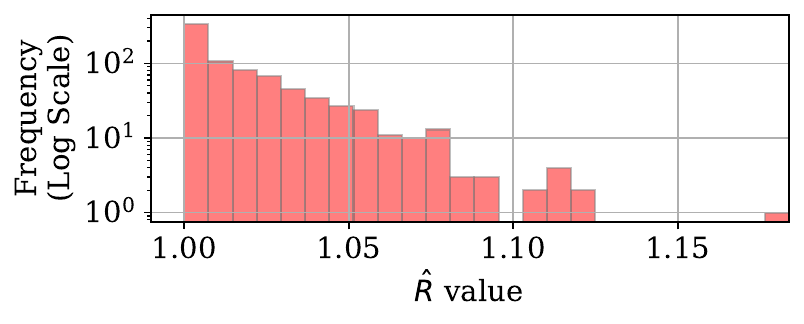}
        \vspace{-0.6cm}
              \captionsetup{font=tiny}
        \caption*{ROAD - Two Chains, One Starting Position.}
        \label{fig:nuscenes_rhat_road1}
    \end{minipage}
        \vskip\baselineskip
    \begin{minipage}[t]{0.32\linewidth}
        \centering
        \includegraphics[width=0.99\linewidth, keepaspectratio]{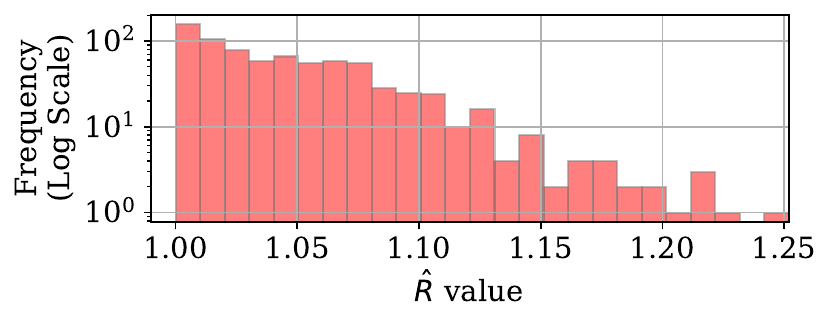}
        \vspace{-0.6cm}
              \captionsetup{font=tiny}
        \caption*{AIDE - Two Chains, Two Starting Positions.}
        \label{fig:nuscenes_rhat_aide2}
    \end{minipage}
    \hfill
    \begin{minipage}[t]{0.32\linewidth}
        \centering
        \includegraphics[width=0.99\linewidth, keepaspectratio]{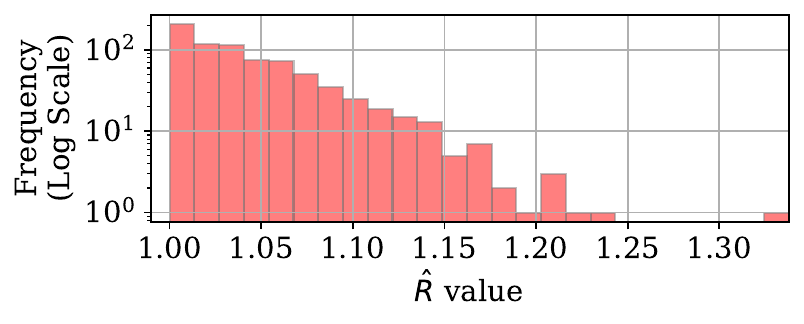}
        \vspace{-0.6cm}
              \captionsetup{font=tiny}
        \caption*{B4C - Two Chains, Two Starting Positions.}
        \label{fig:nuscenes_rhat_b4c2}
    \end{minipage}
    \hfill
    \begin{minipage}[t]{0.32\linewidth}
        \centering
        \includegraphics[width=0.99\linewidth, keepaspectratio]{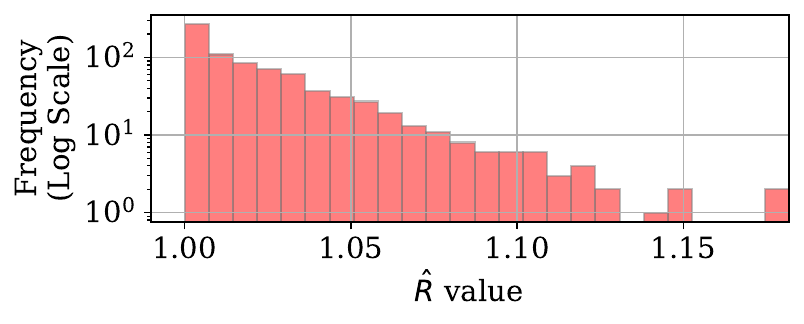}
        \vspace{-0.6cm}
              \captionsetup{font=tiny}
        \caption*{ROAD - Two Chains, Two Starting Positions.}
        \label{fig:nuscenes_rhat_road2}
    \end{minipage}
    
    \caption{Log-scale histograms of the $\hat{R}$ values for two chains for single starting position (top) or two starting positions (bottom), for 100 burn-in samples, target acceptance of 0.7, 50 samples and a prior scale of 1.}
    \label{fig:rhats}
\end{figure}

\begin{table*}[t!]
\centering
\caption{Average top-performance and two standard errors of the mean across five random seeds for LL--HMC with multiple starting positions or chains. S=Starting Positions, C=Chains.}
\resizebox{\linewidth}{!}{
\begin{tabular}{lcc|c|c|c|c||c|c|c|c||c|c|c|c}
\toprule
&  &&\multicolumn{4}{c||}{\textbf{AIDE} \citep{yang2023aide}} & \multicolumn{4}{c||}{\textbf{B4C} \citep{jain2015car}} & \multicolumn{4}{c}{\textbf{ROAD} \citep{singh2022road}} \\
& &&\textbf{Acc ($\uparrow$)} & \textbf{F1 ($\uparrow$)} & \textbf{ACE ($\downarrow$)} & \textbf{rAULC ($\uparrow$)} 
& \textbf{Acc ($\uparrow$)} & \textbf{F1 ($\uparrow$)} & \textbf{ACE ($\downarrow$)} & \textbf{rAULC ($\uparrow$)}
& \textbf{Acc ($\uparrow$)} & \textbf{F1 ($\uparrow$)} & \textbf{ACE ($\downarrow$)} & \textbf{rAULC ($\uparrow$)} \\
\midrule
\multirow{4}{*}{\textbf{LL--HMC}} & \multirow{2}{*}{\small{S=1}} & \small{C=1}   
& 89.00$\pm$0.16 & 83.87$\pm$0.08 & 0.029$\pm$0.002 & 0.68$\pm$0.06
& 93.73$\pm$0.42 & 94.78$\pm$0.29 & 0.027$\pm$0.007 & 0.66$\pm$0.10
& 70.15$\pm$5.05 & 54.18$\pm$4.43 & 0.073$\pm$0.014 & 0.41$\pm$0.12 \\
& & \small{C=2} 
& 88.34$\pm$0.73 & 82.98$\pm$0.25 & 0.045$\pm$0.017 & 0.66$\pm$0.06
& 93.59$\pm$0.42 & 94.74$\pm$0.32 & 0.032$\pm$0.011 & 0.61$\pm$0.03
& 68.16$\pm$5.12 & 53.51$\pm$0.93 & 0.077$\pm$0.017 & 0.16$\pm$0.09 \\
\cmidrule{2-15}
& \multirow{2}{*}{\small{S=2}} & \small{C=1} 
& 89.00$\pm$0.17 & 83.66$\pm$0.32 & 0.029$\pm$0.002 & 0.70$\pm$0.03
& 94.02$\pm$0.11 & 95.00$\pm$0.07 & 0.023$\pm$0.001 & 0.70$\pm$0.02
& 69.16$\pm$4.66 & 52.67$\pm$2.45 & 0.072$\pm$0.015 & 0.57$\pm$0.07 \\
& & \small{C=2} 
& 88.15$\pm$0.46 & 82.56$\pm$0.60 & 0.042$\pm$0.007 & 0.60$\pm$0.03
& 93.59$\pm$0.42 & 94.72$\pm$0.30 & 0.047$\pm$0.019 & 0.52$\pm$0.07
& 70.90$\pm$4.76 & 53.79$\pm$3.05 & 0.081$\pm$0.017 & 0.27$\pm$0.19 \\
\bottomrule
\end{tabular}
}
\label{tab:performance_multi_hmc}
\end{table*}

\paragraph{\textbf{Effective sample size and parameter trajectories.}}Figure \ref{fig:ess} includes a histogram of effective sample size (ESS) values across parameters (x-axis: ESS; y-axis: number of parameters), along with two example parameter trajectories. Consistent with findings in Vellenga et al. \cite{vellenga2024pthmc}, LL--HMC yields a low ESS for most parameters, indicating effective sampling only for a small subset. The example parameter trajectories show no discernible trend or drift and fluctuate around zero, suggesting limited exploration. 

\paragraph{\textbf{Multiple chains and starting positions.}}
Table \ref{tab:performance_multi_hmc} shows the average top-performing configurations across five random seeds for multiple starting (S) positions and multiple chains (C). The results show that increasing the number of chains or adding an additional starting position does not consistently lead to improved performance. When using multiple chains, it is also important to assess whether the chains have explored distinct regions of the posterior and converged to the target distribution. Figure \ref{fig:rhats} shows the $\hat{R}$ values for the scenarios with multiple chains for both  single and multiple starting positions. For all datasets, the $\hat{R}$ values are close to 1 and mostly below 1.1. This indicates that the chains have likely converged to the target distribution but also potentially sample from the same region \citep{izmailov2021bayesian}. %

\begin{table*}[t]
\centering
\caption{Average uncertainty-based OOD detection performance and two standard errors of the mean of the top-performing hyperparameter configurations per random seed. ROC-AUC=Receiver Operating Characteristic - Area Under the Curve, PR=Precision-Recall - Area Under the Curve. Best results are highlighted in \colorbox{best}{Blue}, and the second best in \colorbox{2best}{Red}.} 
\resizebox{\textwidth }{!}{
\begin{tabular}{l|c|c|c|c|c|c||c|c|c|c|c|c||c|c|c}
\toprule
\multicolumn{1}{c|}{} & \multicolumn{6}{c||}{\textbf{AIDE} \citep{yang2023aide}} & \multicolumn{6}{c||}{\textbf{B4C} \citep{jain2015car}} & \multicolumn{3}{c}{\textbf{ROAD} \citep{singh2022road}}  \\ \cmidrule{2-16}
\multicolumn{1}{c|}{} & \multicolumn{3}{c|}{\textbf{OOD min}} & \multicolumn{3}{c||}{\textbf{OOD max}} & \multicolumn{3}{c|}{\textbf{OOD min}} & \multicolumn{3}{c||}{\textbf{OOD max}} & \multicolumn{3}{c}{\textbf{OOD min}} \\
 \multicolumn{1}{c|}{} & \begin{tabular}{@{}c@{}}\textbf{ROC-}\\ \textbf{AUC} ($\uparrow$)\end{tabular} & 
\begin{tabular}{@{}c@{}}\textbf{PR-}\\\textbf{AUC} ($\uparrow$)\end{tabular} & 
\begin{tabular}{@{}c@{}}\textbf{FPR95}\\($\downarrow$)\end{tabular} & \begin{tabular}{@{}c@{}}\textbf{ROC-}\\ \textbf{AUC} ($\uparrow$)\end{tabular} & 
\begin{tabular}{@{}c@{}}\textbf{PR-}\\\textbf{AUC} ($\uparrow$)\end{tabular} & 
\begin{tabular}{@{}c@{}}\textbf{FPR95}\\($\downarrow$)\end{tabular} & \begin{tabular}{@{}c@{}}\textbf{ROC-}\\ \textbf{AUC} ($\uparrow$)\end{tabular} & 
\begin{tabular}{@{}c@{}}\textbf{PR-}\\\textbf{AUC} ($\uparrow$)\end{tabular} & 
\begin{tabular}{@{}c@{}}\textbf{FPR95}\\($\downarrow$)\end{tabular} & \begin{tabular}{@{}c@{}}\textbf{ROC-}\\ \textbf{AUC} ($\uparrow$)\end{tabular} & 
\begin{tabular}{@{}c@{}}\textbf{PR-}\\\textbf{AUC} ($\uparrow$)\end{tabular} & 
\begin{tabular}{@{}c@{}}\textbf{FPR95}\\($\downarrow$)\end{tabular} & \begin{tabular}{@{}c@{}}\textbf{ROC-}\\ \textbf{AUC} ($\uparrow$)\end{tabular} & 
\begin{tabular}{@{}c@{}}\textbf{PR-}\\\textbf{AUC} ($\uparrow$)\end{tabular} & 
\begin{tabular}{@{}c@{}}\textbf{FPR95}\\($\downarrow$)\end{tabular}   \\
\midrule
\textbf{Regular} &   0.55 & 0.19 & 0.91 & 0.82 & 0.94 & 0.41 & 0.51 & 0.37 & 0.90 & 0.64 & 0.82 & 0.84  & 0.19 & 0.25 & 1.00 \\
\textbf{DE (N=5)} &  0.69 & 0.28 & 0.85 & 0.84 & 0.94 &0.35 & 0.52 & 0.39 & 0.90 &  0.79 &  0.90 & \cellcolor{2best} 0.45 & 0.55& 0.41 & 0.88   \\
\textbf{DDU} &  0.73 & 0.31 & 0.49 & 0.80 & 0.90 & 0.45 & 0.63 & 0.44 &0.87 & 0.48 & 0.75 & 0.96 & 0.63 & 0.44 & 0.87  \\  \midrule
\textbf{BBB--LL} & \cellcolor{best} 0.95$\pm$0.02 & \cellcolor{best}0.72$\pm$0.08 & \cellcolor{best} 0.10$\pm$0.05	& \cellcolor{best} 0.96$\pm$0.01 & \cellcolor{best} 0.99$\pm$0.01 & \cellcolor{2best} 0.22$\pm$0.06	& \cellcolor{2best} 0.84$\pm$0.03 & \cellcolor{2best} 0.70$\pm$0.04 & 0.48$\pm$0.21	& \cellcolor{2best} 0.80$\pm$0.02 & \cellcolor{2best} 0.92$\pm$0.01 &  0.69$\pm$0.10 & \cellcolor{2best}	0.60$\pm$0.10 & \cellcolor{2best} 0.49$\pm$0.13 & 0.85$\pm$0.06  \\

\textbf{PE--LL} &  0.69$\pm$0.01 & 0.26$\pm$0.01 & 0.68$\pm$0.02	& 0.93$\pm$0.01 & 0.96$\pm$0.01 & 0.41$\pm$0.01 &	0.74$\pm$0.04 & 0.55$\pm$0.04 & 0.48$\pm$0.03 &	0.72$\pm$0.01 & 0.89$\pm$0.02 & 0.82$\pm$0.02 &	0.44$\pm$0.04 & 0.33$\pm$0.01 & 0.89$\pm$0.04 \\
\textbf{SE} &  0.59$\pm$0.01 & 0.21$\pm$0.01 & 0.89$\pm$0.01 & 	0.88$\pm$0.01 & 0.96$\pm$0.01 & 0.41$\pm$0.01	& 0.72$\pm$0.02 & 0.54$\pm$0.01 & 0.50$\pm$0.02 &	0.71$\pm$0.01 & 0.88$\pm$0.01 & 0.79$\pm$0.01 &	0.45$\pm$0.05 & 0.33$\pm$0.02 & 0.86$\pm$0.10 \\

\textbf{VBLL} & 0.66$\pm$0.04 & 0.25$\pm$0.03 & 0.79$\pm$0.08	&0.89$\pm$0.01 & \cellcolor{2best} 0.97$\pm$0.01 & 0.42$\pm$0.04	& 0.82$\pm$0.05 & 0.64$\pm$0.07 & \cellcolor{2best} 0.41$\pm$0.09 &	0.65$\pm$0.02 & 0.84$\pm$0.02 & 0.88$\pm$0.06 &	0.57$\pm$0.09 & 0.44$\pm$0.11 & \cellcolor{2best} 0.80$\pm$0.05 \\
 


\textbf{LL--HMC} & \cellcolor{2best}0.91$\pm$0.03 & \cellcolor{2best} 0.58$\pm$0.10 & \cellcolor{2best} 0.18$\pm$0.05 & \cellcolor{2best}	0.95$\pm$0.01 & \cellcolor{best}0.99$\pm$0.01 &  \cellcolor{best} 0.14$\pm$0.01	&   \cellcolor{best}  0.85$\pm$0.04 & \cellcolor{best}   0.73$\pm$0.05 &  \cellcolor{best}  0.36$\pm$0.08&  \cellcolor{best}  0.91$\pm$0.01 &  \cellcolor{best}  0.97$\pm$0.01 &  \cellcolor{best}  0.39$\pm$0.05&  \cellcolor{best}  0.87$\pm$0.01 &  \cellcolor{best}  0.72$\pm$0.01 &  \cellcolor{best}  0.31$\pm$0.04 \\ 


\bottomrule
\end{tabular}
}
 \label{tab:road_ood_exp}
\end{table*}

\begin{table*}[t]
\centering
\caption{Average uncertainty-based OOD detection grid search results and two standard errors of the mean for the best average hyperparameter configuration across five random seeds.} 
\resizebox{\textwidth }{!}{
\begin{tabular}{l|c|c|c|c|c|c||c|c|c|c|c|c||c|c|c}
\toprule
\multicolumn{1}{c|}{} & \multicolumn{6}{c||}{\textbf{AIDE} \citep{yang2023aide}} & \multicolumn{6}{c||}{\textbf{B4C} \citep{jain2015car}} & \multicolumn{3}{c}{\textbf{ROAD} \citep{singh2022road}}  \\ \cmidrule{2-16}
\multicolumn{1}{c|}{} & \multicolumn{3}{c|}{\textbf{OOD min}} & \multicolumn{3}{c||}{\textbf{OOD max}} & \multicolumn{3}{c|}{\textbf{OOD min}} & \multicolumn{3}{c||}{\textbf{OOD max}} & \multicolumn{3}{c}{\textbf{OOD min}} \\
 \multicolumn{1}{c|}{} & \begin{tabular}{@{}c@{}}\textbf{ROC-}\\ \textbf{AUC} ($\uparrow$)\end{tabular} & 
\begin{tabular}{@{}c@{}}\textbf{PR-}\\\textbf{AUC} ($\uparrow$)\end{tabular} & 
\begin{tabular}{@{}c@{}}\textbf{FPR95}\\($\downarrow$)\end{tabular} & \begin{tabular}{@{}c@{}}\textbf{ROC-}\\ \textbf{AUC} ($\uparrow$)\end{tabular} & 
\begin{tabular}{@{}c@{}}\textbf{PR-}\\\textbf{AUC} ($\uparrow$)\end{tabular} & 
\begin{tabular}{@{}c@{}}\textbf{FPR95}\\($\downarrow$)\end{tabular} & \begin{tabular}{@{}c@{}}\textbf{ROC-}\\ \textbf{AUC} ($\uparrow$)\end{tabular} & 
\begin{tabular}{@{}c@{}}\textbf{PR-}\\\textbf{AUC} ($\uparrow$)\end{tabular} & 
\begin{tabular}{@{}c@{}}\textbf{FPR95}\\($\downarrow$)\end{tabular} & \begin{tabular}{@{}c@{}}\textbf{ROC-}\\ \textbf{AUC} ($\uparrow$)\end{tabular} & 
\begin{tabular}{@{}c@{}}\textbf{PR-}\\\textbf{AUC} ($\uparrow$)\end{tabular} & 
\begin{tabular}{@{}c@{}}\textbf{FPR95}\\($\downarrow$)\end{tabular} & \begin{tabular}{@{}c@{}}\textbf{ROC-}\\ \textbf{AUC} ($\uparrow$)\end{tabular} & 
\begin{tabular}{@{}c@{}}\textbf{PR-}\\\textbf{AUC} ($\uparrow$)\end{tabular} & 
\begin{tabular}{@{}c@{}}\textbf{FPR95}\\($\downarrow$)\end{tabular}   \\
\midrule
\textbf{BBB--LL} &\cellcolor{best} 0.55$\pm$0.11 & \cellcolor{best} 0.20$\pm$0.04 & \cellcolor{best} 0.84$\pm$0.15	&	0.80$\pm$0.06 & 0.93$\pm$0.02 & 0.58$\pm$0.20	&	0.32$\pm$0.05 & 0.30$\pm$0.02 & 0.99$\pm$0.03	&	 \cellcolor{2best} 0.50$\pm$0.11 & \cellcolor{best}   0.76$\pm$0.07 &  \cellcolor{2best}   0.92$\pm$0.04	&\cellcolor{best} 	0.29$\pm$0.06 & \cellcolor{best} 0.28$\pm$0.02 & \cellcolor{best} 0.97$\pm$0.04  \\

\textbf{PE--LL} &  0.48$\pm$0.01&  0.15$\pm$0.01& 0.93$\pm$0.01	&	0.79$\pm$0.01& 0.92$\pm$0.01& \cellcolor{2best} 0.43$\pm$0.01	& 0.29$\pm$0.01& 0.29$\pm$0.01& 0.99$\pm$0.01	&	0.33$\pm$0.04 & 0.69$\pm$0.01 & 0.99$\pm$0.01	&	0.20$\pm$0.02& 0.25$\pm$0.01& 1.00$\pm$0.00\\

\textbf{SE} &  0.51$\pm$0.01 & 0.17$\pm$0.00& \cellcolor{2best} 0.86$\pm$0.02	&	\cellcolor{2best} 0.81$\pm$0.00&  \cellcolor{2best}0.93$\pm$0.00& \cellcolor{best} 0.41$\pm$0.01	&	0.29$\pm$0.00& 0.29$\pm$0.00& 1.00$\pm$0.00	&	0.32$\pm$0.02 & 0.69$\pm$0.01 & 0.99$\pm$0.01	&	0.17$\pm$0.01 & 0.24$\pm$0.00& 1.00$\pm$0.00\\

\textbf{VBLL} &  0.44$\pm$0.02 & 0.14$\pm$0.01 & 0.96$\pm$0.02	&	0.54$\pm$0.14 & 0.84$\pm$0.07 & 0.88$\pm$0.13	&	\cellcolor{2best} 0.49$\pm$0.10& \cellcolor{2best} 0.36$\pm$0.07 &   \cellcolor{2best} 0.90$\pm$0.12	&	0.48$\pm$0.03 &  0.42$\pm$0.04 & 0.96$\pm$0.02	&	0.16$\pm$0.03 & 0.24$\pm$0.01 & \cellcolor{2best} 0.99$\pm$0.01 \\


\textbf{LL--HMC} &\cellcolor{2best} 0.53$\pm$0.02 &\cellcolor{2best} 0.18$\pm$0.01 & 0.87$\pm$0.05	&	\cellcolor{best} 0.82$\pm$0.06 & \cellcolor{best} 0.94$\pm$0.02 & 0.48$\pm$0.20	& \cellcolor{best} 	0.57$\pm$0.06 & \cellcolor{best} 0.42$\pm$0.04 &\cellcolor{best}  0.78$\pm$0.10	&	\cellcolor{best} 0.57$\pm$0.06 & \cellcolor{2best}   0.74$\pm$0.02  & \cellcolor{best} 0.78$\pm$0.10	& \cellcolor{2best} 0.28$\pm$0.08 & \cellcolor{2best} 0.27$\pm$0.02 & \cellcolor{best} 0.97$\pm$0.03  \\

\bottomrule
\end{tabular}
}
 \label{tab:road_ood_exp_hp}
\end{table*}


\subsection{Uncertainty-based out-of-distribution detection performance}
\paragraph{\textbf{Grid search results.}} Table \ref{tab:road_ood_exp} reports the average OOD detection performance for the top-performing hyperparameter configurations per random seed. Table \ref{tab:road_ood_exp_hp} shows the results of the single hyperparameter configuration that achieved the best average performance across the random seeds. For every OOD scenario, at least one PDL method outperforms the regular softmax baseline. Table \ref{tab:road_ood_exp} shows that the top-performing LL--HMC configuration produces the top or the best uncertainty-based OOD detection results. The other PDL methods also exhibit comparable, and at times superior, performance. However, as observed for in-distribution classification results, Table \ref{tab:road_ood_exp_hp} highlights that the grid searched hyperparameter configurations are in most scenarios not robust to different random seeds and often result in substantial performance degradation. Only the AIDE OOD max scenario (with the most OOD instances) shows comparable performance to the top-performing configurations.

\begin{figure*}[t!]
\centering
\begin{tikzpicture}
\begin{groupplot}[
group style={
group size=3 by 2,
 horizontal sep=1.0cm,
vertical sep=0.75cm,
 },
 width=5.5cm,
 height=4.0cm,
 grid=major,
 xlabel={Number of samples/predictions},
 xtick={1,5,10,15,20,25,30,35,40,45,50},
 tick label style={font=\small},
label style={font=\small},
 y tick label style={
/pgf/number format/fixed,
 /pgf/number format/precision=2,
/pgf/number format/fixed zerofill=true,
 },
 tick label style={font=\fontsize{4}{4}\selectfont},
 xlabel style={yshift=0.1cm}, 
]

\nextgroupplot[
title={{\textbf{AIDE} \citep{yang2023aide}}},
title style={yshift=-0.2cm},
 ylabel={\small{PR-AUC}},
ymin=0.15,
ymax=0.66,
]

\addplot[mark=*, thin, line width=0.25pt, mark options={fill=orange}, mark size=1.0pt, error bars/.cd,
 y dir=both,
 y explicit,
error bar style={line width=0.1pt, color=orange}] coordinates {
(5.0,0.4099293594955154) += (0,0.0611586526848039) -=(0,0.0611586526848039)
(10.0,0.4708471672582329) += (0,0.06596160100760632) -=(0,0.06596160100760632)
(15.0,0.46847685413525325) += (0,0.06786657155694765) -=(0,0.06786657155694765)
(20.0,0.5961098446503498) += (0,0.06015507569947847) -=(0,0.06015507569947847)
(25.0,0.5619578990892682) += (0,0.0612660126788448) -=(0,0.0612660126788448)
(30.0,0.587875672510149) += (0,0.054593452295624796) -=(0,0.054593452295624796)
(35.0,0.5542912596218069) += (0,0.0583492582662243) -=(0,0.0583492582662243)
(40.0,0.5699888898962179) += (0,0.05678440536488734) -=(0,0.05678440536488734)
(45.0,0.5558766945490639) += (0,0.06268800160241772) -=(0,0.06268800160241772)
(50.0,0.5903266926606767) += (0,0.07500588661606554) -=(0,0.07500588661606554)
};

\addplot[mark=triangle*, thin, line width=0.25pt, mark options={fill=cyan}, mark size=2pt,error bars/.cd,
 y dir=both,
y explicit,
error bar style={line width=0.1pt, color=cyan}] coordinates {
(5,0.206597056129927) += (0,0.0017465905234542) -=(0,0.0017465905234542)
(10,0.206923153624713) += (0,0.00305923557638305) -=(0,0.00305923557638305)
(15,0.207293790859179) += (0,0.00354230011203544) -=(0,0.00354230011203544)
(20,0.207752941332176) += (0,0.00368609183154441) -=(0,0.00368609183154441)
(25,0.20803022139541) += (0,0.00315258737416246) -=(0,0.00315258737416246)
(30,0.208482846513983) += (0,0.00121611497633725) -=(0,0.00121611497633725)
(35,0.205702302169141) += (0,0.00233200954669881) -=(0,0.00233200954669881)
(40,0.206329405723692) += (0,0.00273402670658221) -=(0,0.00273402670658221)
(45,0.207862360300522) += (0,0.00318044891623496) -=(0,0.00318044891623496)
(50,0.206884871100107) += (0,0.00263379413199315) -=(0,0.00263379413199315)

};

\addplot[mark=square*, thin, line width=0.25pt, mark options={fill=red}, mark size=1.0pt,
error bars/.cd,
 y dir=both,
 y explicit,
error bar style={line width=0.1pt, color=yellow}] coordinates {
(5,0.20694511344929709) += (0,0.001373024874313264) -=(0,0.001373024874313264)
(10,0.20611171048491678) += (0,0.0009710375647568551) -=(0,0.0009710375647568551)
(15,0.20637550980293162) += (0,0.000658744036075908) -=(0,0.000658744036075908)
(20,0.2063492428356545) += (0,0.0005075677891823795) -=(0,0.0005075677891823795)
(25,0.2067879494966308) += (0,0.0005741695420387532) -=(0,0.0005741695420387532)
(30,0.20682007591505136) += (0,0.0005395029330917639) -=(0,0.0005395029330917639)
(35,0.2064812404309772) += (0,0.000746011032822557) -=(0,0.000746011032822557)
(40,0.20629989745120106) += (0,0.0005379654165682829) -=(0,0.0005379654165682829)
(45,0.20635113018699108) += (0,0.0004337772666133956) -=(0,0.0004337772666133956)
(50,0.20655758244982697) += (0,0.0003972238421868066) -=(0,0.0003972238421868066)
};

\addplot[mark=diamond*, thin, line width=0.25pt, mark options={fill=yellow}, mark size=1.0pt,
error bars/.cd,
y dir=both,
y explicit,
error bar style={line width=0.1pt, color=yellow}] coordinates {
(5,0.38177465178644876) += (0,0.05353110292857039) -=(0,0.05353110292857039)
(10,0.4214417583457424) += (0,0.01893322197300767) -=(0,0.01893322197300767)
(15,0.4166176479774652) += (0,0.04085449754162002) -=(0,0.04085449754162002)
(20,0.4302313222856375) += (0,0.022941019011776953) -=(0,0.022941019011776953)
(25,0.4538529905132734) += (0,0.02964893808546523) -=(0,0.02964893808546523)
(30,0.45759031304906833) += (0,0.02982390886549429) -=(0,0.02982390886549429)
(35,0.46095961793450313) += (0,0.023537446545199623) -=(0,0.023537446545199623)
(40,0.4617348365783642) += (0,0.04175402035974026) -=(0,0.04175402035974026)
(45,0.46653975967457456) += (0,0.048281145328221614) -=(0,0.048281145328221614)
(50,0.4834894076461963) += (0,0.015383569420799786) -=(0,0.015383569420799786)
};

\addplot[mark=halfcircle*, thin, line width=0.25pt, mark options={fill=magenta}, mark size=1.5pt] coordinates {
(5,82.24)
};

\addplot[mark=oplus*, thin, line width=0.25pt, mark options={fill=pink}, mark size=1.5pt] coordinates {
(1,0.20)
};

\addplot[mark=pentagon*, thin, line width=0.25pt, mark options={fill=green}, mark size=1.5pt,
error bars/.cd,
y dir=both,
y explicit,
error bar style={line width=0.1pt, color=green}] coordinates {
(1,0.16784053115103267) += (0,0.010862086427357984) -=(0,0.010862086427357984)
};

\nextgroupplot[
title={{\textbf{B4C} \citep{jain2015car}}},
 title style={yshift=-0.2cm},
ylabel={\small{PR-AUC}},
ymax=0.65,
ymin=0.35,
]

\addplot[mark=*, thin, line width=0.25pt, mark options={fill=orange}, mark size=1.0pt,error bars/.cd,
 y dir=both,
y explicit,
error bar style={line width=0.1pt, color=orange}] coordinates {
(5.0,0.4638115966944718) += (0,0.06263507963334694) -=(0,0.06263507963334694)
(10.0,0.46837843860822764) += (0,0.03929424611442142) -=(0,0.03929424611442142)
(15.0,0.5383142087951663) += (0,0.0597395982855219) -=(0,0.0597395982855219)
(20.0,0.5015212997930913) += (0,0.02829283734024349) -=(0,0.02829283734024349)
(25.0,0.5367705848030446) += (0,0.022485559864272183) -=(0,0.022485559864272183)
(30.0,0.5518880202900334) += (0,0.03407983637151061) -=(0,0.03407983637151061)
(35.0,0.5709447653908899) += (0,0.0769850406691456) -=(0,0.0769850406691456)
(40.0,0.5831610515156422) += (0,0.0563969018049688) -=(0,0.0563969018049688)
(45.0,0.5558371519949898) += (0,0.05581561722959842) -=(0,0.05581561722959842)
(50.0,0.5695414652006476) += (0,0.049748689551465544) -=(0,0.049748689551465544)
};

\addplot[mark=triangle*, thin, line width=0.25pt, mark options={fill=cyan}, mark size=1pt,error bars/.cd,
 y dir=both,
 y explicit,
error bar style={line width=0.1pt, color=cyan}] coordinates {
(5,0.388887543489991) += (0,0.00311444284845945) -=(0,0.00311444284845945)
(10,0.389200067551477) += (0,0.00259077984889378) -=(0,0.00259077984889378)
(15,0.390401353785549) += (0,0.00132358122043397) -=(0,0.00132358122043397)
(20,0.391116552369851) += (0,0.00184023303465676) -=(0,0.00184023303465676)
(25,0.391361941738688) += (0,0.00132555372754428) -=(0,0.00132555372754428)
(30,0.391546567338954) += (0,0.00143462012635332) -=(0,0.00143462012635332)
(35,0.39043902377285) += (0,0.00135201730366195) -=(0,0.00135201730366195)
(40,0.391520530764576) += (0,0.00202468151060397) -=(0,0.00202468151060397)
(45,0.390528943386747) += (0,0.00171498562411969) -=(0,0.00171498562411969)
(50,0.391301699623872) += (0,0.00133479426537446) -=(0,0.00133479426537446)

};

\addplot[mark=square*, thin, line width=0.25pt, mark options={fill=red}, mark size=1.0pt,
error bars/.cd,
y dir=both,
y explicit,
error bar style={line width=0.1pt, color=red}] coordinates {
(5,0.4206446511658267) += (0,0.00460980486790938) -=(0,0.00460980486790938)
(10,0.4194405869966336) += (0,0.0035345981442183187) -=(0,0.0035345981442183187)
(15,0.4213187852732581) += (0,0.002975979508316335) -=(0,0.002975979508316335)
(20,0.4199438772617269) += (0,0.0016168051918349074) -=(0,0.0016168051918349074)
(25,0.4203765071938733) += (0,0.001856711833077699) -=(0,0.001856711833077699)
(30,0.42074098327383525) += (0,0.0032297818789886475) -=(0,0.0032297818789886475)
(35,0.4233719281478428) += (0,0.001037373570778641) -=(0,0.001037373570778641)
(40,0.4190402100703186) += (0,0.002717013237190011) -=(0,0.002717013237190011)
(45,0.4192048775709722) += (0,0.0017838535081197776) -=(0,0.0017838535081197776)
(50,0.4206813168449681) += (0,0.001979401777005892) -=(0,0.001979401777005892)
};

\addplot[mark=diamond*, thin, line width=0.25pt,
mark options={fill=yellow},
mark size=1.0pt,
error bars/.cd,
y dir=both,
y explicit,
error bar style={line width=0.1pt, color=yellow}] coordinates {
(5,0.49154039080155254) += (0,0.029341496660601556) -=(0,0.029341496660601556)
(10,0.4786783124541153) += (0,0.046594966526135836) -=(0,0.046594966526135836)
(15,0.4858752502869305) += (0,0.03643797444855078) -=(0,0.03643797444855078)
(20,0.49316223276116683) += (0,0.05662706336336582) -=(0,0.05662706336336582)
(25,0.5055797138457091) += (0,0.03157582500057088) -=(0,0.03157582500057088)
(30,0.5000315859462228) += (0,0.03793796850444391) -=(0,0.03793796850444391)
(35,0.5094513789672975) += (0,0.01639906660142857) -=(0,0.01639906660142857)
(40,0.4878801810924019) += (0,0.03814407001476142) -=(0,0.03814407001476142)
(45,0.487995741214408) += (0,0.03930492837330372) -=(0,0.03930492837330372)
(50,0.507522857379817) += (0,0.025232759972828693) -=(0,0.025232759972828693)
};

\addplot[mark=halfcircle*, thin, line width=0.25pt, mark options={fill=magenta}, mark size=1.5pt] coordinates {
(5,0.39)
};

\addplot[mark=oplus*, thin, line width=0.25pt, mark options={fill=pink}, mark size=1.5pt] coordinates {
(1,0.37)
};

\addplot[mark=pentagon*, thin, line width=0.25pt, mark options={fill=green}, mark size=1.5pt,
error bars/.cd,
 y dir=both,
y explicit,
error bar style={line width=0.1pt, color=green}] coordinates {
(1,0.46620109960672795) += (0,0.047805178351586735) -=(0,0.047805178351586735)
};

\nextgroupplot[
title={{\textbf{ROAD} \citep{singh2022road}}},
 title style={yshift=-0.2cm},
 ylabel={\small{PR-AUC}},
 ymin=0.2,
 ymax=0.45
]

\addplot[mark=*, thin, line width=0.25pt, mark options={fill=orange}, mark size=1.0pt, error bars/.cd,
 y dir=both,
 y explicit,
error bar style={line width=0.1pt, color=orange}] coordinates {
(5.0,0.38773302582836655) += (0,0.06927663458620247) -=(0,0.06927663458620247)
(10.0,0.3846003238762753) += (0,0.06946059174627063) -=(0,0.06946059174627063)
(15.0,0.32491859375872256) += (0,0.03606627073289094) -=(0,0.03606627073289094)
(20.0,0.39729878889193204) += (0,0.05093776633649451) -=(0,0.05093776633649451)
(25.0,0.37972716205254164) += (0,0.06648795556275817) -=(0,0.06648795556275817)
(30.0,0.33951794870287716) += (0,0.03920956973347966) -=(0,0.03920956973347966)
(35.0,0.38624654790624013) += (0,0.07775981881775736) -=(0,0.07775981881775736)
(40.0,0.35987718338549146) += (0,0.04119934271810508) -=(0,0.04119934271810508)
(45.0,0.3808373415939105) += (0,0.08027787441865955) -=(0,0.08027787441865955)
(50.0,0.4123227895682818) += (0,0.06473062325494665) -=(0,0.06473062325494665)
};

\addplot[mark=triangle*, thin, line width=0.25pt, mark options={fill=cyan}, mark size=1pt,error bars/.cd,
 y dir=both,
 y explicit,
error bar style={line width=0.1pt, color=cyan}] coordinates {
(5,0.263640935127315) += (0, 0.00314024347491678) -=(0, 0.00314024347491678)
(10,0.261869460410927) += (0, 0.00251039464151634) -=(0, 0.00251039464151634)
(15,0.263270568660843) += (0, 0.00242108935926339) -=(0, 0.00242108935926339)
(20,0.263274216199517) += (0, 0.00187659386685288) -=(0, 0.00187659386685288)
(25,0.263805198034818) += (0, 0.00180380488440164) -=(0,0.00180380488440164)
(30,0.263404589113462) += (0, 0.00119894210076869) -=(0,0.00119894210076869)
(35,0.263890956008082) += (0, 0.00142530288154616) -=(0,0.00142530288154616)
(40,0.26311402456255) += (0,0.00164960471978395) -=(0,0.00164960471978395)
(45,0.263780362982253) += (0,0.0015008035329665) -=(0,0.0015008035329665)
(50,0.263267408171233) += (0,0.00170841548066841) -=(0,0.00170841548066841)
};

\addplot[mark=square*, thin, line width=0.25pt, mark options={fill=red}, mark size=1.0pt,
error bars/.cd,
 y dir=both,
y explicit,
error bar style={line width=0.1pt, color=red}] coordinates {
(5,0.3133075629044341) += (0,0.01218731778942207) -=(0,0.01218731778942207)
(10,0.3104261371113608) += (0,0.005470732865910389) -=(0,0.005470732865910389)
(15,0.3138265304819588) += (0,0.005856994784797176) -=(0,0.005856994784797176)
(20,0.3196035307762033) += (0,0.00790886567990158) -=(0,0.00790886567990158)
(25,0.32096080747240985) += (0,0.00833132646399127) -=(0,0.00833132646399127)
(30,0.31792014751682107) += (0,0.004430483838032742) -=(0,0.004430483838032742)
(35,0.3177472908417346) += (0,0.003392476562095908) -=(0,0.003392476562095908)
(40,0.3178155109752583) += (0,0.003980072054699508) -=(0,0.003980072054699508)
(45,0.32215951534707027) += (0,0.006032049968494871) -=(0,0.006032049968494871)
(50,0.3215322735884013) += (0,0.003270997191830635) -=(0,0.003270997191830635)
};

\addplot[mark=diamond*, thin, line width=0.25pt, mark options={fill=yellow}, mark size=1.0pt, error bars/.cd,
 y dir=both, y explicit,
error bar style={line width=0.1pt, color=yellow}] coordinates {
(5,0.30259105891012805) += (0,0.025997233083679124) -=(0,0.025997233083679124)
(10,0.2913667249900822) += (0,0.016209804868846067) -=(0,0.016209804868846067)
(15,0.31814389406627586) += (0,0.03973041935650117) -=(0,0.03973041935650117)
(20,0.3287216845015036) += (0,0.0503565554763321) -=(0,0.0503565554763321)
(25,0.31062515571996696) += (0,0.044754320959089945) -=(0,0.044754320959089945)
(30,0.28833208941681676) += (0,0.022108741344449857) -=(0,0.022108741344449857)
(35,0.3302366083064697) += (0,0.05503833240035041) -=(0,0.05503833240035041)
(40,0.3045102230648385) += (0,0.02073995834927245) -=(0,0.02073995834927245)
(45,0.3042644592758631) += (0,0.02220499653733075) -=(0,0.02220499653733075)
(50,0.30972900759294825) += (0,0.027734898495034876) -=(0,0.027734898495034876)
};

\addplot[mark=halfcircle*, thin, line width=0.25pt, mark options={fill=magenta}, mark size=1.5pt] coordinates {
(5,0.41)
};

\addplot[mark=oplus*, thin, line width=0.25pt, mark options={fill=pink}, mark size=1.5pt] coordinates {
(1,0.24955)
};

\addplot[mark=pentagon*, thin, line width=0.25pt, mark options={fill=green}, mark size=1.5pt,
error bars/.cd,
y dir=both,
y explicit,
error bar style={line width=0.1pt, color=green}] coordinates {
(1,0.24497345524411732) +=(0,0.00616422784742784) -=(0,0.00616422784742784)
};

\nextgroupplot[
title=\empty,
 ylabel={\small{PR-AUC}},
 ymin=0.85,
ymax=0.98
]

\addplot[mark=*, thin, line width=0.25pt, mark options={fill=orange}, mark size=1.0pt, error bars/.cd,
y dir=both,
 y explicit,
error bar style={line width=0.1pt, color=orange}] coordinates {
(5.0,0.8890952355627204) += (0,0.02047053526188267) -=(0,0.02047053526188267)
(10.0,0.8953625367929087) += (0,0.01246797746193798) -=(0,0.01246797746193798)
(15.0,0.9067725793580275) += (0,0.013519808461528654) -=(0,0.013519808461528654)
(20.0,0.8969385261953808) += (0,0.01036496620579844) -=(0,0.01036496620579844)
(25.0,0.9130454847262927) += (0,0.014946356767675757) -=(0,0.014946356767675757)
(30.0,0.8938326988036012) += (0,0.01201948574187212) -=(0,0.01201948574187212)
(35.0,0.8970527741214148) += (0,0.015264353856984263) -=(0,0.015264353856984263)
(40.0,0.9083495177397062) += (0,0.014328574100584288) -=(0,0.014328574100584288)
(45.0,0.9127346098767953) += (0,0.00698748386415729) -=(0,0.00698748386415729)
(50.0,0.9005882102907485) += (0,0.01466810260714856) -=(0,0.01466810260714856)
};

\addplot[mark=triangle*, thin, line width=0.25pt, mark options={fill=cyan}, mark size=1pt,error bars/.cd,
 y dir=both,
 y explicit,
error bar style={line width=0.1pt, color=cyan}] coordinates {

(5,0.928506286220776) += (0,0.000734413629306157) -=(0,0.000734413629306157)
(10,0.929271199548355) += (0,0.000425000269828696) -=(0,0.000425000269828696)
(15,0.929162027915687) += (0,0.000560941799247343) -=(0,0.000560941799247343)
(20,0.929211982128556) += (0,0.000311745537649016) -=(0,0.000311745537649016)
(25,0.929082762510006) += (0,0.000397726468606998) -=(0,0.000397726468606998)
(30,0.929487925667901) += (0,0.000649161789004993) -=(0,0.000649161789004993)
(35,0.929329897723539) += (0,0.000430878838722062) -=(0,0.000430878838722062)
(40,0.929541701043627) += (0,0.000512409095274379) -=(0,0.000512409095274379)
(45,0.929209215211515) += (0,0.000387047022626515) -=(0,0.000387047022626515)
(50,0.929329737938024) += (0,0.000561129715452201) -=(0,0.000561129715452201)
};

\addplot[mark=square*, thin, line width=0.25pt, mark options={fill=red}, mark size=1.0pt,
error bars/.cd,
 y dir=both,
 y explicit,
error bar style={line width=0.1pt, color=yellow}] coordinates {
(5,0.9555835895870072) += (0,0.002832811409249749) -=(0,0.002832811409249749)
(10,0.9587466540049675) += (0,0.0019056627083049182) -=(0,0.0019056627083049182)
(15,0.9597025633785752) += (0,0.0013939626388902047) -=(0,0.0013939626388902047)
(20,0.9602738656910887) += (0,0.0009153579402506462) -=(0,0.0009153579402506462)
(25,0.960161140329007) += (0,0.0010255319525492982) -=(0,0.0010255319525492982)
(30,0.9601480522823737) += (0,0.0009191390494881075) -=(0,0.0009191390494881075)
(35,0.9608012772991531) += (0,0.000983793732688009) -=(0,0.000983793732688009)
(40,0.9600423600296191) += (0,0.000951478299868778) -=(0,0.000951478299868778)
(45,0.9602279128121924) += (0,0.0007558661646736411) -=(0,0.0007558661646736411)
(50,0.9604300014730894) += (0,0.0010174092592789104) -=(0,0.0010174092592789104)
};

\addplot[mark=diamond*, thin, line width=0.25pt, mark options={fill=yellow}, mark size=1.0pt,
error bars/.cd,
y dir=both,
y explicit,
error bar style={line width=0.1pt, color=yellow}] coordinates {
(5,0.9617172352105422) += (0,0.0034651347012903784) -=(0,0.0034651347012903784)
(10,0.9646612480139651) += (0,0.002391206411039989) -=(0,0.002391206411039989)
(15,0.9691497394501779) += (0,0.001712411808605333) -=(0,0.001712411808605333)
(20,0.971158174692748) += (0,0.001375620955681129) -=(0,0.001375620955681129)
(25,0.9706999801065308) += (0,0.001210874051052671) -=(0,0.001210874051052671)
(30,0.9720006924024169) += (0,0.001200270967006817) -=(0,0.001200270967006817)
(35,0.9714917236460485) += (0,0.001485601366579603) -=(0,0.001485601366579603)
(40,0.9711898294705111) += (0,0.0022849187140955524) -=(0,0.0022849187140955524)
(45,0.9719698549924789) += (0,0.0009724128543781708) -=(0,0.0009724128543781708)
(50,0.9727679502293537) += (0,0.0007986064197992981) -=(0,0.0007986064197992981)
};

\addplot[mark=halfcircle*, thin, line width=0.25pt, mark options={fill=magenta}, mark size=1.5pt] coordinates {
(5,0.94)
};

\addplot[mark=oplus*, thin, line width=0.25pt, mark options={fill=pink}, mark size=1.5pt] coordinates {
(1,0.94)
};

\addplot[mark=pentagon*, thin, line width=0.25pt, mark options={fill=green}, mark size=1.5pt,
error bars/.cd,
y dir=both,
y explicit,
error bar style={line width=0.1pt, color=green}] coordinates {
(1,0.9419845204867905) += (0,0.012282229562846944) -=(0,0.012282229562846944)
};

\nextgroupplot[
title= \empty,
title style={yshift=-0.2cm},
 ylabel={\small{PR-AUC}},
 ymax=0.97,
 ymin=0.78,
]

\addplot[mark=*, thin, line width=0.25pt, mark options={fill=orange}, mark size=1.0pt,error bars/.cd,
y dir=both,
y explicit,
error bar style={line width=0.1pt, color=orange}] coordinates {
(5.0,0.8890952355627204) += (0,0.02047053526188267) -=(0,0.02047053526188267)
(10.0,0.8953625367929087) += (0,0.01246797746193798) -=(0,0.01246797746193798)
(15.0,0.9067725793580275) += (0,0.013519808461528654) -=(0,0.013519808461528654)
(20.0,0.8969385261953808) += (0,0.01036496620579844) -=(0,0.01036496620579844)
(25.0,0.9130454847262927) += (0,0.014946356767675757) -=(0,0.014946356767675757)
(30.0,0.8938326988036012) += (0,0.01201948574187212) -=(0,0.01201948574187212)
(35.0,0.8970527741214148) += (0,0.015264353856984263) -=(0,0.015264353856984263)
(40.0,0.9083495177397062) += (0,0.014328574100584288) -=(0,0.014328574100584288)
(45.0,0.9127346098767953) += (0,0.00698748386415729) -=(0,0.00698748386415729)
(50.0,0.9005882102907485) += (0,0.01466810260714856) -=(0,0.01466810260714856)
};

\addplot[mark=triangle*, thin, line width=0.25pt, mark options={fill=cyan}, mark size=1pt,error bars/.cd,
y dir=both,
 y explicit,
error bar style={line width=0.1pt, color=cyan}] coordinates {
(5,0.837712690299853) += (0,0.00293174035336785) -=(0,0.00293174035336785)
(10,0.837536975655889) += (0,0.00244074565522067) -=(0,0.00244074565522067)
(15,0.839048665406508) += (0,0.00135288761101275) -=(0,0.00135288761101275)
(20,0.839718175171287) += (0,0.00150997236248267) -=(0,0.00150997236248267)
(25,0.838602931239869) += (0,0.00131160240101608) -=(0,0.00131160240101608)
(30,0.839134209927864) += (0,0.00135957719499665) -=(0,0.00135957719499665)
(35,0.838900814730659) += (0,0.00139051854318541) -=(0,0.00139051854318541)
(40,0.838050485938717) += (0,0.00178767699904112) -=(0,0.00178767699904112)
(45,0.838403433008571) += (0,0.00150399376015297) -=(0,0.00150399376015297)
(50,0.839243671699682) += (0,0.00121778689193249) -=(0,0.00121778689193249)
};

\addplot[mark=square*, thin, line width=0.25pt, mark options={fill=red}, mark size=1.0pt,
error bars/.cd,
 y dir=both,
 y explicit,
error bar style={line width=0.1pt, color=red}] coordinates {
(5,0.8720259503349436) += (0,0.0019019623864402633) -=(0,0.0019019623864402633)
(10,0.8757909638851944) += (0,0.001515904940566373) -=(0,0.001515904940566373)
(15,0.8757736588520493) += (0,0.0013690623348630718) -=(0,0.0013690623348630718)
(20,0.8754886673420307) += (0,0.002493356066266395) -=(0,0.002493356066266395)
(25,0.8748057411117042) += (0,0.0014246995655760824) -=(0,0.0014246995655760824)
(30,0.8758860975688926) += (0,0.0010697960683935824) -=(0,0.0010697960683935824)
(35,0.8745137286778558) += (0,0.000931580980482559) -=(0,0.000931580980482559)
(40,0.8742627982872954) += (0,0.000859664409545465) -=(0,0.000859664409545465)
(45,0.8743424201502226) += (0,0.0005312108151598462) -=(0,0.0005312108151598462)
(50,0.874138561338419) += (0,0.001574229983756852) -=(0,0.001574229983756852)
};

\addplot[mark=diamond*, thin, line width=0.25pt,
mark options={fill=yellow},
mark size=1.0pt,
error bars/.cd,
 y dir=both,
 y explicit,
error bar style={line width=0.1pt, color=yellow}] coordinates {
(5,0.934234387079315) += (0,0.007482862024564887) -=(0,0.007482862024564887)
(10,0.9463293321330107) += (0,0.007834248512217627) -=(0,0.007834248512217627)
(15,0.9480493321202473) += (0,0.006899738096230493) -=(0,0.006899738096230493)
(20,0.9510705083455543) += (0,0.004403889758786524) -=(0,0.004403889758786524)
(25,0.9561847616749584) += (0,0.004935406059881822) -=(0,0.004935406059881822)
(30,0.9603580205893574) += (0,0.0027788470535201127) -=(0,0.0027788470535201127)
(35,0.9578881533621375) += (0,0.002279184518381539) -=(0,0.002279184518381539)
(40,0.9560222759258465) += (0,0.0030825319800790515) -=(0,0.0030825319800790515)
(45,0.9601004612890331) += (0,0.001813134265007449) -=(0,0.001813134265007449)
(50,0.9595021218847638) += (0,0.0020104443722019485) -=(0,0.0020104443722019485)
};

\addplot[mark=halfcircle*, thin, line width=0.25pt, mark options={fill=magenta}, mark size=1.5pt] coordinates {
(5,0.90)
};

\addplot[mark=oplus*, thin, line width=0.25pt, mark options={fill=pink}, mark size=1.5pt] coordinates {
(1,0.82)
};

\addplot[mark=pentagon*, thin, line width=0.25pt, mark options={fill=green}, mark size=1.5pt,
error bars/.cd,
 y dir=both,
y explicit,
error bar style={line width=0.1pt, color=green}] coordinates {
(1,0.7971741487324044) += (0,0.013486337831006126) -=(0,0.013486337831006126)
};

\end{groupplot}
\node[font=\bfseries, rotate=90] at ($(group c1r1.west)+(-1.00cm,0)$) {OOD min};

 \node[font=\bfseries, rotate=90] at ($(group c1r1.west)+(-1.00cm,-3.2)$) {OOD max};
\end{tikzpicture}

\vspace*{0.3cm}
\begin{tikzpicture}
\begin{axis}[
hide axis,
 xmin=0, xmax=1, ymin=0, ymax=1,
legend style={at={(0.0,0.0)}, anchor=center, legend columns=7},
legend entries={\tiny{BBB--LL},\tiny{PE--LL},\tiny{SE},\tiny{VBLL}, \tiny{LL--HMC},\tiny{DE},\tiny{Regular},}
]

\addlegendimage{mark=*, thin, line width=0.25pt, mark options={fill=orange}, mark size=2.0pt}
\addlegendimage{mark=triangle*, thin, line width=0.25pt, mark options={fill=cyan}, mark size=2.0pt}
\addlegendimage{mark=square*, thin, line width=0.25pt, mark options={fill=red}, mark size=2pt}
\addlegendimage{mark=pentagon*, thin, line width=0.25pt, mark options={fill=green}, mark size=2.0pt}
\addlegendimage{mark=diamond*, thin, line width=0.25pt, mark options={fill=yellow}, mark size=2.0pt}
\addlegendimage{mark=halfcircle*, thin, line width=0.25pt, mark options={fill=magenta}, mark size=2.0pt}
\addlegendimage{mark=oplus*, thin, line width=0.25pt, mark options={fill=pink}, mark size=2.0pt}

\end{axis}
\end{tikzpicture}
\caption{Influence of the number of samples or predictions on the OOD detection across 10 different random seeds for the average best-performing hyperparameter configuration of the included PDL methods. For each number of samples, the whiskers indicate two standard  errors of the mean. Markers without visible whiskers indicate small standard errors.} 
\label{fig:ood_comparison}

\end{figure*}
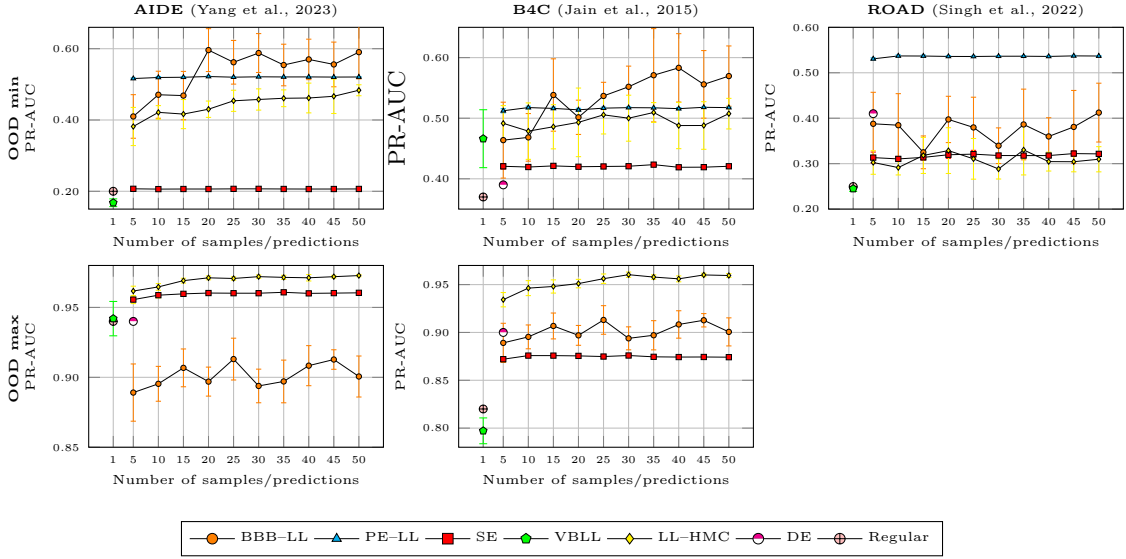

\definecolor{darkgray176}{RGB}{176,176,176}
\definecolor{steelblue31119180}{RGB}{31,119,180}

\begin{figure}[t!]
    \centering
    \begin{tikzpicture}
        \begin{groupplot}[
            group style={
                group size=3 by 2, 
                horizontal sep=1.2cm,
                vertical sep=0.3cm,
            },
            width=5.0cm,
            height=4.0cm,
            ybar,
            enlargelimits=0.15,
            ylabel={\small {PR-AUC}},
            symbolic x coords={BBB--LL, SE, PE--LL, VBLL, LL--HMC},
            xtick=data,
            grid=major,
            tick label style={font=\tiny},
    xlabel style={yshift=0.1cm, rotate=60},  
            every node near coord/.append style={
                yshift=3pt,
                font=\footnotesize,
    label style={font=\tiny},
    y tick label style={
        /pgf/number format/fixed,
        /pgf/number format/precision=2,
        /pgf/number format/fixed zerofill=true, 
    },
    tick label style={font=\fontsize{4}{4}\selectfont},
            },
        ]

        \nextgroupplot[title={ \textbf{AIDE} \citep{yang2023aide}},
         title style={yshift=-0.2cm}, xlabel=\empty,  xticklabels=\empty,
           ymin=0.17, ymax=0.44,]
        \addplot[
            fill=steelblue31119180!60,
            draw=steelblue31119180,
            error bars/.cd,
            y dir=both,
            y explicit,
        ] coordinates {
        (SE, 0.18634) +- (0.001151,0.001151)
(BBB--LL, 0.43635) +- (0.01884,0.01884)
(PE--LL, 0.206248) +- (0.0002,0.0002)
(VBLL, 0.17674) +- (0.00578,0.00578)
(LL--HMC, 0.44611) +- (0.00939,0.00939)

        };

        \nextgroupplot[title={ \textbf{B4C}  \citep{jain2015car}},
          title style={yshift=-0.2cm},
   xlabel=\empty,  xticklabels=\empty,  ymin=0.35, ymax=0.53,]
        \addplot[
            fill=steelblue31119180!60,
            draw=steelblue31119180,
            error bars/.cd,
            y dir=both,
            y explicit,
        ] coordinates {
(BBB--LL, 0.458215) +- (0.0139062,0.0139062)
(SE, 0.517133) +- (0.0049198,0.0049198)
(PE--LL, 0.390672) +- (0.00167,0.00167)
(VBLL, 0.446607) +- (0.013007,0.013007)
(LL--HMC,0.495702) +- (0.0135222,0.0135222)
        };

     \nextgroupplot[title={ \textbf{ROAD}  \citep{singh2022road}},
   title style={yshift=-0.2cm},  ymin=0.24, ymax=0.42,]
        \addplot[
            fill=steelblue31119180!60,
            draw=steelblue31119180,
            error bars/.cd,
            y dir=both,
            y explicit,
        ] coordinates {
(SE, 0.259985) +- (0.0002554,0.0002554)
(BBB--LL, 0.30883) +- (0.00871,0.00871)
(PE--LL, 0.252689) +- (0.0022,0.0022)
(VBLL, 0.2627) +- (0.00814,0.00814)
(LL--HMC, 0.38528) +- (0.00623,0.00623)

        };

        \nextgroupplot[title=\empty,
         title style={yshift=-0.2cm},
           ymin=0.85, ymax=1,]
        \addplot[
            fill=steelblue31119180!60,
            draw=steelblue31119180,
            error bars/.cd,
            y dir=both,
            y explicit,
        ] coordinates {
          (SE, 0.955661) +- (0.0010662,0.0010662)
        (BBB--LL, 0.93663) +- (0.00173,0.00173)
(PE--LL, 0.929376) +- (0.000428,0.000428)
(VBLL, 0.93656) +- (0.0032,0.0032)
(LL--HMC, 0.96756) +- (0.00072,0.00072)

        };

        \nextgroupplot[title=\empty,
   title style={yshift=-0.2cm},
     ymin=0.65, ymax=0.9,]
        \addplot[
            fill=steelblue31119180!60,
            draw=steelblue31119180,
            error bars/.cd,
            y dir=both,
            y explicit,
        ] coordinates {	
(SE,0.869604) +- (0.001336,0.001336)
(BBB--LL, 0.86651) +- (0.00418,0.00418)
(PE--LL, 0.83) +- (0.001,0.0001)
(VBLL, 0.75418) +- (0.0063,0.0063)
(LL--HMC, 0.84314) +- (0.00818,0.00818)

        };

        \end{groupplot}
         
   \node[font=\bfseries, rotate=90] at ($(group c1r1.west)+(-1.10cm,0)$) {OOD min};
   
  \node[font=\bfseries, rotate=90] at ($(group c1r1.west)+(-1.10cm,-2.7cm)$) {OOD max};
    \end{tikzpicture}
    \caption{Average uncertainty-based OOD detection performance (PR-AUC) across the included LL--PDL methods for the best performing hyperparameter configuration across 100 random seeds.}
    \label{fig:placeholder_groupplot}
\end{figure}
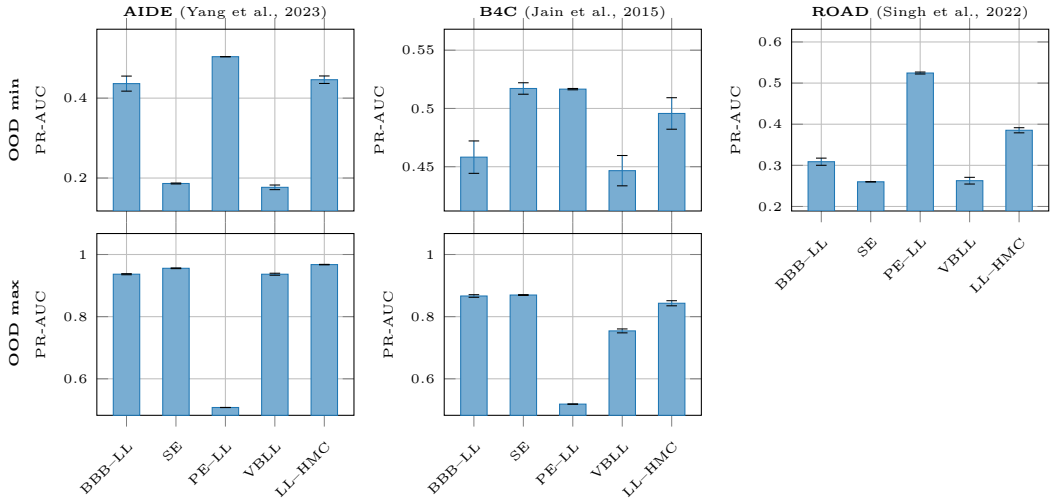

\begin{figure}[t]
    \centering
    \begin{tikzpicture}
    \begin{groupplot}[
        group style={
            group size=3 by 1,
            horizontal sep=1.2cm,
            vertical sep=1.0cm,
        },
        width=5.5cm,
        height=4.0cm,
        grid=major,
        xlabel={Number of samples},
        xtick={2,5,10,15,20,25,30,35,40,45,50},
        xmin=2,
        xmax=50,
        tick label style={font=\small},
        label style={font=\small},
        y tick label style={
            /pgf/number format/fixed,
            /pgf/number format/precision=2,
            /pgf/number format/fixed zerofill=true,
        },
        xlabel style={yshift=0.1cm},
    ]
    \nextgroupplot[
        title={{\textbf{AIDE} \citep{yang2023aide}}},
        ylabel={\small{PR-AUC}},
        title style={yshift=-0.2cm},
        ymax=0.3,
        ymin=0.2
    ]
\path [fill=steelblue31119180, fill opacity=0.2]
(axis cs:2,0.20082696612975)
--(axis cs:2,0.20082696612975)
--(axis cs:3,0.206269136026063)
--(axis cs:4,0.211883652161684)
--(axis cs:5,0.216906773173809)
--(axis cs:6,0.221433673101807)
--(axis cs:7,0.224979655732907)
--(axis cs:8,0.229382069005079)
--(axis cs:9,0.229767420802002)
--(axis cs:10,0.229355053070612)
--(axis cs:11,0.231222586786899)
--(axis cs:12,0.234677046653289)
--(axis cs:13,0.237397688913769)
--(axis cs:14,0.23862981623995)
--(axis cs:15,0.237755315125637)
--(axis cs:16,0.23780934081883)
--(axis cs:17,0.238106650987553)
--(axis cs:18,0.239101091432751)
--(axis cs:19,0.240559812380572)
--(axis cs:20,0.240419581962129)
--(axis cs:21,0.240742894576274)
--(axis cs:22,0.241704158679544)
--(axis cs:23,0.243020111748291)
--(axis cs:24,0.243840423565301)
--(axis cs:25,0.244065060984747)
--(axis cs:26,0.243177961002371)
--(axis cs:27,0.243099436840492)
--(axis cs:28,0.244526920061491)
--(axis cs:29,0.244624914698632)
--(axis cs:30,0.244895391983439)
--(axis cs:31,0.245602701877174)
--(axis cs:32,0.245082200393341)
--(axis cs:33,0.246141289139533)
--(axis cs:34,0.247118844682639)
--(axis cs:35,0.247634315603031)
--(axis cs:36,0.248008790761979)
--(axis cs:37,0.248823552300667)
--(axis cs:38,0.249464297576278)
--(axis cs:39,0.249881043162904)
--(axis cs:40,0.250754016221559)
--(axis cs:41,0.251222522614405)
--(axis cs:42,0.250764158966271)
--(axis cs:43,0.251362445393215)
--(axis cs:44,0.251262416340393)
--(axis cs:45,0.251625930786636)
--(axis cs:46,0.251370426302328)
--(axis cs:47,0.251217220464025)
--(axis cs:48,0.251526597033637)
--(axis cs:49,0.251361473495114)
--(axis cs:50,0.251747240973589)
--(axis cs:50,0.282507634660928)
--(axis cs:50,0.282507634660928)
--(axis cs:49,0.282275325499589)
--(axis cs:48,0.282602678489762)
--(axis cs:47,0.282452290992835)
--(axis cs:46,0.282762859865336)
--(axis cs:45,0.283171461006649)
--(axis cs:44,0.282945849698275)
--(axis cs:43,0.28318279325004)
--(axis cs:42,0.282710106972456)
--(axis cs:41,0.283309634021306)
--(axis cs:40,0.282957039543784)
--(axis cs:39,0.282197711588964)
--(axis cs:38,0.281931152516512)
--(axis cs:37,0.281461472695229)
--(axis cs:36,0.280840139220452)
--(axis cs:35,0.280673631384622)
--(axis cs:34,0.280360933294929)
--(axis cs:33,0.279597117801791)
--(axis cs:32,0.278808820793172)
--(axis cs:31,0.279623642640749)
--(axis cs:30,0.279187601038192)
--(axis cs:29,0.279236370741613)
--(axis cs:28,0.279467523009777)
--(axis cs:27,0.278357732047506)
--(axis cs:26,0.278784093261736)
--(axis cs:25,0.27999135461144)
--(axis cs:24,0.279995848494187)
--(axis cs:23,0.279420136385134)
--(axis cs:22,0.278264237011443)
--(axis cs:21,0.277432611950079)
--(axis cs:20,0.277301988292157)
--(axis cs:19,0.277601771213806)
--(axis cs:18,0.276088429966025)
--(axis cs:17,0.275126420971812)
--(axis cs:16,0.275002956352188)
--(axis cs:15,0.274993269909988)
--(axis cs:14,0.27578080585786)
--(axis cs:13,0.274095527693655)
--(axis cs:12,0.270925392998617)
--(axis cs:11,0.26680326334707)
--(axis cs:10,0.265692734417512)
--(axis cs:9,0.266897081631096)
--(axis cs:8,0.266827652600438)
--(axis cs:7,0.262152958243399)
--(axis cs:6,0.259282092471034)
--(axis cs:5,0.254636287753316)
--(axis cs:4,0.245409408343973)
--(axis cs:3,0.237948204454689)
--(axis cs:2,0.20082696612975)
--cycle;

\addplot [semithick, steelblue31119180]
table {%
2 0.20082696612975
3 0.222108670240376
4 0.228646530252828
5 0.235771530463563
6 0.240357882786421
7 0.243566306988153
8 0.248104860802758
9 0.248332251216549
10 0.247523893744062
11 0.249012925066984
12 0.252801219825953
13 0.255746608303712
14 0.257205311048905
15 0.256374292517812
16 0.256406148585509
17 0.256616535979683
18 0.257594760699388
19 0.259080791797189
20 0.258860785127143
21 0.259087753263177
22 0.259984197845493
23 0.261220124066713
24 0.261918136029744
25 0.262028207798094
26 0.260981027132053
27 0.260728584443999
28 0.261997221535634
29 0.261930642720123
30 0.262041496510816
31 0.262613172258962
32 0.261945510593256
33 0.262869203470662
34 0.263739888988784
35 0.264153973493827
36 0.264424464991216
37 0.265142512497948
38 0.265697725046395
39 0.266039377375934
40 0.266855527882671
41 0.267266078317855
42 0.266737132969363
43 0.267272619321628
44 0.267104133019334
45 0.267398695896642
46 0.267066643083832
47 0.26683475572843
48 0.2670646377617
49 0.266818399497351
50 0.267127437817259
};
    \nextgroupplot[
        title={{\textbf{B4C} \citep{jain2015car}}},
        ylabel={\small{PR-AUC}},
        title style={yshift=-0.2cm},
        ymin=0.45,
        ymax=0.55
    ]
\path [fill=steelblue31119180, fill opacity=0.2]
(axis cs:2,0.472049297050022)
--(axis cs:2,0.472049297050022)
--(axis cs:3,0.476261201376686)
--(axis cs:4,0.481034184563371)
--(axis cs:5,0.482228398719696)
--(axis cs:6,0.479344463150629)
--(axis cs:7,0.47854270436869)
--(axis cs:8,0.478588663804916)
--(axis cs:9,0.478332621535614)
--(axis cs:10,0.479846845030431)
--(axis cs:11,0.480268591565185)
--(axis cs:12,0.481720124114031)
--(axis cs:13,0.484502206630871)
--(axis cs:14,0.486774456943587)
--(axis cs:15,0.490105304636759)
--(axis cs:16,0.492052585886778)
--(axis cs:17,0.491745423657405)
--(axis cs:18,0.492734341346875)
--(axis cs:19,0.49509066973402)
--(axis cs:20,0.494116411552949)
--(axis cs:21,0.494621710796874)
--(axis cs:22,0.495533139519828)
--(axis cs:23,0.495936464959795)
--(axis cs:24,0.496708724883093)
--(axis cs:25,0.495988096660825)
--(axis cs:26,0.494508046637702)
--(axis cs:27,0.494589347443993)
--(axis cs:28,0.494180917937972)
--(axis cs:29,0.494288734680759)
--(axis cs:30,0.493670548192374)
--(axis cs:31,0.494991130571111)
--(axis cs:32,0.495288904440143)
--(axis cs:33,0.495142404321849)
--(axis cs:34,0.496047927061433)
--(axis cs:35,0.495276188033123)
--(axis cs:36,0.494932011555301)
--(axis cs:37,0.495791314663435)
--(axis cs:38,0.496083816534049)
--(axis cs:39,0.497194626363594)
--(axis cs:40,0.497564907246973)
--(axis cs:41,0.497710426531636)
--(axis cs:42,0.497396992848516)
--(axis cs:43,0.498208835378659)
--(axis cs:44,0.498626257115228)
--(axis cs:45,0.499002288890598)
--(axis cs:46,0.499173174819543)
--(axis cs:47,0.499628830553708)
--(axis cs:48,0.49963994022018)
--(axis cs:49,0.499858610816221)
--(axis cs:50,0.500258377256011)
--(axis cs:50,0.530517185596643)
--(axis cs:50,0.530517185596643)
--(axis cs:49,0.530223204246891)
--(axis cs:48,0.530110484828306)
--(axis cs:47,0.530222024648719)
--(axis cs:46,0.529896538009646)
--(axis cs:45,0.52985841251074)
--(axis cs:44,0.529586503211321)
--(axis cs:43,0.529283883172318)
--(axis cs:42,0.528579058206734)
--(axis cs:41,0.529020881685359)
--(axis cs:40,0.528995959546056)
--(axis cs:39,0.528739147345399)
--(axis cs:38,0.527737485488303)
--(axis cs:37,0.527557564154529)
--(axis cs:36,0.526780126970904)
--(axis cs:35,0.52715877825787)
--(axis cs:34,0.527962244261834)
--(axis cs:33,0.527055300455652)
--(axis cs:32,0.527140492802301)
--(axis cs:31,0.526712543239267)
--(axis cs:30,0.525161663449809)
--(axis cs:29,0.525427515690935)
--(axis cs:28,0.524815756293175)
--(axis cs:27,0.524726832586691)
--(axis cs:26,0.524112331154937)
--(axis cs:25,0.524925025353184)
--(axis cs:24,0.525048939577477)
--(axis cs:23,0.523732678625799)
--(axis cs:22,0.522806076120896)
--(axis cs:21,0.521467437632591)
--(axis cs:20,0.520436072041672)
--(axis cs:19,0.520920168575401)
--(axis cs:18,0.51802721113951)
--(axis cs:17,0.51660774494)
--(axis cs:16,0.516364032836584)
--(axis cs:15,0.513651244382871)
--(axis cs:14,0.50941663882394)
--(axis cs:13,0.5063141550256)
--(axis cs:12,0.50263181464334)
--(axis cs:11,0.501048421568952)
--(axis cs:10,0.49982715490037)
--(axis cs:9,0.498365678627083)
--(axis cs:8,0.499782573011329)
--(axis cs:7,0.501324797834761)
--(axis cs:6,0.503736318452476)
--(axis cs:5,0.507423315083751)
--(axis cs:4,0.505138595464774)
--(axis cs:3,0.496598074474388)
--(axis cs:2,0.472049297050022)
--cycle;

\addplot [semithick, steelblue31119180]
table {%
2 0.472049297050022
3 0.486429637925537
4 0.493086390014073
5 0.494825856901724
6 0.491540390801552
7 0.489933751101726
8 0.489185618408122
9 0.488349150081348
10 0.489836999965401
11 0.490658506567068
12 0.492175969378685
13 0.495408180828236
14 0.498095547883763
15 0.501878274509815
16 0.504208309361681
17 0.504176584298703
18 0.505380776243192
19 0.508005419154711
20 0.50727624179731
21 0.508044574214732
22 0.509169607820362
23 0.509834571792797
24 0.510878832230285
25 0.510456561007005
26 0.50931018889632
27 0.509658090015342
28 0.509498337115573
29 0.509858125185847
30 0.509416105821092
31 0.510851836905189
32 0.511214698621222
33 0.51109885238875
34 0.512005085661633
35 0.511217483145497
36 0.510856069263103
37 0.511674439408982
38 0.511910651011176
39 0.512966886854497
40 0.513280433396514
41 0.513365654108497
42 0.512988025527625
43 0.513746359275489
44 0.514106380163274
45 0.514430350700669
46 0.514534856414595
47 0.514925427601214
48 0.514875212524243
49 0.515040907531556
50 0.515387781426327
};
    \nextgroupplot[
        title={{\textbf{ROAD} \citep{singh2022road}}},
        ylabel={\small{PR-AUC}},
        title style={yshift=-0.2cm},
        ymax=0.8,
        ymin=0.2
    ]
\path [fill=steelblue31119180, fill opacity=0.2]
(axis cs:2,0.275117458168383)
--(axis cs:2,0.275117458168383)
--(axis cs:3,0.299131309279073)
--(axis cs:4,0.327027154713565)
--(axis cs:5,0.348331343190482)
--(axis cs:6,0.365331262217934)
--(axis cs:7,0.37701779377515)
--(axis cs:8,0.392452469202735)
--(axis cs:9,0.394536265982295)
--(axis cs:10,0.406819436640144)
--(axis cs:11,0.417704942452028)
--(axis cs:12,0.423955361639726)
--(axis cs:13,0.431453231149226)
--(axis cs:14,0.445770207434816)
--(axis cs:15,0.452742206752329)
--(axis cs:16,0.458391293814999)
--(axis cs:17,0.466381475039342)
--(axis cs:18,0.468769023439449)
--(axis cs:19,0.47527522999373)
--(axis cs:20,0.47496052442839)
--(axis cs:21,0.483302360218524)
--(axis cs:22,0.482551878874507)
--(axis cs:23,0.493938066885996)
--(axis cs:24,0.494879067600715)
--(axis cs:25,0.49722744215485)
--(axis cs:26,0.498928557588683)
--(axis cs:27,0.504679817827822)
--(axis cs:28,0.509337051021471)
--(axis cs:29,0.508729026365266)
--(axis cs:30,0.510887727587358)
--(axis cs:31,0.515269090353243)
--(axis cs:32,0.516193283278065)
--(axis cs:33,0.523002974354718)
--(axis cs:34,0.5294977098895)
--(axis cs:35,0.531325459536842)
--(axis cs:36,0.535831587746147)
--(axis cs:37,0.53759735167599)
--(axis cs:38,0.538815846118453)
--(axis cs:39,0.539126663983706)
--(axis cs:40,0.549960344295658)
--(axis cs:41,0.550235141344492)
--(axis cs:42,0.551503733911316)
--(axis cs:43,0.554787416235092)
--(axis cs:44,0.552842689191243)
--(axis cs:45,0.551202459442872)
--(axis cs:46,0.553795949443208)
--(axis cs:47,0.558824036137439)
--(axis cs:48,0.558321515966149)
--(axis cs:49,0.559992478578359)
--(axis cs:50,0.560753277949136)
--(axis cs:50,0.739484979625226)
--(axis cs:50,0.739484979625226)
--(axis cs:49,0.738445089927039)
--(axis cs:48,0.736430311305528)
--(axis cs:47,0.73654185852547)
--(axis cs:46,0.730993136169602)
--(axis cs:45,0.728010530954271)
--(axis cs:44,0.72925754556456)
--(axis cs:43,0.730599110556421)
--(axis cs:42,0.72644468415384)
--(axis cs:41,0.7243506044773)
--(axis cs:40,0.723116326619058)
--(axis cs:39,0.711199292352092)
--(axis cs:38,0.710384985480261)
--(axis cs:37,0.708539041710656)
--(axis cs:36,0.706043218285911)
--(axis cs:35,0.700726059860097)
--(axis cs:34,0.698193608849744)
--(axis cs:33,0.690866105435825)
--(axis cs:32,0.683476791294466)
--(axis cs:31,0.682195793690637)
--(axis cs:30,0.677323439519928)
--(axis cs:29,0.674701162060829)
--(axis cs:28,0.67466380122864)
--(axis cs:27,0.668954249414169)
--(axis cs:26,0.66225025901629)
--(axis cs:25,0.659797532353257)
--(axis cs:24,0.656286490981316)
--(axis cs:23,0.653939018278663)
--(axis cs:22,0.64063048869585)
--(axis cs:21,0.640480437120543)
--(axis cs:20,0.630585188448819)
--(axis cs:19,0.629782142171948)
--(axis cs:18,0.621359802630562)
--(axis cs:17,0.617290421665991)
--(axis cs:16,0.606991530729663)
--(axis cs:15,0.598921927524825)
--(axis cs:14,0.589264514800586)
--(axis cs:13,0.571840147998813)
--(axis cs:12,0.562818705454511)
--(axis cs:11,0.55581548392746)
--(axis cs:10,0.543613459714645)
--(axis cs:9,0.530493025248699)
--(axis cs:8,0.528564828077903)
--(axis cs:7,0.510183824350839)
--(axis cs:6,0.498245465522197)
--(axis cs:5,0.480065744867161)
--(axis cs:4,0.453490057613887)
--(axis cs:3,0.415080439351548)
--(axis cs:2,0.275117458168383)
--cycle;

\addplot [semithick, steelblue31119180]
table {%
2 0.275117458168383
3 0.357105874315311
4 0.390258606163726
5 0.414198544028822
6 0.431788363870066
7 0.443600809062995
8 0.460508648640319
9 0.462514645615497
10 0.475216448177395
11 0.486760213189744
12 0.493387033547118
13 0.501646689574019
14 0.517517361117701
15 0.525832067138577
16 0.532691412272331
17 0.541835948352666
18 0.545064413035006
19 0.552528686082839
20 0.552772856438605
21 0.561891398669534
22 0.561591183785179
23 0.573938542582329
24 0.575582779291016
25 0.578512487254054
26 0.580589408302486
27 0.586817033620995
28 0.592000426125055
29 0.591715094213048
30 0.594105583553643
31 0.59873244202194
32 0.599835037286266
33 0.606934539895271
34 0.613845659369622
35 0.61602575969847
36 0.620937403016029
37 0.623068196693323
38 0.624600415799357
39 0.625162978167899
40 0.636538335457358
41 0.637292872910896
42 0.638974209032578
43 0.642693263395756
44 0.641050117377902
45 0.639606495198571
46 0.642394542806405
47 0.647682947331454
48 0.647375913635838
49 0.649218784252699
50 0.650119128787181
};
    \end{groupplot}
    \end{tikzpicture}
    \caption{Effects of additional dependent \textcolor{black}{$\theta_{LL}$ parameter} samples  from the same chain of the top-performing LL--HMC configuration for the OOD min scenario for different 100 random seeds. The light blue area illustrates the standard deviation of the cumulative of the PR-AUC for each additional sample.}
    \label{fig:dependent_samples_ood}
\end{figure}
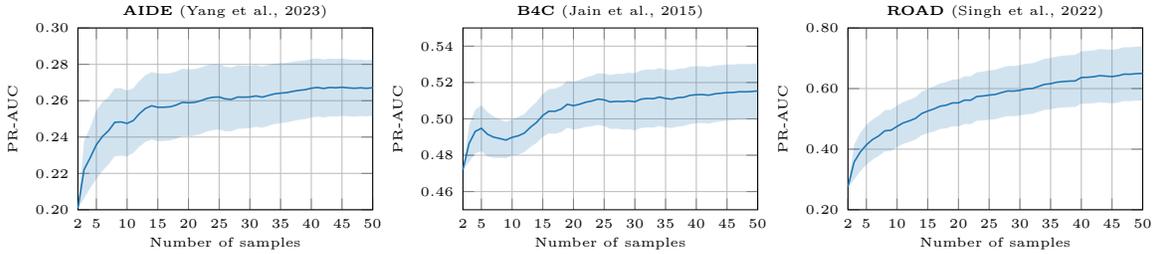

 
\paragraph{\textbf{Influence of additional predictions or samples.}} Figure \ref{fig:ood_comparison} visualizes the influence of additional samples or predictions for the best performing hyperparameter configurations from Table \ref{tab:road_ood_exp_hp} for 10 random seeds for the different numbers of samples or predictions. The BBB--LL results for the OOD min scenarios for the AIDE and B4C datasets show a slight improvement with additional samples, but this trend is not visible for the other scenarios or for the ROAD dataset. Opposed to the results in Table \ref{tab:road_ood_exp}, LL--HMC for 10 different random seeds for each number of samples does not yield the highest performance on average for all scenarios. Similar to the in-distribution results, there is no single LL--PDL method consistently performs better in all scenarios.  

\paragraph{\textbf{Comparing the best hyperparameter configurations. }} Figure \ref{fig:placeholder_groupplot} shows the average PR-AUC across 100 random seeds for the best hyperparameter configurations from Table \ref{tab:road_ood_exp_hp}. The OOD detection performance of the PE--LL and SE is relatively stable across scenarios, with average PR-AUC values comparable to those reported in Table \ref{tab:road_ood_exp}. In contrast, BBB--LL and LL--HMC exhibit greater variability in average performance across seeds and configurations depending on the OOD scenario. Overall, evaluating the top-performing configurations across multiple seeds suggests that no single LL--PDL method consistently outperforms the others.

\paragraph{\textbf{Influence of additional dependent LL--HMC samples.}}
Figure \ref{fig:dependent_samples_ood} shows \textcolor{black}{the relation between the complexity (additional $\theta_{LL}$ parameter samples)} from the same HMC chain and the PR-AUC performance. The results are based on the best-performing grid search configuration. Compared to the modest F1-score gains in Figure \ref{fig:dependent_samples}, PR-AUC improves slightly more with additional same-chain samples in the OOD min scenario. However, the PR-AUC improvement stabilizes with more samples. 




\begin{table*}[t]
\centering
\caption{Average OOD top-performance and two standard errors of the mean across five random seeds for LL--HMC with multiple starting positions or chains. S=Starting Positions, C=Chains.}

\resizebox{\linewidth}{!}{
\begin{tabular}{lcc|c|c|c|c|c|c||c|c|c|c|c|c||c|c|c}
\toprule
\multicolumn{3}{c|}{} & \multicolumn{6}{c||}{\textbf{AIDE} \citep{yang2023aide}} & \multicolumn{6}{c||}{\textbf{B4C} \citep{jain2015car}} & \multicolumn{3}{c}{\textbf{ROAD} \citep{singh2022road}}  \\ \cmidrule{4-18}
\multicolumn{3}{c|}{} & \multicolumn{3}{c|}{\textbf{OOD min}} & \multicolumn{3}{c||}{\textbf{OOD max}} & \multicolumn{3}{c|}{\textbf{OOD min}} & \multicolumn{3}{c||}{\textbf{OOD max}} & \multicolumn{3}{c}{\textbf{OOD min}} \\
 \multicolumn{3}{c|}{} & \begin{tabular}{@{}c@{}}\textbf{ROC-}\\ \textbf{AUC} ($\uparrow$)\end{tabular} & 
\begin{tabular}{@{}c@{}}\textbf{PR-}\\\textbf{AUC} ($\uparrow$)\end{tabular} & 
\begin{tabular}{@{}c@{}}\textbf{FPR95}\\($\downarrow$)\end{tabular} & \begin{tabular}{@{}c@{}}\textbf{ROC-}\\ \textbf{AUC} ($\uparrow$)\end{tabular} & 
\begin{tabular}{@{}c@{}}\textbf{PR-}\\\textbf{AUC} ($\uparrow$)\end{tabular} & 
\begin{tabular}{@{}c@{}}\textbf{FPR95}\\($\downarrow$)\end{tabular} & \begin{tabular}{@{}c@{}}\textbf{ROC-}\\ \textbf{AUC} ($\uparrow$)\end{tabular} & 
\begin{tabular}{@{}c@{}}\textbf{PR-}\\\textbf{AUC} ($\uparrow$)\end{tabular} & 
\begin{tabular}{@{}c@{}}\textbf{FPR95}\\($\downarrow$)\end{tabular} & \begin{tabular}{@{}c@{}}\textbf{ROC-}\\ \textbf{AUC} ($\uparrow$)\end{tabular} & 
\begin{tabular}{@{}c@{}}\textbf{PR-}\\\textbf{AUC} ($\uparrow$)\end{tabular} & 
\begin{tabular}{@{}c@{}}\textbf{FPR95}\\($\downarrow$)\end{tabular} & \begin{tabular}{@{}c@{}}\textbf{ROC-}\\ \textbf{AUC} ($\uparrow$)\end{tabular} & 
\begin{tabular}{@{}c@{}}\textbf{PR-}\\\textbf{AUC} ($\uparrow$)\end{tabular} & 
\begin{tabular}{@{}c@{}}\textbf{FPR95}\\($\downarrow$)\end{tabular}   \\
\midrule

  \multirow{4}{*}{\textbf{LL--HMC}}& \multirow{2}{*}{\small{S=1}} &  \small{C=1} & 0.90$\pm$0.01 & 0.52$\pm$0.01 & 0.21$\pm$0.02	&	0.95$\pm$0.00 & 0.99$\pm$0.00 & 0.14$\pm$0.01	&	0.83$\pm$0.01 & 0.70$\pm$0.01 & 0.41$\pm$0.03	&	0.91$\pm$0.01 & 0.97$\pm$0.00 & 0.39$\pm$0.05	&	0.86$\pm$0.01 & 0.73$\pm$0.02 & 0.36$\pm$0.05 \\
 &    &   \small{C=2}   & 0.91$\pm$0.03 & 0.58$\pm$0.10 & 0.18$\pm$0.05	&	0.95$\pm$0.00 & 0.98$\pm$0.00 & 0.15$\pm$0.00	&	0.85$\pm$0.04 & 0.73$\pm$0.05 & 0.36$\pm$0.08	&	0.85$\pm$0.04 & 0.73$\pm$0.05 & 0.36$\pm$0.08	&	0.87$\pm$0.01 & 0.72$\pm$0.01 & 0.31$\pm$0.04  \\  \cmidrule{2-18}
& \multirow{2}{*}{\small{S=2}} & \small{C=1} &  0.64$\pm$0.03 & 0.35$\pm$0.02 & 0.97$\pm$0.03	&	0.89$\pm$0.03 & 0.97$\pm$0.01 & 0.40$\pm$0.18	&	0.85$\pm$0.01 & 0.70$\pm$0.02 & 0.34$\pm$0.03	&	0.90$\pm$0.01 & 0.96$\pm$0.01 & 0.42$\pm$0.06	&	0.38$\pm$0.05 & 0.39$\pm$0.01 & 1.00$\pm$0.00  \\
& &  \small{C=2}  &  0.66$\pm$0.07 & 0.35$\pm$0.01 & 0.89$\pm$0.21	&	0.88$\pm$0.01 & 0.97$\pm$0.00 & 0.42$\pm$0.06	&	0.85$\pm$0.00 & 0.70$\pm$0.02 & 0.34$\pm$0.03	&	0.89$\pm$0.01 & 0.96$\pm$0.00 & 0.39$\pm$0.06	&	0.50$\pm$0.06 & 0.42$\pm$0.01 & 0.99$\pm$0.01 \\ 
\bottomrule
\end{tabular}
}
\label{tab:ood_multi_llhmc}
\end{table*}

\paragraph{\textbf{Multiple chains and starting positions.}}
Table \ref{tab:ood_multi_llhmc} reports OOD detection performance using the top-performing grid search configuration across five seeds, comparing different numbers of HMC chains (C) and starting positions (S). LL--HMC with a single starting position consistently yields strong OOD detection performance across the datasets, regardless of the number of chains. In contrast, using two starting positions results in less consistent performance. Specifically, in the AIDE OOD min and ROAD OOD min scenarios, LL--HMC with two starting positions produces higher FPR95 values, indicating reduced reliability in distinguishing in- and out-of-distribution instances. This suggests LL--HMC’s performance is sensitive to the learned latent representations, especially in how they guide the initialization and exploration of posterior modes. As with the in-distribution classification performance, increasing the number of chains or starting positions does not consistently improve OOD detection, and in some cases degrades performance.

\section{Discussion and Conclusion}
We evaluated and explored the effectiveness of applying Hamiltonian Monte Carlo (HMC) sampling to the final layer of deep neural networks for video-based driver action and intention recognition. Applying HMC to sample the parameters of the last layer reduces the required computational cost compared to sampling the entire network. While last layer HMC (LL--HMC) benefits from the Markov-Chain Monte Carlo sampling ability to explore complex and potentially multi-modal posterior distributions for the last layer's parameters, its reliance on fixed pre-trained representations comes at the cost of not capturing the uncertainty inherent in the feature extraction layers of a deep neural network (DNN). Furthermore, as already indicated by the toy example, increasing the number of posterior samples improves the uncertainty estimations but saturates after a certain point.

Compared to the softmax-based baseline, the last layer probabilistic deep learning (PDL)  approaches demonstrated improved performance under specific hyperparameter settings. However, the classification and OOD detection results across grid searches yielded significantly lower performance dropped compared to the best performance for a single grid search. The effect of additional predictions by evaluating the best performing hyperparameter configuration across 100 random seeds for each last layer PDL method and dataset showed that LL--HMC yields competitive results. Additional \textcolor{black}{last layer parameter} samples or predictions\textcolor{black}{, which increases the computational complexity,} did not consistently enhance the in-distribution or OOD performance for the included last layer PDL approaches. \textcolor{black}{Future work could further examine the effects of not only sampling the last layer of the DNN but extracting the latent representation from earlier stages of the network to apply the LL–PDL methods to the subsequent remaining layers. Alternatively, it would be interesting to assess the effects of extending the single last layer to multiple layers for generating predictions and uncertainty estimates. } 

While driving action and intention recognition are examples of realistic safety-critical applications, last layer PDL approaches can also be applied in other domains. For example, to support human-robot interaction through action and intention recognition \cite{qi2024computer}, elderly care \cite{pereira2009elder}, predictive maintenance \cite{benker2021utilizing,nguyen2022probabilistic}, medical imaging \cite{zou2023review}, and fraud detection \cite{habibpour2023uncertainty}. These tasks require different modeling approaches, which could provide deeper insights into how the underlying model representations affect the last layer uncertainty estimation performance.


None of the last layer PDL methods were able to consistently identify all OOD instances when the most or least prevailing driving maneuvers were removed from the datasets. This raises broader considerations for the design of safety-critical systems such as advanced driver assistance and autonomous driving technologies. For instance, when a model predicts two maneuvers (e.g., ‘\textit{left turn}’ vs. ‘\textit{U-turn}’) with similar high confidence and low uncertainty, the system defaults to pick the class with the highest assigned confidence, despite the potential overlap. In such a scenario, it could be interesting to see whether predicting sets \citep{hullermeier2022quantification,wang2024credal} of potential maneuvers improves the performance. More broadly, if PDL methods are unable to reliably flag all OOD scenarios, we should also consider the potential outcome of the recommendations given the current driving scene.  By leveraging a more decision-theoretic approach that incorporates generic task knowledge (e.g., penalizing intended maneuvers that potentially harm other road users), we might be able to produce better and safer recommendations \citep{cobb2018loss}.

\bibliographystyle{ACM-Reference-Format}
\bibliography{sample-base}

\clearpage

\appendix
\section*{Appendices}
\section{Thinning effects for the regression toy example}

\textcolor{black}{To further illustrate practical considerations when using LL--HMC, we visualize the effects of thinning the number of retained last-layer samples on the predictive uncertainty for the toy regression example \cite{lippe2022uva} as described in \cref{sec:3.1}. Because HMC samples from a Markov chain, successive draws can be auto-correlated, meaning that the effective number of independent draws is smaller than the nominal number of stored samples. This effect is commonly quantified using the effective sample size (ESS, \cref{eq:ess}), which decreases as within-chain autocorrelation increases. One simple mitigation is \emph{thinning}, i.e., retaining only every $K$-th post-burn-in sample to reduce autocorrelation among the stored parameter samples.} 

\begin{figure}[t]
    \centering
      \begin{subfigure}[b]{0.49\linewidth}
        \centering
        \includegraphics[width=\linewidth, trim=2.85cm 1.2cm 1.5cm 1.4cm, clip]{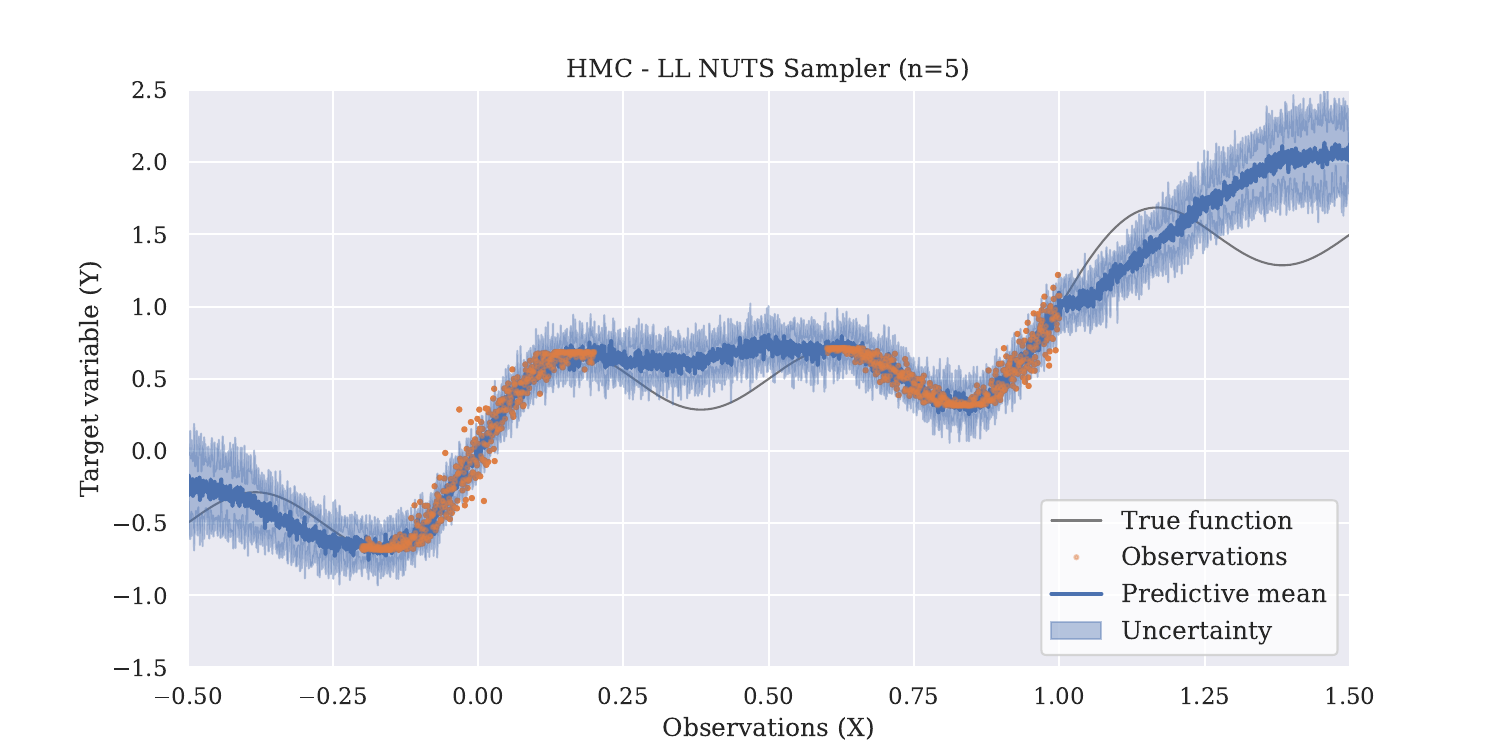} 
       \captionsetup{font=tiny}
         \caption{LL--HMC (N=5)}
    \end{subfigure}
    \hfill
      \begin{subfigure}[b]{0.49\linewidth}
        \centering
        \includegraphics[width=\linewidth, trim=2.85cm 1.2cm 1.5cm 1.4cm, clip]{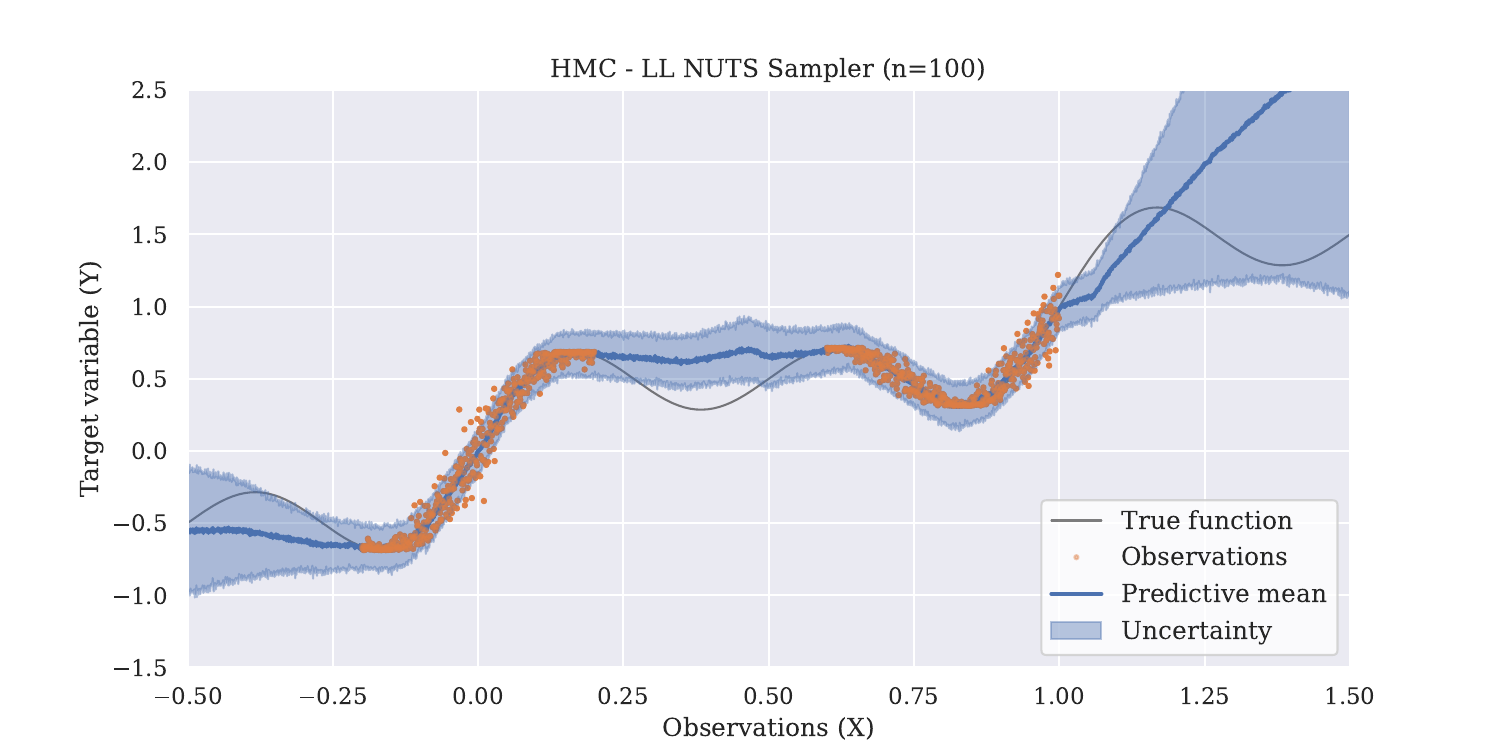} 
       \captionsetup{font=tiny}
         \caption{LL--HMC (N=100)}
    \end{subfigure}
    \hfill
      \begin{subfigure}[b]{0.49\linewidth}
        \centering
        \includegraphics[width=\linewidth, trim=2.85cm 1.2cm 1.5cm 1.4cm, clip]{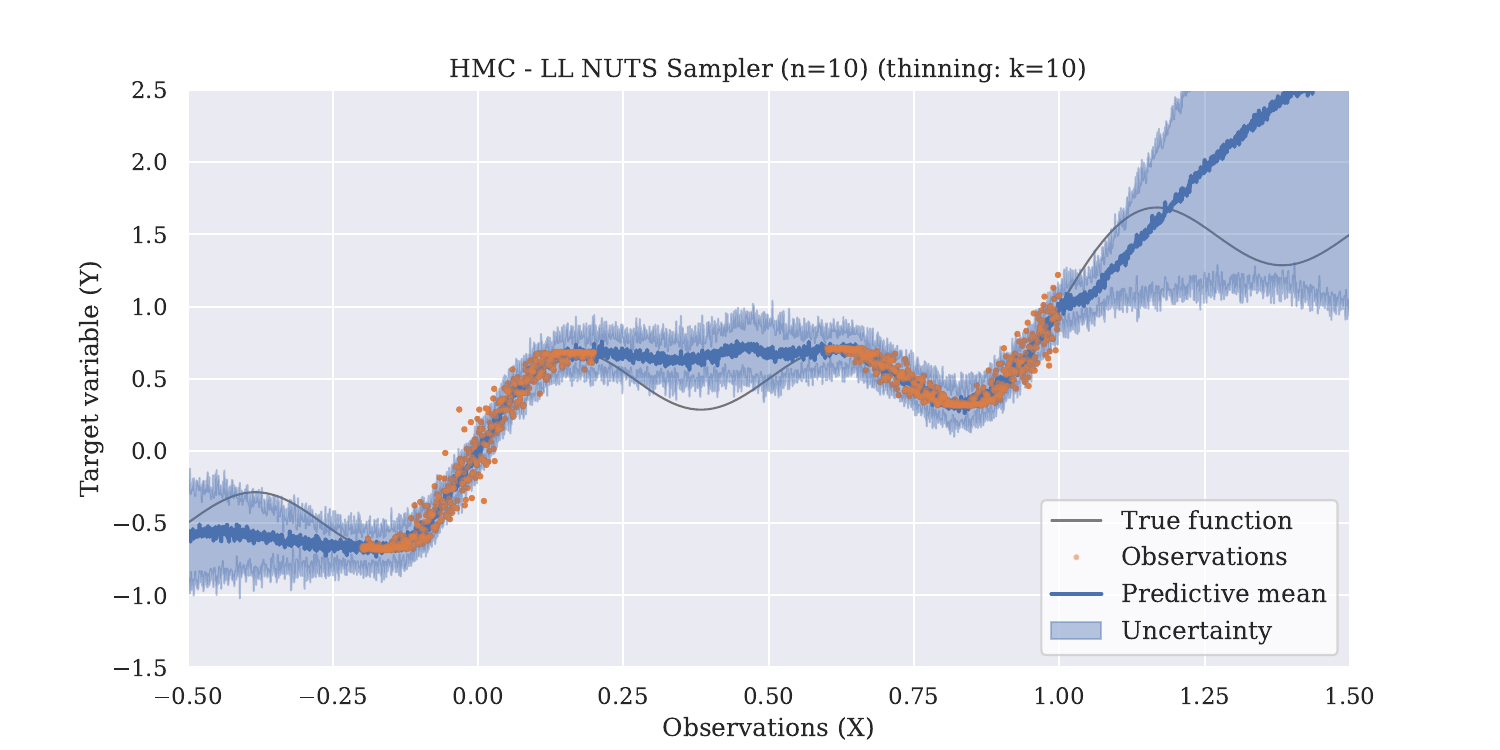} %
       \captionsetup{font=tiny}
        \caption{LL--HMC (N=10, K=10)}
    \end{subfigure}
    \hfill
      \begin{subfigure}[b]{0.49\linewidth}
        \centering
        \includegraphics[width=\linewidth, trim=2.85cm 1.2cm 1.5cm 1.4cm, clip]{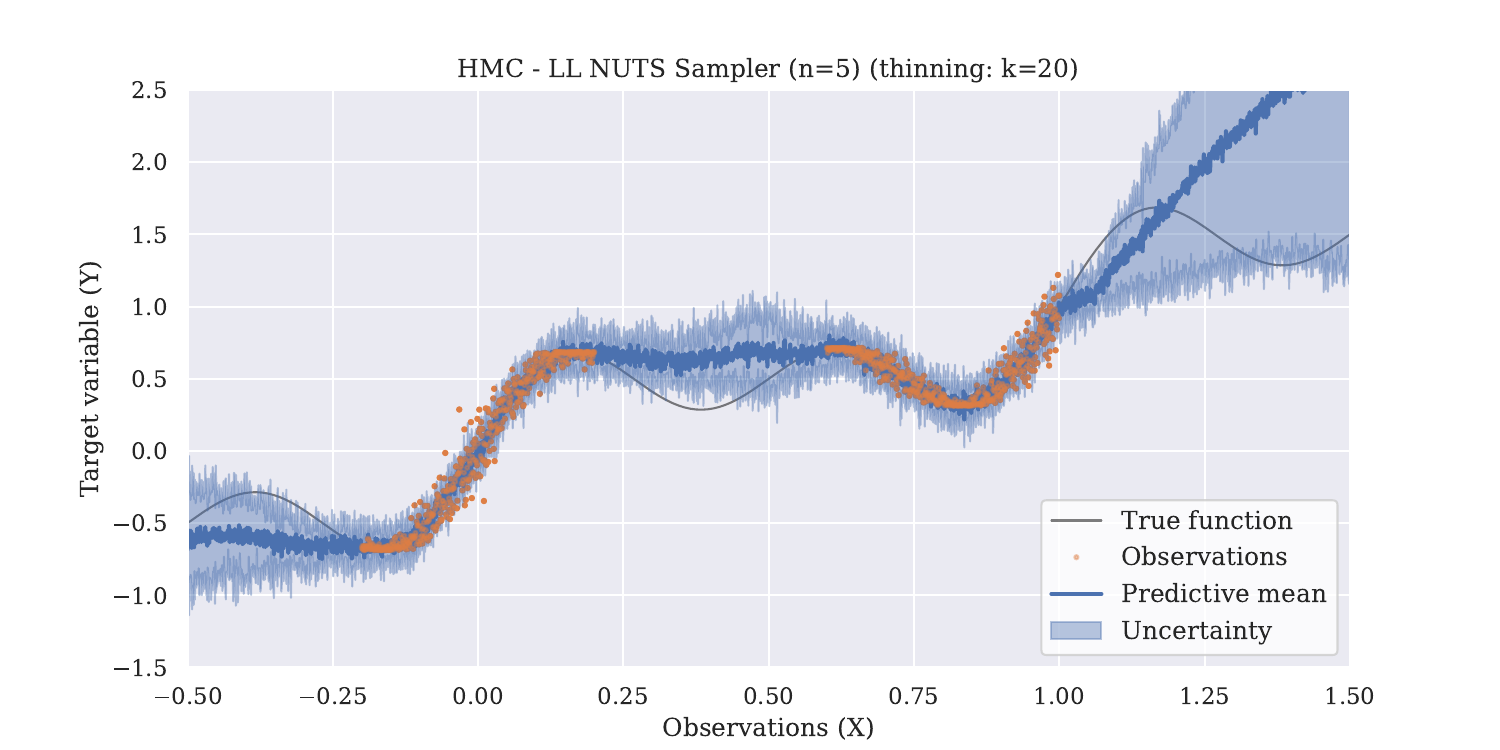} %
       \captionsetup{font=tiny}
        \caption{LL--HMC (N=5, K=20)}
    \end{subfigure}

    \caption{Thinning effects on the uncertainty estimation for a regression toy example for the last layer Hamiltonian Monte-Carlo method. The y-axis is the target variable, the x-axis the observations. The gray line is the original function, the orange dots represent the noisy observations, the blue line the mean prediction, and the light blue area the uncertainty estimation. N=number of parameter samples, K=the thinning interval of samples to keep.}
        \label{fig:thin_llhmc}
\end{figure}

\textcolor{black}{Figure~\ref{fig:thin_llhmc} visualizes the thinning effects. The upper row shows the uncertainty estimates for LL--HMC with either 5 or 100 retained samples (after 100 burn-in samples). The bottom row visualizes the effects of thinning: retaining every $10^{th}$ or $20^{th}$  sample (yielding 5 or 10 samples from the original 100 samples). Compared to the baseline of 100 retained samples (top right), both thinning examples exhibit similar uncertainty bands but noisier predictive means. Compared to the baseline of only 5 samples without thinning (top left), the thinned examples in the bottom row demonstrate improved uncertainty estimates, reflecting the benefits of broader chain exploration. However, given that the additional computational cost of retaining more last-layer samples is relatively negligible compared to the backbone forward pass, thinning may be unnecessary in practice.}

\clearpage
\section{Intermediate latent representation performance}

\begin{figure}[h]
        \centering
        \includegraphics[trim={0.4cm 9.2cm 18.0cm 0cm},clip, width=0.50\textwidth, keepaspectratio]{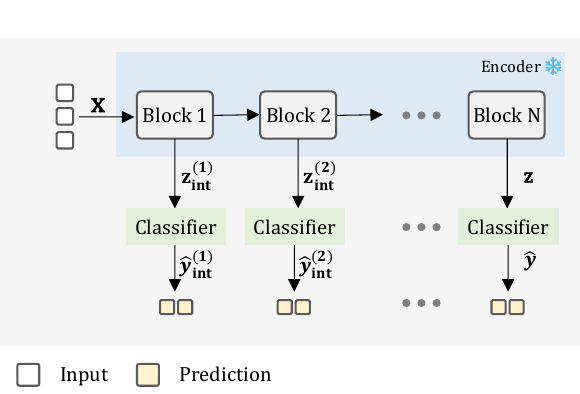}
        \caption{Schematic intermediate representation classification overview. The intermediate latent representation $z_{int}^{i}$ can be extracted at different depths from the encoder, and used to produce intermediate classifications. }\label{fig:intermed}
    \end{figure}


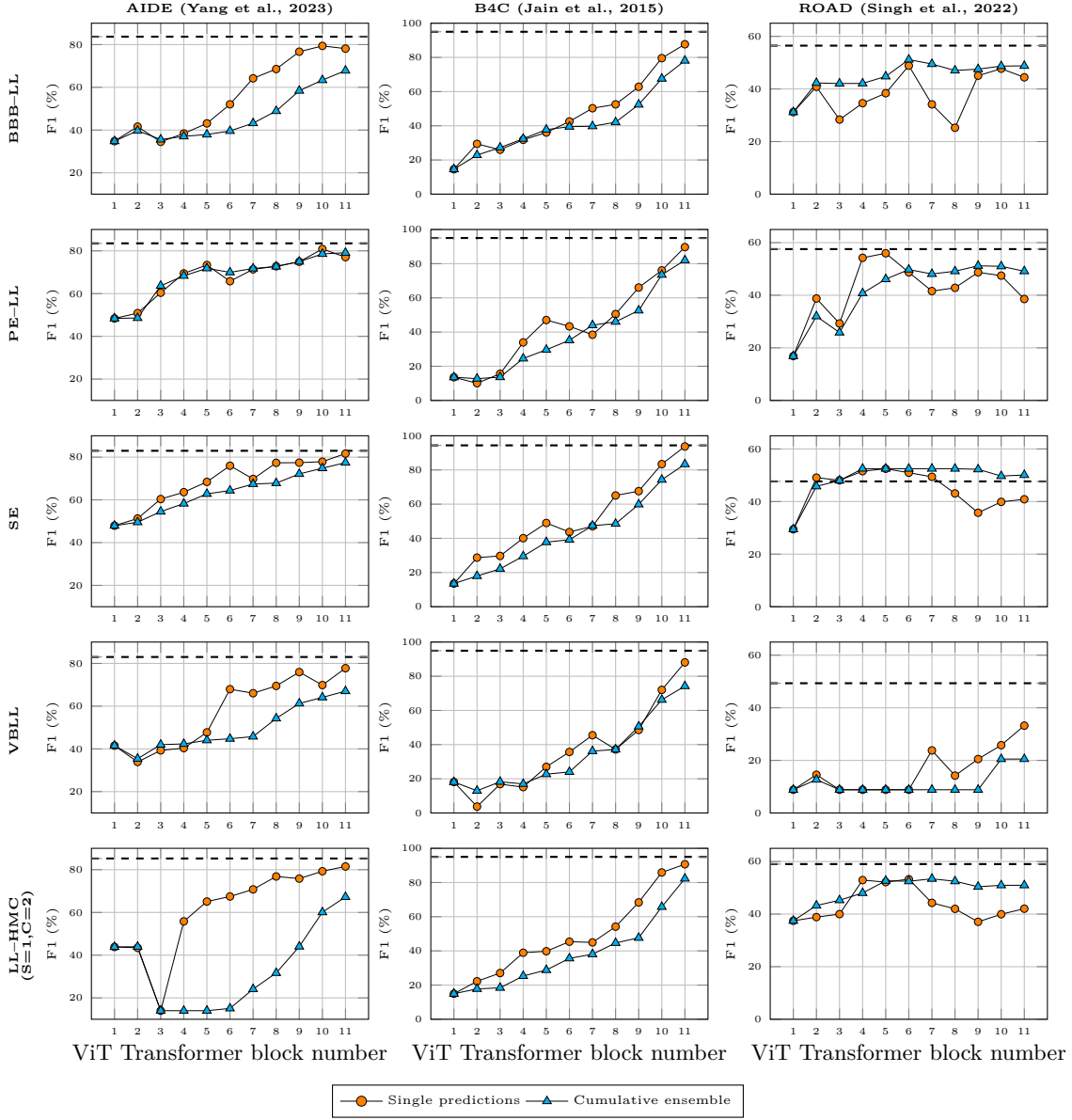
\begin{figure*}[h]
\begin{tikzpicture}
\begin{groupplot}[
    group style={
        group size=3 by 5,  
        horizontal sep=0.88cm,
        vertical sep=0.5cm,  
    },
    width=5.5cm,  
    height=4.00cm,  
    grid=major,
    xlabel={ViT Transformer block number},
    xtick={1,2,3,4,5,6,7,8,9,10,11,12(LL)},
    tick label style={font=\small},
    label style={font=\small},
    y tick label style={
        /pgf/number format/fixed,
        /pgf/number format/precision=0,
        /pgf/number format/fixed zerofill=true,
    },
    tick label style={font=\fontsize{4}{4}\selectfont},
    ylabel style={yshift=-0.1cm},  
    xlabel style={yshift=0.1cm},  
    title style={}, 
]

\node[font=\bfseries\tiny, rotate=90] at ($(group c1r1.west)+(-1.1cm,-0.0)$) {BBB--LL};
\node[font=\bfseries\tiny, rotate=90] at ($(group c1r1.west)+(-1.1cm,-2.9cm)$) {PE--LL};
\node[font=\bfseries\tiny, rotate=90] at ($(group c1r1.west)+(-1.1cm,-5.8cm)$) {SE};
\node[font=\bfseries\tiny, rotate=90] at ($(group c1r1.west)+(-1.1cm,-8.8cm)$) {VBLL};
\node[font=\bfseries\tiny, rotate=90] at ($(group c1r1.west)+(-1.1cm,-11.7cm)$) {LL--HMC};
\node[font=\bfseries\tiny, rotate=90] at ($(group c1r1.west)+(-0.9cm,-11.7cm)$) {(S=1,C=2)};

\nextgroupplot[
    title={},  
    ylabel={\small{F1 (\%)}},
    xtick={1,2,3,4,5,6,7,8,9,10,11},
    xlabel=\empty, 
    ymin=10,
    ymax=90, 
     extra y ticks={83.70},
     extra y tick labels={}, 
    extra y tick style={grid=major, grid style={thick, dashed, black}}
]

\addplot[mark=*, thin, line width=0.25pt, mark options={fill=orange}, mark size=1.5pt] coordinates {
(1, 34.80756)
(2, 41.61931)
(3, 34.48399)
(4, 38.3959)
(5, 43.11464)
(6, 52.05468)
(7, 64.20065)
(8, 68.55294)
(9, 76.66792)
(10, 79.38153)
(11, 78.11334)
};

\addplot[mark=triangle*, thin, line width=0.25pt, mark options={fill=cyan}, mark size=2pt] coordinates {
(1, 34.80756)
(2, 39.70194)
(3, 35.578)
(4, 37.04696)
(5, 37.90946)
(6, 39.58183)
(7, 43.2181)
(8, 48.90854)
(9, 58.40118)
(10, 63.37307)
(11, 67.82723)
};


\nextgroupplot[
    title={},  
    ylabel={\small{F1 (\%)}},
    xtick={1,2,3,4,5,6,7,8,9,10,11},
    xlabel=\empty, 
    ymin=0,
    ymax=100, 
     extra y ticks={95.00},
     extra y tick labels={}, 
    extra y tick style={grid=major, grid style={thick, dashed, black}}
]

\addplot[mark=*, thin, line width=0.25pt, mark options={fill=orange}, mark size=1.5pt] coordinates {
(1, 14.56977)
(2, 29.41549)
(3, 25.97255)
(4, 31.81229)
(5, 36.04545)
(6, 42.5057)
(7, 50.27513)
(8, 52.52793)
(9, 62.81554)
(10, 79.58001)
(11, 87.68717)
};

\addplot[mark=triangle*, thin, line width=0.25pt, mark options={fill=cyan}, mark size=2pt] coordinates {
(1, 14.56977)
(2, 22.86667)
(3, 27.24107)
(4, 32.42375)
(5, 37.75556)
(6, 39.47939)
(7, 39.75452)
(8, 42.07074)
(9, 52.39683)
(10, 67.56133)
(11, 78.10365)
};


\nextgroupplot[
    title={},  
    ylabel={\small{F1 (\%)}},
    xtick={1,2,3,4,5,6,7,8,9,10,11},
    xlabel=\empty, 
    ymin=0,
    ymax=65, 
     extra y ticks={56.47},
     extra y tick labels={}, 
    extra y tick style={grid=major, grid style={thick, dashed, black}}
]

\addplot[mark=*, thin, line width=0.25pt, mark options={fill=orange}, mark size=1.5pt] coordinates {
(1, 31.19747)
(2, 40.82603)
(3, 28.35099)
(4, 34.59857)
(5, 38.38608)
(6, 48.86124)
(7, 34.14086)
(8, 25.25275)
(9, 45.05373)
(10, 47.70917)
(11, 44.42916)
};

\addplot[mark=triangle*, thin, line width=0.25pt, mark options={fill=cyan}, mark size=2pt] coordinates {
(1, 31.19747)
(2, 42.28817)
(3, 42.06203)
(4, 42.11146)
(5, 44.76579)
(6, 51.17573)
(7, 49.48107)
(8, 47.02707)
(9, 47.53458)
(10, 48.68311)
(11, 48.78504)
};



\nextgroupplot[
    title={},  
    ylabel={\small{F1 (\%)}},
    xtick={1,2,3,4,5,6,7,8,9,10,11},
    xlabel=\empty, 
    ymin=10,
    ymax=90, 
     extra y ticks={83.52},
     extra y tick labels={}, 
    extra y tick style={grid=major, grid style={thick, dashed, black}}
]

\addplot[mark=*, thin, line width=0.25pt, mark options={fill=orange}, mark size=1.5pt] coordinates {
(1, 48.37391)
(2, 50.85772)
(3, 60.40841)
(4, 69.44792)
(5, 73.36362)
(6, 65.7891)
(7, 71.35559)
(8, 72.77942)
(9, 74.89236)
(10, 80.8762)
(11, 77.06119)
};

\addplot[mark=triangle*, thin, line width=0.25pt, mark options={fill=cyan}, mark size=2pt] coordinates {
(1, 48.37391)
(2, 48.58562)
(3, 63.65119)
(4, 68.31964)
(5, 71.82754)
(6, 69.94979)
(7, 71.74029)
(8, 72.66157)
(9, 74.97421)
(10, 78.57321)
(11, 79.11401)
};


\nextgroupplot[
    title={},  
    ylabel={\small{F1 (\%)}},
    xtick={1,2,3,4,5,6,7,8,9,10,11},
    xlabel=\empty, 
    ymin=0,
    ymax=100, 
     extra y ticks={95.03},
     extra y tick labels={}, 
    extra y tick style={grid=major, grid style={thick, dashed, black}}
]

\addplot[mark=*, thin, line width=0.25pt, mark options={fill=orange}, mark size=1.5pt] coordinates {
(1, 13.57317)
(2, 10.06396)
(3, 15.59454)
(4, 33.92625)
(5, 47.02066)
(6, 43.34239)
(7, 38.43244)
(8, 50.50615)
(9, 66.04923)
(10, 76.06624)
(11, 89.69433)
};

\addplot[mark=triangle*, thin, line width=0.25pt, mark options={fill=cyan}, mark size=2pt] coordinates {
(1, 13.57317)
(2, 12.75251)
(3, 13.62854)
(4, 24.52641)
(5, 29.65018)
(6, 35.17379)
(7, 44.0076)
(8, 46.02523)
(9, 52.62523)
(10, 73.50839)
(11, 81.99567)
};


\nextgroupplot[
    title={},  
    ylabel={\small{F1 (\%)}},
    xtick={1,2,3,4,5,6,7,8,9,10,11},
    xlabel=\empty, 
    ymin=0,
    ymax=65, 
     extra y ticks={57.54},
     extra y tick labels={}, 
    extra y tick style={grid=major, grid style={thick, dashed, black}}
]

\addplot[mark=*, thin, line width=0.25pt, mark options={fill=orange}, mark size=1.5pt] coordinates {
(1, 16.86518)
(2, 38.75458)
(3, 29.26438)
(4, 54.22914)
(5, 55.96379)
(6, 48.70592)
(7, 41.58425)
(8, 42.82384)
(9, 48.70138)
(10, 47.45985)
(11, 38.56271)
};

\addplot[mark=triangle*, thin, line width=0.25pt, mark options={fill=cyan}, mark size=2pt] coordinates {
(1, 16.86518)
(2, 31.9888)
(3, 25.83333)
(4, 40.77227)
(5, 46.10682)
(6, 49.7619)
(7, 48.08367)
(8, 49.13485)
(9, 51.18355)
(10, 51.00361)
(11, 49.13485)
};



\nextgroupplot[
    title={},  
    ylabel={\small{F1 (\%)}},
    xtick={1,2,3,4,5,6,7,8,9,10,11},
    xlabel=\empty, 
    ymin=10,
    ymax=90, 
     extra y ticks={82.97},
     extra y tick labels={}, 
    extra y tick style={grid=major, grid style={thick, dashed, black}}
]

\addplot[mark=*, thin, line width=0.25pt, mark options={fill=orange}, mark size=1.5pt] coordinates {
(1, 47.92156)
(2, 51.28769)
(3, 60.37662)
(4, 63.59172)
(5, 68.41436)
(6, 75.98272)
(7, 69.79106)
(8, 77.33989)
(9, 77.41162)
(10, 77.86767)
(11, 81.66656)
};

\addplot[mark=triangle*, thin, line width=0.25pt, mark options={fill=cyan}, mark size=2pt] coordinates {
(1, 47.92156)
(2, 49.54622)
(3, 54.51997)
(4, 58.25786)
(5, 62.79981)
(6, 64.37686)
(7, 67.41963)
(8, 67.84083)
(9, 72.13373)
(10, 74.82318)
(11, 77.42545)
};


\nextgroupplot[
    title={},  
    ylabel={\small{F1 (\%)}},
    xtick={1,2,3,4,5,6,7,8,9,10,11},
    xlabel=\empty, 
    ymin=0,
    ymax=100, 
     extra y ticks={94.39},
     extra y tick labels={}, 
    extra y tick style={grid=major, grid style={thick, dashed, black}}
]

\addplot[mark=*, thin, line width=0.25pt, mark options={fill=orange}, mark size=1.5pt] coordinates {
(1, 13.53434)
(2, 28.70708)
(3, 29.70619)
(4, 40.08207)
(5, 48.9715)
(6, 43.69256)
(7, 47.0734)
(8, 65.07611)
(9, 67.60291)
(10, 83.41823)
(11, 93.741)
};

\addplot[mark=triangle*, thin, line width=0.25pt, mark options={fill=cyan}, mark size=2pt] coordinates {
(1, 13.53434)
(2, 17.95271)
(3, 22.14991)
(4, 29.53797)
(5, 37.75152)
(6, 39.19328)
(7, 47.39876)
(8, 48.53209)
(9, 59.84462)
(10, 74.27719)
(11, 83.35656)
};


\nextgroupplot[
    title={},  
    ylabel={\small{F1 (\%)}},
    xtick={1,2,3,4,5,6,7,8,9,10,11},
    xlabel=\empty, 
    ymin=0,
    ymax=65, 
     extra y ticks={47.65},
     extra y tick labels={}, 
    extra y tick style={grid=major, grid style={thick, dashed, black}}
]

\addplot[mark=*, thin, line width=0.25pt, mark options={fill=orange}, mark size=1.5pt] coordinates {
(1, 29.44511)
(2, 49.09091)
(3, 48.08367)
(4, 51.52295)
(5, 52.46429)
(6, 50.95238)
(7, 49.40736)
(8, 43.07416)
(9, 35.72022)
(10, 39.8658)
(11, 40.85708)
};

\addplot[mark=triangle*, thin, line width=0.25pt, mark options={fill=cyan}, mark size=2pt] coordinates {
(1, 29.44511)
(2, 45.7669)
(3, 47.96001)
(4, 52.48299)
(5, 52.48299)
(6, 52.48299)
(7, 52.48299)
(8, 52.48299)
(9, 52.24638)
(10, 49.68223)
(11, 50.08464)
};



\nextgroupplot[
    title={},  
    ylabel={\small{F1 (\%)}},
    xtick={1,2,3,4,5,6,7,8,9,10,11},
    xlabel=\empty, 
    ymin=10,
    ymax=90, 
     extra y ticks={83.00},
     extra y tick labels={}, 
    extra y tick style={grid=major, grid style={thick, dashed, black}}
]

\addplot[mark=*, thin, line width=0.25pt, mark options={fill=orange}, mark size=1.5pt] coordinates {
(1, 41.5104)
(2, 33.87967)
(3, 39.28984)
(4, 40.35489)
(5, 47.72237)
(6, 67.93976)
(7, 66.0523)
(8, 69.45391)
(9, 75.93762)
(10, 69.86359)
(11, 77.75914)
};

\addplot[mark=triangle*, thin, line width=0.25pt, mark options={fill=cyan}, mark size=2pt] coordinates {
(1, 41.5104)
(2, 35.40693)
(3, 41.97037)
(4, 42.33288)
(5, 44.02737)
(6, 44.74939)
(7, 45.7584)
(8, 54.25977)
(9, 61.25627)
(10, 64.09861)
(11, 67.01982)
};


\nextgroupplot[
    title={},  
    ylabel={\small{F1 (\%)}},
    xtick={1,2,3,4,5,6,7,8,9,10,11},
    xlabel=\empty, 
    ymin=0,
    ymax=100, 
     extra y ticks={94.93},
     extra y tick labels={}, 
    extra y tick style={grid=major, grid style={thick, dashed, black}}
]

\addplot[mark=*, thin, line width=0.25pt, mark options={fill=orange}, mark size=1.5pt] coordinates {
(1, 18.11492)
(2, 3.75)
(3, 16.87273)
(4, 15.13342)
(5, 27.0007)
(6, 35.69139)
(7, 45.50221)
(8, 37.16727)
(9, 48.55402)
(10, 72.02053)
(11, 88.08161)
};

\addplot[mark=triangle*, thin, line width=0.25pt, mark options={fill=cyan}, mark size=2pt] coordinates {
(1, 18.11492)
(2, 13.01723)
(3, 18.25154)
(4, 17.01258)
(5, 22.69958)
(6, 24.00761)
(7, 36.19728)
(8, 37.15934)
(9, 50.6051)
(10, 66.22472)
(11, 74.21875)
};


\nextgroupplot[
    title={},  
    ylabel={\small{F1 (\%)}},
    xtick={1,2,3,4,5,6,7,8,9,10,11},
    xlabel=\empty, 
    ymin=0,
    ymax=65, 
     extra y ticks={49.34},
     extra y tick labels={}, 
    extra y tick style={grid=major, grid style={thick, dashed, black}}
]

\addplot[mark=*, thin, line width=0.25pt, mark options={fill=orange}, mark size=1.5pt] coordinates {
(1, 8.83652)
(2, 14.54545)
(3, 8.83652)
(4, 8.83652)
(5, 8.83652)
(6, 8.83652)
(7, 23.79832)
(8, 14.21246)
(9, 20.51517)
(10, 25.78406)
(11, 33.22437)
};

\addplot[mark=triangle*, thin, line width=0.25pt, mark options={fill=cyan}, mark size=2pt] coordinates {
(1, 8.83652)
(2, 12.68997)
(3, 8.83652)
(4, 8.83652)
(5, 8.83652)
(6, 8.83652)
(7, 8.83652)
(8, 8.83652)
(9, 8.83652)
(10, 20.51517)
(11, 20.51517)
};



\nextgroupplot[
    title={},  
    ylabel={\small{F1 (\%)}},
    xtick={1,2,3,4,5,6,7,8,9,10,11},
    ymin=10,
    ymax=90, 
     extra y ticks={85.24},
     extra y tick labels={}, 
    extra y tick style={grid=major, grid style={thick, dashed, black}}
]

\addplot[mark=*, thin, line width=0.25pt, mark options={fill=orange}, mark size=1.5pt] coordinates {
(1, 43.82956)
(2, 43.44611)
(3, 14.00213)
(4, 55.81161)
(5, 65.1037)
(6, 67.45548)
(7, 70.78094)
(8, 76.82769)
(9, 75.83238)
(10, 79.27046)
(11, 81.50745)
};

\addplot[mark=triangle*, thin, line width=0.25pt, mark options={fill=cyan}, mark size=2pt] coordinates {
(1, 43.82956)
(2, 43.94482)
(3, 14.00213)
(4, 14.00213)
(5, 14.00213)
(6, 15.07482)
(7, 24.11139)
(8, 31.65505)
(9, 44.01879)
(10, 60.08508)
(11, 67.23819)
};


\nextgroupplot[
    title={},  
    ylabel={\small{F1 (\%)}},
    xtick={1,2,3,4,5,6,7,8,9,10,11},
    ymin=0,
    ymax=100, 
     extra y ticks={95.00},
     extra y tick labels={}, 
    extra y tick style={grid=major, grid style={thick, dashed, black}}
]

\addplot[mark=*, thin, line width=0.25pt, mark options={fill=orange}, mark size=1.5pt] coordinates {
(1, 14.94949)
(2, 22.21822)
(3, 27.02882)
(4, 38.91637)
(5, 39.78317)
(6, 45.38431)
(7, 44.95006)
(8, 54.17743)
(9, 68.37018)
(10, 85.88435)
(11, 90.68394)
};

\addplot[mark=triangle*, thin, line width=0.25pt, mark options={fill=cyan}, mark size=2pt] coordinates {
(1, 14.94949)
(2, 17.75241)
(3, 18.40663)
(4, 25.30447)
(5, 28.79075)
(6, 35.62768)
(7, 38.05286)
(8, 44.64466)
(9, 47.64526)
(10, 65.74323)
(11, 82.30475)
};


\nextgroupplot[
    title={},  
    ylabel={\small{F1 (\%)}},
    xtick={1,2,3,4,5,6,7,8,9,10,11},
    ymin=0,
    ymax=65, 
     extra y ticks={59.01},
     extra y tick labels={}, 
    extra y tick style={grid=major, grid style={thick, dashed, black}}
]

\addplot[mark=*, thin, line width=0.25pt, mark options={fill=orange}, mark size=1.5pt] coordinates {
(1, 37.45106)
(2, 38.84074)
(3, 39.98501)
(4, 52.9289)
(5, 52.15129)
(6, 53.2795)
(7, 44.26237)
(8, 41.9908)
(9, 37.05128)
(10, 39.93982)
(11, 42.04927)
};

\addplot[mark=triangle*, thin, line width=0.25pt, mark options={fill=cyan}, mark size=2pt] coordinates {
(1, 37.45106)
(2, 43.25089)
(3, 45.26906)
(4, 48.0142)
(5, 52.64746)
(6, 52.48299)
(7, 53.40875)
(8, 52.48299)
(9, 50.39683)
(10, 50.94785)
(11, 50.94785)
};


\end{groupplot}

\node[font=\bfseries\tiny] at ($(group c1r1.north)+(0,0.2cm)$) {\textbf{AIDE} \citep{yang2023aide}};
\node[font=\bfseries\tiny] at ($(group c2r1.north)+(0,0.2cm)$) {\textbf{B4C} \citep{jain2015car}};
\node[font=\bfseries\tiny] at ($(group c3r1.north)+(0,0.2cm)$) {\textbf{ROAD} \citep{singh2022road}};

\end{tikzpicture}

\vspace{0.2cm}
\begin{tikzpicture}
\begin{axis}[
    hide axis,
    xmin=0, xmax=1, ymin=0, ymax=1,
    legend style={at={(0.0,0.0)}, anchor=center, legend columns=3},
    legend entries={\tiny{Single predictions},\tiny{Cumulative ensemble}}
]

\addlegendimage{mark=*,  thin, line width=0.25pt, mark options={fill=orange}, mark size=2.0pt}
\addlegendimage{mark=triangle*, thin, line width=0.25pt, mark options={fill=cyan}, mark size=2.0pt}

\end{axis}
\end{tikzpicture}

\caption{Performance overview for the single and cumulative ensembles based on the intermediate latent representations. The dashed lines indicate the best performance for each LL--PDL method performed on the final latent representation.}
\label{fig:inter}


\end{figure*}

For deeper DNNs, previous studies have indicated that the intermediate latent representations can already be used to classify before the final layer of a network \citep{montavon2011kernel,papyan2020prevalence,kothapallineural} (see Figure \ref{fig:intermed}). To reduce the required computations to produce uncertainty estimations further, \cite{qendro2021early} and \cite{meronen2024fixing} showed that using probabilistic last layers trained on these intermediate representations of a single network can improve the performance. Subsequently, these intermediate predictions can be incrementally used to create a model-internal ensemble, which combines the predictions from different depths of a network to improve the performance \citep{meronen2024fixing}.  This potentially enables to not having to compute the entire network but to rely on intermediate representations for confident prediction and uncertainty estimation. 

Processing a single 16-frame video requires approximately $361 \times 10^9$ FLOPs for a ViT-Base model, where each Transformer block accounts for $29.8 \times 10^9$ FLOPs. Given the resource-constrained operating environment of DAR and DIR, and prior work highlighting the utility of intermediate representations for classification \citep{papyan2020prevalence, qendro2021early, kothapallineural, meronen2024fixing}, we evaluate LL--PDL methods' performance when trained on the representations extracted from the intermediate layers.  We extract the latent features from each block via average pooling and apply the top-performing  configurations to these intermediate features. Following \citep{qendro2021early, meronen2024fixing}, we assess whether predictions, either from individual layers or ensembles in the form of accumulated predictions from previous intermediate layers surpass the full network performance.


\paragraph{\textbf{Results.}} Figure \ref{fig:inter} illustrates the F1-score of the LL--PDL methods based on the intermediate latent representations of the ViT-Base model. For the AIDE and B4C datasets, the performance starts to saturate when relying on deeper latent representations, but never exceeds the final latent representation performance as reported in Table \ref{tab:performance}. For the ROAD dataset, the performance exhibits more variability, potentially due to the smaller dataset size. Only for the SE on the ROAD dataset, we observe improved performance based on the intermediate representation.









\end{document}